QUEEN MARY, UNIVERSITY OF LONDON

# Performing Bayesian Risk Aggregation using Discrete Approximation Algorithms with Graph Factorization

**Peng Lin**

December 2014

Submitted in partial fulfilment of the requirements of the degree of Doctor of Philosophy

# Declaration of originality

I, Peng Lin, confirm that the research included within this thesis is my own work or that where it has been carried out in collaboration with, or supported by others, that this is duly acknowledged below and my contribution indicated. Previously published material is also acknowledged below.

I attest that I have exercised reasonable care to ensure that the work is original, and does not to the best of my knowledge break any UK law, infringe any third party's copyright or other Intellectual Property Right, or contain any confidential material.

I accept that the College has the right to use plagiarism detection software to check the electronic version of the thesis.

I confirm that this thesis has not been previously submitted for the award of a degree by this or any other university.

Signature: 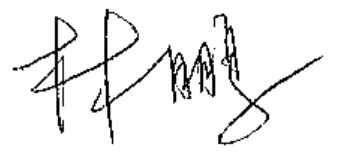

Date: December 31st, 2014

*To my parents*

# Acknowledgement

Firstly, I would like to thank my principal supervisor, Prof. Martin Neil. Without his enthusiasm and the effort spent on research discussions I would not have produced this work. He provided me with sound advice, rigorous standards and also free space for independent thinking.

I would also like to thank my second supervisor, Prof. Norman Fenton. He was a constant support and ever responsive, even when busy.

My parents are far away from the UK, but they are close to my research and they have followed every step of my progression. Without this constant support and care this research would not have been possible.

I thank those academics with whom I had stimulating and challenging research discussions: Dr. William Marsh (Queen Mary) and Dr. Talya Meltzer (Hebrew University). Dr. Anthony Constantinou and Dr. Barbaros Yet have conducted proof reading for this thesis. I thank all members in the Risk and Information Management group of Queen Mary.

I thank Yihan Tao for keeping me alive in good health. I warmly thank my friends and colleagues at Queen Mary, whose friendship helped sustain me during the difficulties, encountered my PhD studies and helped lighten the load.

Finally I thank Prof. Neil Alford (Imperial College), who recommended me to pursue a PhD in Queen Mary.

# Abstract

Risk aggregation is a popular method used to estimate the sum of a collection of financial assets or events, where each asset or event is modelled as a random variable. Applications, in the financial services industry, include insurance, operational risk, stress testing, and sensitivity analysis, but the problem is widely encountered in many other application domains.

This thesis has contributed two algorithms to perform Bayesian risk aggregation when model exhibit hybrid dependency and high dimensional inter-dependency. The first algorithm operates on a subset of the general problem, with an emphasis on convolution problems, in the presence of continuous and discrete variables (so called hybrid models) and the second algorithm offer a universal method for general purpose inference over much wider classes of Bayesian Network models.

The first algorithm is called the *Bayesian Factorization and Elimination* (BFE) algorithm which performs convolution on the hybrid models required to aggregate risk in the presence of causal dependencies. This algorithm exploits a number of advances from the field of Bayesian Networks, covering methods to approximate statistical and conditionally deterministic functions to factorize multivariate distributions for efficient computation. This algorithm aims to support the representation of Bayesian "views" in an explicit causal dependent structure, whilst providing the computational framework for evaluating convolution models. Such causal models would involve discrete explanatory (regime switching) variables, hybrid mixtures of dependent discrete and continuous variables, and high dimensional inter-dependent continuous variables.

The second algorithm developed is called *Dynamic Discretized Belief Propagation* (DDBP). It combines a dynamic discretization approach, to approximate continuous variables, with a new *Triplet Region Construction* (TRC) algorithm to perform inference on high dimensional hybrid models. The TRC algorithm is an optimized region graph approach based on graph factorization and Generalized Belief Propagation (GBP), which reduces the model complexity from exponential to polynomial. Proofs and experiments show that the algorithm converges, meets the requirements for a balanced, maximum entropy normal region graph and does not restrain the model to any particular parameterization. DDBP offers a general purpose solution to inference in hybrid Bayesian Networks of any size regardless of dimensionality, provided the model is binary factorizable, which may be inconvenient to solve by traditional algorithms. Experiments show that it produces comparably accurate result with exact values.

## Table of Contents

# Glossary of Abbreviations

| | |
|---|---|
| BF | Binary Factorization |
| BFE | Bayesian Factorization and Elimination |
| BFG | Binary Factorized Graph |
| BM | Bethe Method |
| BN | Bayesian Network |
| BP | Belief Propagation |
| CDF | Compound Density Factorization |
| CG-DCCD | Conditional Gaussian-Densely Connected Chain DAG |
| CI | Conditional Independence |
| CPD | Conditional Probability Distribution |
| CTE | Cluster Tree Elimination |
| CVM | Cluster Variation Method |
| DAG | Directed Acyclic Graph |
| DCCD | Densely Connected Chain DAG |
| DD | Dynamic Discretization |
| DDBP | Dynamic Discretized Belief Propagation |
| DDJT | Dynamic Discretized Junction Tree |
| EDBP | Edge Deletion Belief Propagation |
| EP | Expectation Propagation |
| FFT | Fast Fourier Transform |
| FG | Factor Graph |
| GBP | Generalized Belief Propagation |
| IJGP | Iterative Join Graph Propagation |
| JGM | Junction Graph Method |
| KL | Kullback-Leibler |
| LBA | Log Based Aggregation |
| LBP | Loopy Belief Propagation |
| MC | Monte Carlo |
| MC-BU | Mini-Clustering Belief Updating |
| MCMC | Markov Chain Monte Carlo |
| MGD | Multivariate Gaussian Distribution |
| MN | Markov Network |
| NBP | Non-parametric Belief Propagation |
| NPT | Node Probability Table |
| PBP | Particle Belief Propagation |
| RG | Region Graph |
| SP | Survey Propagation |
| TLSBP | Truncated Loop Series Belief Propagation |
| TRC | Triplet Region Construction |
| VE | Variable Elimination |
| VI | Variational Inference |

# List of Figures

# List of Tables

# List of Algorithms

# 1. Introduction

## 1.1. Motivation

Risk aggregation is a popular method used to estimate the sum of a collection of financial assets or events, where each asset or event is modelled as a random variable. Existing techniques make a number of assumptions about these random variables. Firstly, they are almost always continuous. Secondly, should they be dependent, these dependencies are best represented by correlation functions, such as copulas (Politou & Giudici, 2009) , where marginal distribution functions are linked by some dependence structure. These statistical methods have tended to model associations between variables as a purely phenomenological artefact extant in historical statistical data. Recent experience, at least since the beginning of the current financial crisis in 2007, has amply demonstrated the inability of these assumptions to handle non-linear effects or “shocks” on financial assets and events, resulting in models that are inadequate for prediction, stress testing and model comprehension (IMF, 2009), (Laeven & Valencia, 2008).

It has been extensively argued that modelling dependence as correlation is insufficient, since it ignores any views that the analyst may, quite properly, hold about those causal influences that help generate and explain the statistical data observed (Meucci, 2008), (Rebonato, 2010). Such causal influences are commonplace and permeate all levels of economic and financial discourse. For example, does a dramatic fall in equity prices cause an increase in equity implied volatilities or is it an increase in implied volatility that causes a fall in equity prices? The answer is trivial in this case, since a fall in equity prices is well known to affect implied volatility, but correlation alone contains no information about the direction of causation. To incorporate causation we need to involve the analyst or expert and “fold into” the model views of how discrete events interact and the effects of this interaction on the aggregation of risk. This approach extends the methodological boundaries last pushed back by the celebrated Black–Litterman model (Black & Litterman, 1991). In that approach a risk manager’s role is as an active participant in the risk modelling, and the role of the model is to accommodate their subjective "views", expressed as Bayesian priors of expectations and variances of asset returns. This thesis aims to represent these Bayesian "views" in an explicit causal structure, whilst providing the computational framework for solutions. Such causal models would involve discrete explanatory (regime switching) variables, hybrid dependent variables, and high dimensional inter-dependent variables. A causal risk aggregation model might incorporate expert derived views about macro-economic, behavioural, operational or strategic factors that might influence the assets or events under "normal" or "abnormal" conditions. Applications of the approach include insurance, stress testing, operational risk and sensitivity analysis, but the problem is widely encountered in many other application domains.

At its heart risk aggregation requires the sum of $n$ random variables. In practice this involves the use of two well-known mathematical operations: *n-fold* convolution (for a fixed value of $n$) and *N-fold* convolution (Heckman & Meyers, 1983), defined as the compound sum of a frequency distribution, $N$, and a severity distribution, $S$, where the number of constant *n-fold* convolutions is determined stochastically. Currently numerical tools such as Panjer's recursion (Panjer, 1981), Fast Fourier transforms (Heckman & Meyers, 1983) and Monte Carlo (MC) simulation (Meyers, 1980) perform risk aggregation numerically using parameters derived from historical data to estimate the distributions for both $S$ and $N$. These approaches produce acceptable results, but they have not been designed to cope with the new modelling challenges outlined above. In the context of modelling general dependencies among severity variables, a popular approach is to use copulas, both to model the dependent variables and to perform risk aggregation.

One aim of this thesis is to show how we can carry out a stochastic risk aggregation (*N-fold* convolution) in a causal Bayesian framework, in such a way that subjective views about inter-dependencies can be explicitly modelled and numerically evaluated i.e. where discrete and continuous variables are inter-dependent and may influence $N$ and $S$ in complex, non-linear ways. Conventional approaches, such as the copula based risk aggregation framework (Bruneton, 2011), cannot perform deconvolution because the model is usually not identifiable since it involves an inverse transform of the aggregated model, resulting in multiple solutions. The algorithms described in this thesis support deconvolution (illustrated in chapter 4) because the discrete nature of the approximation is more tractable.

We see this as the first of many financial modelling problems that are amenable to this new approach.

This thesis describes the development in two parts. The first part involves Chapters 2 and 4, which describe a *Bayesian Factorization and Elimination* (BFE) algorithm that performs convolution on the hybrid models required to aggregate risk in the presence of causal dependencies. This algorithm exploits a number of advances from the field of Bayesian Networks (BNs) (Pearl, 1988) (F. V. Jensen & Nielsen, 2009), covering methods to approximate statistical and conditionally deterministic functions and to factorize multivariate distributions for efficient computation.

The second part is covered in Chapters 3 and 5, where the aim is to solve the problem on general inference on high dimensional models of any form, including non-Gaussian and discrete cases, and which can be extended to many classes of hybrid Bayesian Networks. In this way risk aggregation on high dimensional models can be carried out, but, as a natural by product, so can approximate inference on many other more general problems. A discrete approximation inference algorithm called *Dynamically Discretized Belief Propagation* (DDBP) is presented to perform inference on models exhibiting high dimensionality; a major computation barrier existed for current discrete approximation algorithms, such as the Junction Tree (JT) algorithm. DDBP uses graph factorization, provided the model is binary factorizable, and uses Generalized Belief Propagation (GBP) (Yedidia, Freeman, & Weiss, 2005) for discrete approximation. It significantly reduces the computational complexity from (worst case) exponential to polynomial compared with Junction Tree based approaches. The DDBP algorithm contains a sub algorithm called *Triplet Region Construction* (TRC), derived from GBP (Yedidia et al., 2005) that builds an optimal region graph satisfying the maxent-normal (Yedidia et al., 2005) property and guaranteeing a balanced propagation structure. The TRC algorithm solves several related problems, e.g. how to determine the graph topology, interaction strength and ensures the counting number of the region graph is balanced. When coupled with Dynamic Discretization (DD) the TRC algorithm produces approximate results on a wide class of BN models and experiments demonstrate that such models converge and are accurate.

The BFE and DDBP algorithms are major contributions of this research. They both rely on factorizing the graph structure in principled ways and both use discrete approximation methods, using dynamic discretization (DD), to approximate continuous variables. The thesis compares the results produced from these against conventional numerical algorithm, such as Markov Chain Monte Carlo (MCMC) (Gamerman & Lopes, 2006), and with analytical closed form solutions. The algorithms developed by this research, however, have the advantages to be more flexible, extendable and also just as accurate as the existing state of the art.

# 1.2. Research Hypotheses

The main research objective of this thesis is to perform a stochastic risk aggregation using a causal Bayesian framework, typically such models could contain independent random variables, hybrid dependent random variables and high dimensional inter-dependent random variables. So this research is carried out to answer the following four research hypotheses.

- First, can we perform a Bayesian stochastic risk aggregation using discrete approximation approach to accurately address the compound density when severity variables are independent?
- Second, can we perform a Bayesian stochastic risk aggregation using discrete approximation approach to accurately address the compound density when severity variables are hybrid and dependent?
- Third, can we accurately de-convolute the causal explanatory variables by using discrete approximation approach for the Bayesian stochastic risk aggregation model?
- Fourth, can we perform a Bayesian stochastic risk aggregation using discrete approximation approach to accurately address the compound density when severity variables exhibiting high dimensionality?

This thesis will describe the solutions for all these hypotheses, supported by experiments and proofs. The reader should be aware however that the research results demonstrated answer hypothetical questions beyond those outlined above and as such the thesis exceeds the ambitions of the original research agenda.

## 1.3. Structure of the Thesis

Chapter 2 reviews the background of Bayesian Network inference with an emphasis on factorization, the central orientation of this thesis. It covers Bayes' theorem, Bayesian networks and other supporting structures (such as Markov Graphs). The graph factorization approach is presented, which is then used for the BFE and DDBP algorithms. It also illustrates how deterministic conditional functions are factorized for continuous variables.

Chapter 3 covers the inference approaches for Bayesian networks. It reviews the current popular inference approaches, e.g. JT, VI, MCMC and GBP.

Chapter 4 introduces the BFE algorithm for *N-fold* convolution models. The severity models considered are independent and hybrid dependent random variables. This chapter presents the factorization and variable elimination techniques, which are central components of BFE algorithm. Next the de-convolution of the Bayesian risk aggregation model is illustrated. This chapter focuses on research hypotheses one, two and three.

Chapter 5 develops an optimal region graph construction algorithm called *triplet region construction* (TRC), which is a sub algorithm for DDBP. It covers the mechanics of constructing an optimized region graph derived from GBP algorithm, which is automatic and satisfies maxent-normal property. It then develops the DDBP algorithm by combining TRC and dynamic discretization. The experiments in this chapter illustrate the Bayesian risk aggregation on high dimensional models, and also on general models. This chapter focuses on research hypothesis four.

Chapter 6 concludes the thesis, discusses the overall value of this research and highlights several improvements to the BFE and DDBP algorithms to achieve better performance.

## 1.4. Publications

Until the submission of this thesis:

1. Lin, P., Neil, M., & Fenton, N. (2014). Risk aggregation in the presence of discrete causally connected random variables. *Annals of Actuarial Science*, 8 (2), pp 298-319 doi:10.1017/S1748499514000098
2. Lin, P., Neil, M., & Fenton, N. (2014). Inference for high dimensional Bayesian network models using dynamically discretized belief propagation, submitted to *Information Sciences.*

# 2. Belief Networks and Graph Factorization

This chapter provides an overview of the fundamental probability concepts central to this thesis. We will discuss Bayesian probability and (touch on) causality. Bayesian networks and other network structures will be covered in more detail and approaches to graph factorization will be given particular emphasis. The concepts presented in this chapter are used to describe the algorithms presented in the following chapters of this thesis.

## 2.1. Bayesian Probability

Bayesian probability (Cox, 1961) (Finetti, 1970) provides a way to reason coherently toward the uncertainty we face in the real world. It reflects personal knowledge and any belief about uncertainty is assumed to be provisional on experience gained to date (the prior), is then updated by new experience and data (the likelihood) to provide a new personal disposition about the uncertainty (the posterior). Probabilities are quantitative measures of this uncertainty on a unit scale and subject to the axioms of probability theory (Devore, 2011).

However, probability is not only about numbers, but also about the structure of reasoning (Pearl, 1988), which can be used to reason causally (i.e. from cause to effect or vice versa). Indeed in recent decades Bayesian probability has gained in popularity because Bayes' theorem supports reasoning about causal propositions of how the variables are related along with the beliefs (probabilities) between these variables.

Bayesian probability was initially proposed by Thomas Bayes in 18$^{th}$ century, and pioneered and popularised by Laplace (Stigler, 1986). It can be expressed as Equation 2.1.

$$p(\theta \mid E) = \frac{p(\theta, E)}{p(E)} = \frac{p(E \mid \theta) \cdot p(\theta)}{p(E)} \tag{2.1}$$

The last term of Equation 2.1 is known as Bayes' rule (Fenton & Neil, 2012), and is constantly used in this thesis. Here variable $E$ is an observable data point/evidence and $\theta$ is a model parameter/hypothesis. $p(\theta, E)$ is the joint probability of $\theta$ and $E$, which represent the problem domain containing these two variables. By the chain rule of probability (Russell & Norvig, 2010) the last term of Equation 2.1 factorizes the joint probability into conditional probability (the likelihood) and the prior. If we can organise this factorize appropriately then the number of parameters might be substantially reduced (will discuss in section 2.2). Such factorization might reflect the causal nature of the problem or may be subject to mathematical rules governing the decomposition lying behind the parameterization.

In Bayesian interpretation of probability, probability is a measure of belief, and $p(\theta)$ is the probability of some hypotheses or parameters, $\theta$, referred to as the *prior*. It is the belief about $\theta$ without knowing any information about the evidence, $E$. $p(E \mid \theta)$ is the conditional belief about the evidence, $E$, given information about the hypotheses or parameters, $\theta$, and is referred to as the *likelihood*. $p(E)$ is used as a *normalizing factor* which is the probability of observing the evidence over all hypotheses. $p(E)$ is a marginal probability obtained by the *marginalization* operation over the joint probability $p(\theta, E)$, and in the discrete form can be expressed as Equation 2.2.

$$p(E) = \sum_{\theta} p(\theta, E) = \sum_{\theta} p(E \mid \theta) \cdot p(\theta) \tag{2.2}$$

# 2.2. Factorization and Bayesian Belief Networks

Direct computation of joint probability is inefficient or even intractable. The complexity problem of the computation of joint probability can be overcome by exploiting conditional independence (Dawid, 1979) (Pearl, 1988).

Following the chain rule of probability, we can factorize the joint probability $p(X_1, X_2, ..., X_n)$ into Equation 2.3.

$$p(X_1, X_2, ..., X_n) = p(X_1) \cdot p(X_2 \mid X_1) ... \cdot p(X_n \mid X_1, ..., X_{n-1})$$
$$= \prod_{i=1}^{n} p(X_i \mid X_1, ..., X_{i-1}) \quad (2.3)$$

Factorization Equation 2.3 does not imply any benefit for parameter reduction, however, domain knowledge usually allow us to identify a subset of *parent variables*, $pa\{X_i\} \subseteq \{X_1, X_2, ..., X_{i-1}\}$, such that given $pa\{X_i\}$, variable $X_i$ is conditional independent of all variables in $\{X_1, X_2, ..., X_n\} \setminus pa\{X_i\}$. The parameterization might be therefore being expressed as Equation 2.4, thus reducing dimensionality.

$$p(X_i \mid X_1, ..., X_{i-1}) = p(X_i \mid pa\{X_i\}) \quad (2.4)$$

And further we can modularise the joint probability distribution as the product of these local *child variables*, $X_i$, conditioned on parents, $pa(X_i)$, as shown in Equation 2.5.

$$p(X_1, X_2, ..., X_n) = \prod_{i=1}^{n} p(X_i \mid pa\{X_i\}) \quad (2.5)$$

In this way we use independence and dependence assumptions to reduce the parameter complexity. We can consider a BN (defined below) as simple a set of conditional independence assumptions (Barber, 2012) and can be used to simplify the representation/factorization of joint distributions. In the remainder of this section, we give a recap of the properties of BNs.

A BN is a graphical probabilistic model that represents a set of random variables and the conditional independence assumptions between those variables. A BN can be represented by a *directed acyclic graph* (DAG), with a directed arc pointing from a parent variable to child variable, and the $i\,th$ node in the graph corresponding to the factor $p(X_i \mid pa\{X_i\})$ (Barber, 2012).

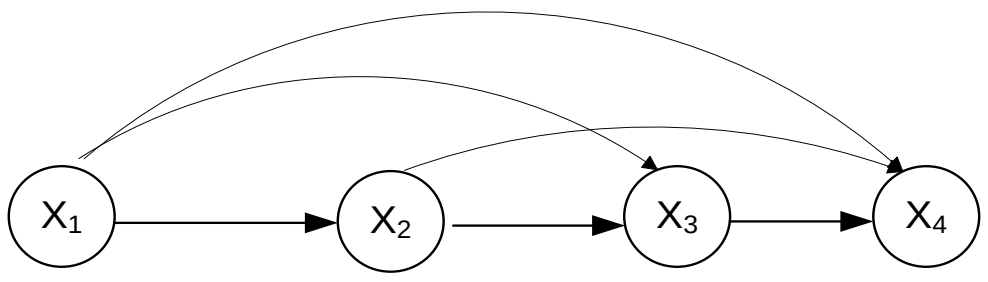

Figure 2.1 BN with four variables distribution

A Bayesian Network (BN) consists of two main elements:

1. *Qualitative*: This is given by a DAG $G$, with nodes representing random variables (parent or child), which can be discrete or continuous, and may or may not be observable, and directed arcs (from parent to child) representing causal or influential relationships between variables. See Figure 2.1 for an example BN.
2. *Quantitative*: A probability distribution associated with each node $X_i$. For a node with parents this is a *conditional probability distribution* (CPD), $p\left(X_i \mid pa\{X_i\}\right)$ that defines the probabilistic relationship of node given its respective parents $pa\{X_i\}$. For each node $X_i$ without parents, called root nodes, this is their marginal probability distribution $p(X_i)$. If $X_i$ is discrete, the CPD can be represented as a *Node Probability Table* (NPT) (Fenton & Neil, 2012), $p\left(X_i \mid pa\{X_i\}\right)$, which lists the probability that $X_i$ takes, on each of its different values, for each combination of values of its parents $pa\{X_i\}$. For continuous variables, the CPDs represent conditional probability density functions.

Together, the qualitative and quantitative parts of the BN encode all relevant information contained in a full joint probability model. The conditional independence (CI) assertions about the variables, represented by the absence of arcs, allow factorization of the underlying joint probability distribution in a compact way as a product of CPDs. More detailed information on BNs can be obtained from (F. V. Jensen & Nielsen, 2009) (Fenton & Neil, 2012) (Koller & Friedman, 2009) (Pearl, 1993) (S. L. Lauritzen, 1996).

Any joint probability distribution factorized by the chain rule can be represented as a complete graph, where each pair of nodes is connected by a directed edge. A variation of BNs that is complete graph is called DCCD and is defined:

> **Definition:** A DAG, $G$, factorized by the chain rule is defined as a *Densely Connected Chain DAG* (DCCD) if each pair of variables are connected by a directed edge, and each variable $X_i$ is directly connected to all of its parents $pa\{X_i\}$. The DCCD representation encodes no independence constraints.

A DAG $G$ that is ***not*** a DCCD, is referred to as a ***sparse graph***. A DCCD can be discrete, continuous or hybrid (appearing discrete and continuous variables simultaneously) model. DCCDs will be used to illustrate the algorithms presented in the rest of this thesis.

Figure 2.1 is a DCCD, representing a four variables distribution $p(X_1, X_2, X_3, X_4)$, which can represent an arbitrary four dimensional joint distribution. If there is no conditional independence information available for this graph, the NPT space complexity (complexity for defining NPT entries) is $O(d^n)$, where $n$ is the number of random variables and $d$ is the number of discretized states of a variable. Conditional independence assumption on a DCCD model implies that there exists some edges can be removed, resulting in a sparse graph. By using conditional independence assumption, we can reduce:

1. NPT space complexity: to store full joint distribution requires $O(d^n)$.
2. Inference complexity: to compute some queries requires $O(d^n)$ steps.

However, CI assumption is problem tailored and model specific, which is not a general method for reducing the NPT complexity, i.e. not applicable to DCCD models. We will introduce NPT decomposition approach in section 2.4.

The *Markov assumption* states that each random variable, $X$, is independent of its non-descendents given its parents, $pa\{X\}$ (Koller & Friedman, 2009). This can be written as $Ind(X; NonDesc(X) | pa\{X\})$. A DAG, $G$, is an *I-map* of a distribution $p$ if all Markov assumptions implied by $G$ are satisfied by $p$, assuming $G$ and $p$ both use the same set of random variables (Koller & Friedman, 2009).

A DCCD $G$ suffices to encode all Markov assumptions for any distribution $p$, so a DCCD, $G$, is an I-map for arbitrary distribution $p$ (Barber, 2012), which is sufficient to represent arbitrary $p$ if $p$ can be factorized in the form of Equation 2.5.

So any BN model can be viewed as a DCCD with some edges being removed under the CI assumptions. Conversely, any non-DCCD model can be converted to a DCCD model by adding appropriate edges, and then CPDs need to be changed accordingly with some parameters set to zero to imply the CI assumptions. So theoretically any BN models can be represented by DCCD models. Any inference algorithm that works on the worst case (the DCCD case) BN will work for any BN (this is the case in Chapter 5 for inference task).

An arbitrary dimensional BN model can be represented as:

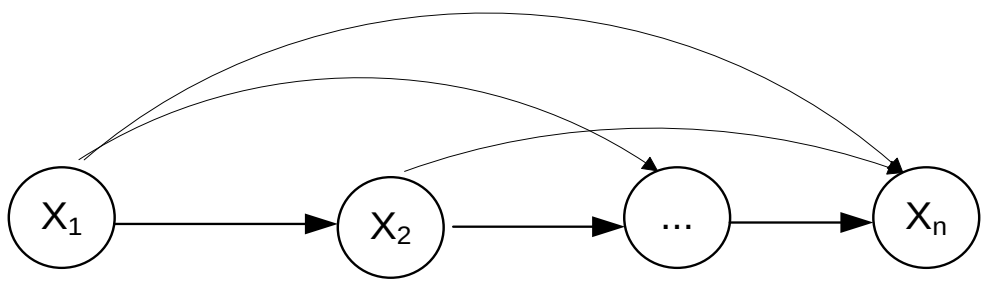

Figure 2.2 $n$ dimensional DCCD

Figure 2.2 shows an $n$ dimensional DCCD, with the dimensionality $n$ can be arbitrary. The DCCD models are general form of any BN model, where the difference is only on dimensions. So in the following sections/chapters, we will use DCCD models to illustrate concepts and develops. In chapter 5, the TRC algorithm is particularly designed for inference on DCCD models.

## 2.3. Causation and BN Structuring

The graphical structure and CI assumptions of a BN can be used to encode causal information into the model. This offers a unique mechanism to fuse information from data, say in the form of correlations but also in the form of subjective probabilities and reflecting, in the structure, statements about how the world might operate. We refer the details of discussions on causation in BNs to (Pearl, 2000) (Fenton & Neil, 2012). In this thesis, risk aggregation with discrete causal variables is presented in Chapter 4, where it demonstrates how causation dependencies are captured along with the aggregation. The causation relationships between random variables e.g. *cause to consequence or vice versa* are quantified by Bayesian inference. In this way, we can propagate/infer the observed data from either cause or consequence variable to other variables. The prior distributions are revised into posterior distributions after accounting for the newly observed data. Revising prior belief into posterior after observing new relevant evidence is believed to be consistent with human reasoning. The causal explanations can be modelled explicitly by three dependence connections in BBN as shown by the three following conditional independence assumptions:

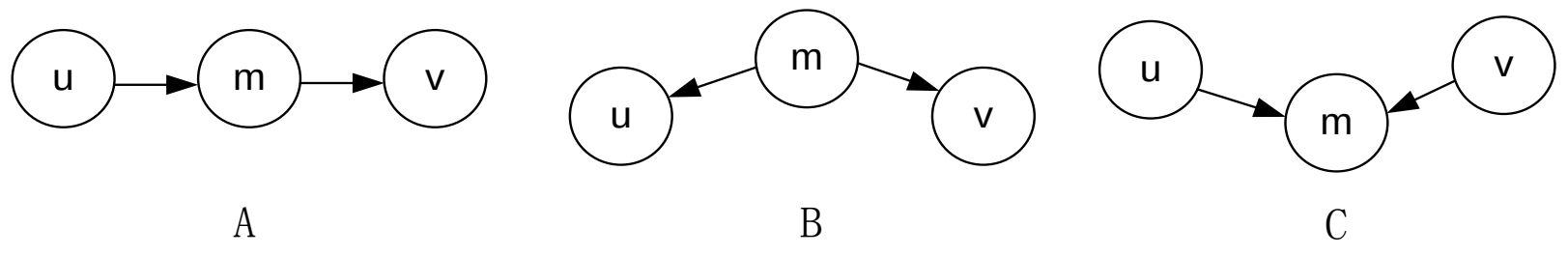

Figure 2.3 Dependence connections and its associated chain rule probability

A: serial connection, $p(u, m, v) = p(u) \cdot p(m \mid u) \cdot p(v \mid m)$

B: diverging connection, $p(u, m, v) = p(m) \cdot p(m \mid u) \cdot p(m \mid v)$

C: converging connection, $p(u, m, v) = p(u) \cdot p(v) \cdot p(m \mid u, v)$

Typically to scale up a BN it suffices to use the above three kinds of dependence connections shown in Figure 2.3. Formally, we introduce the DAG concepts of d-connection and d-separation ("d" stands for directional) that are central to determine CI in any BN with structure given by the DAG. Let $\mathbf{P}$ be a trial, (that is, a collection of edges which is like a path, but each of whose edges may have any direction) from node $u$ to $v$. Then $\mathbf{P}$ is said to be d-separated by a set of nodes $\mathbf{Z}$, if and only if one of the following holds.

1. $\mathbf{P}$ contains a serial connection, $u \to m \to v$, such that the middle node $m$ is in $\mathbf{Z}$.
2. $\mathbf{P}$ contains a diverging connection, $u \leftarrow m \to v$, such that the middle node $m$ is in $\mathbf{Z}$.
3. $\mathbf{P}$ contains a converging connection, $u \to m \leftarrow v$, such that the middle node $m$ is not in $\mathbf{Z}$ and no descendent of $m$ is in $\mathbf{Z}$.

Thus $u$ and $v$ are said to be d-separated by $\mathbf{Z}$ if all trails between them are d-separated. If $u$ and $v$ are not d-separated they are said to be d-connected. In a serial connection of Figure 2.3, if we instantiate $m$ then $u$ and $v$ are d-separated given $m$ (blocked), entering evidence at $u$ does not affect $v$ and vice versa. The same phenomenon applies for a diverging connection case. In the case of a converging connection, entering evidence on $u$ does not affect $v$ and vice versa since they are d-separated. If the certainty of $m$ changes then $u$ and $v$ become dependent. The notations we introduced here will be applied to chapter 4 and 5.

A BN $\mathbf{B}$ is often first developed by creating a causal DAG $G$, such that $\mathbf{B}$ satisfies the Markov assumption with respect to $G$. One then ascertain the CPDs of each variable given its parents in $G$. If in the case where the variables are discrete, if we define the joint distribution of **B** to be the product of these CPDs, then **B** is a Bayesian network with respect to $G$.

In the case where the variables are approximated by piece-wise constant partitions (discrete approximation of a continuous variable), the product of these CPDs can be significantly computational inefficient. The next section discusses, from a computation point of view, how the size of NPTs can be reduced, which is a crucial mechanism used to develop efficient numerical algorithms in Chapter 4 and 5.

## 2.4. Decomposing Node Probability Tables

The dimension of an NPT is simply the enumeration of all combinations of child and parent variable discrete states where the complexity is exponential in the number of variables involved.

This section explains an approach to NPT decomposition called *Binary Factorization* (BF) (Neil, Chen, & Fenton, 2012), which is able to factorize conditional expressions used in the definition of an NPT in such a way that it guarantees that each conditional continuous variable in the BN has no more than two parent variables. In this way the size of the NPT is reduced from an exponential to the sum of NPTs of polynomial size.

Conditional expressions that can be binary factorized in this way include conditionally deterministic expressions and statistical distributions whose parameters are defined as functions of parent variables.

NPT decomposition is not only important in reducing the complexity of the NPTs but also facilitates the development of the risk aggregation algorithm (Chapter 4) and inference algorithms (Chapter 5) in this thesis.

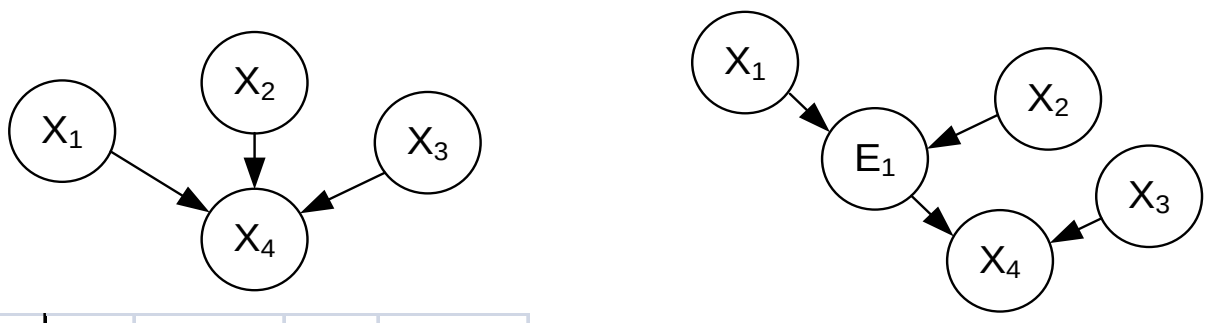

| NPT | X1 | a1 | … | a25 |
|---|---|---|---|---|
| | X2 | b1 … | … | …b25 |
| | X3 | c1…c25 | … | c1…c25 |
| X4 | d1 | … | … | … |
| | … | … | … | … |
| | d25 | … | … | … |

(a)

| NPT | E1 | e1 | … | e25 |
|---|---|---|---|---|
| | X3 | c1…c25 | … | c1…c25 |
| X4 | d1 | … | … | … |
| | … | … | … | … |
| | d25 | … | … | … |

(b)

Figure 2.4 Binary factorization mechanics:

(a) NPT table for $p(X_4 \mid X_1, X_2, X_3)$

(b) NPT table for $p(X_4 \mid E_1, X_3)$ after binary factorization

Consider the conditionally deterministic function for a variable, $X_4$, and its parents: $X_4 = X_1 + X_2 + X_3$, as shown in Figure 2.4 (a). If all four variables are continuous, and each variable is approximated by 25 piece-wise constant partitions, the overall NPT size for a single CPD $p(X_4 \mid X_1, X_2, X_3)$ is $25^4$. To reduce the computation complexity, we use BF to factorize the conditional expressions defined for each conditional NPT, so the resulting NPTs defined on continuous conditioned variables will only contain three variables. In this way the maximum NPT size for its CPD is $25^3$, as shown in Figure 2.4 (b) and the size of the conditional NPTs is now $(m-1)n^3 < n^m$ where $m$ is the number of parent variables and $n$ is the number of states.

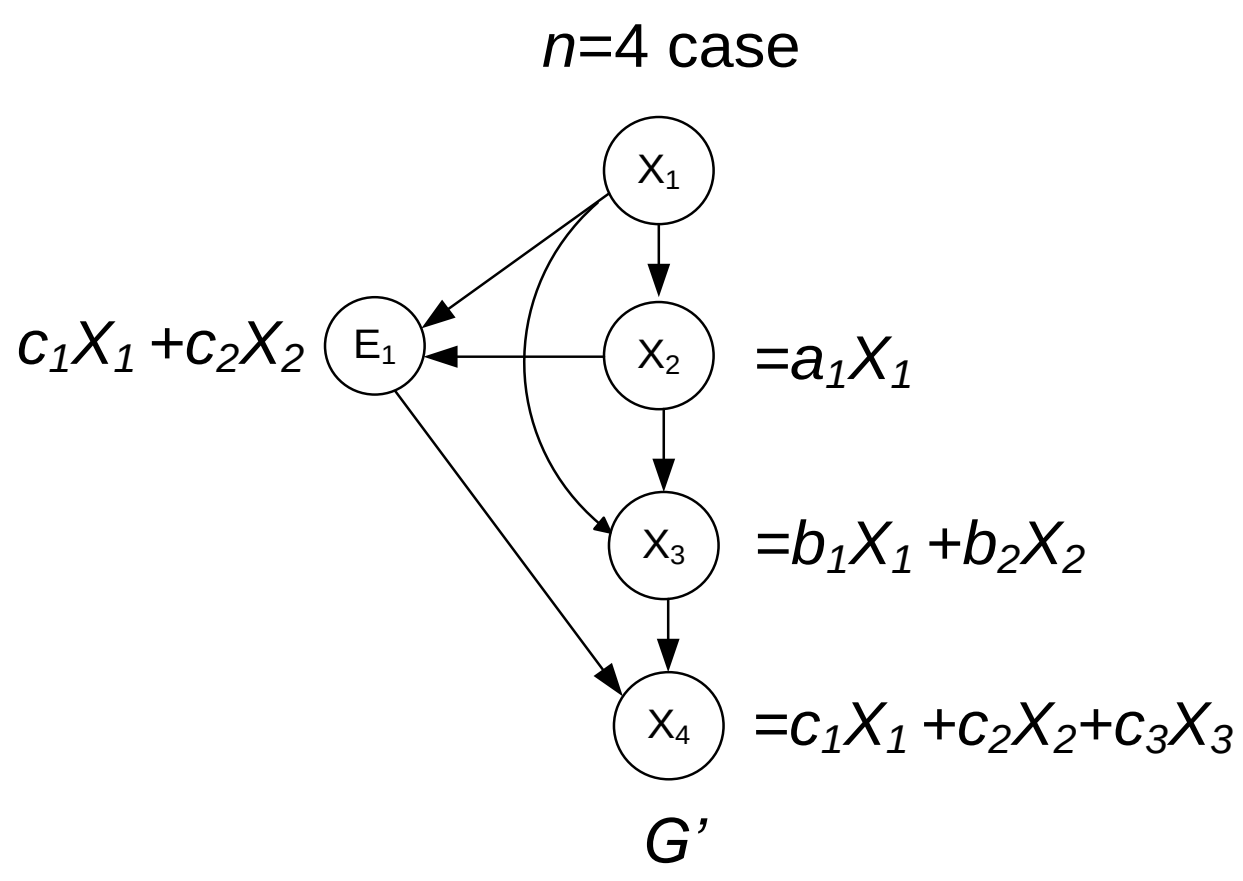

Figure 2.5 Binary factorization of 4 dimensional DCCD model

If all four variables in Figure 2.1 are continuous and the associated CPDs are conditional deterministic functions, we can perform BF process and obtain its BF formatted version $G'$ in Figure 2.5.

In Figure 2.5, the BF process factorizes the conditional deterministic expression for Figure 2.1:

$$X_4 = c_1X_1 + c_2X_2 + c_3X_3$$

into $X_4 = E_1 + c_3X_3$, where $E_1 = c_1X_1 + c_2X_2$ is an intermediate variable added to substitute the other two parents. By using the BF process, each continuous variable has only two parents. This produces a maximal discretized NPT of size $m^3$ ($m$ is the average number of piece-wise constant partitions for each variable) rather than $m^4$ for the case we considered. Note that the BF process will not work on a CPD with three or more parameters that cannot be decomposed, such as a hyper Geometric distribution.

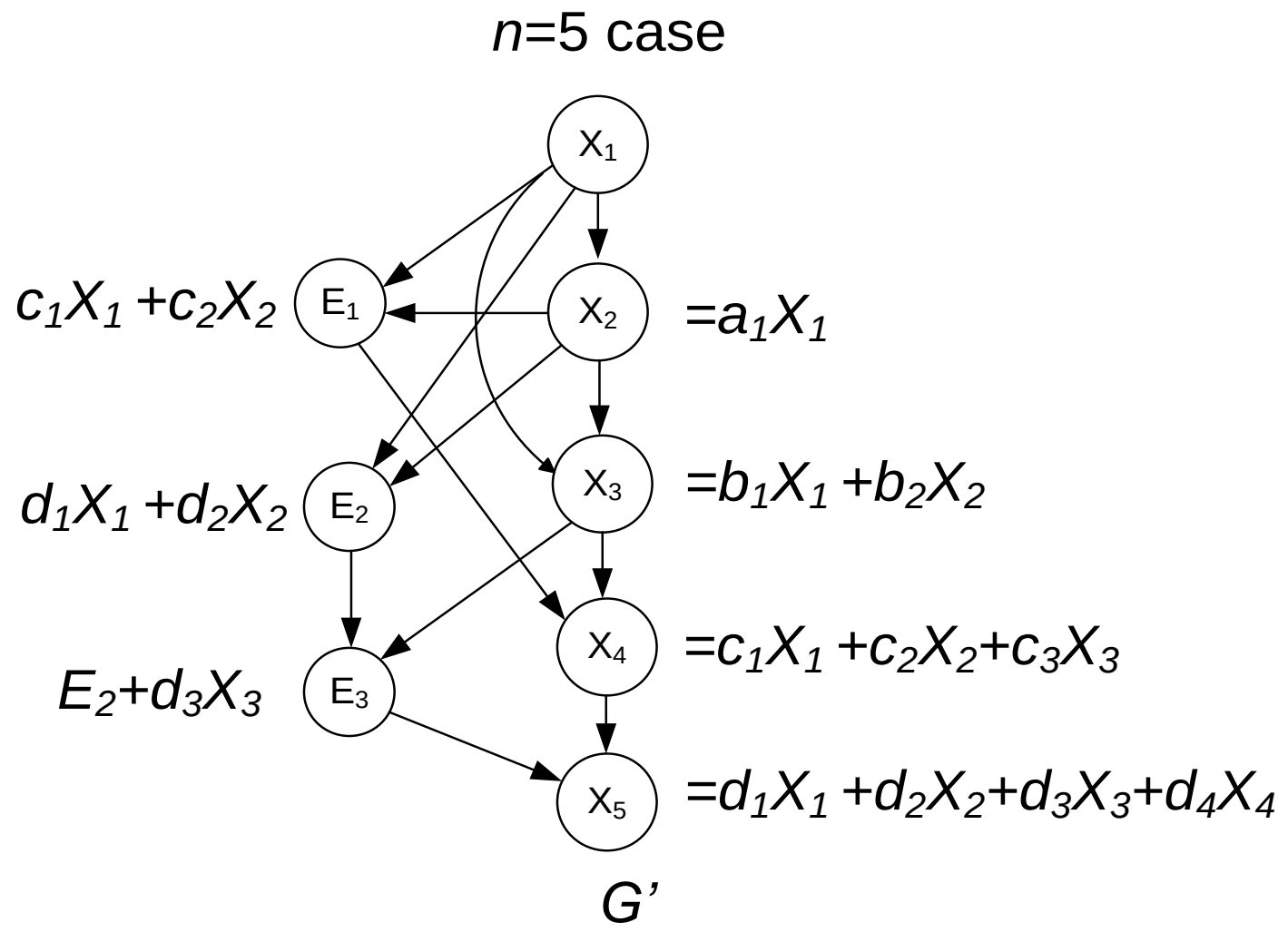

Figure 2.6 Binary factorization of 5 dimensional DCCD model

Similarly, we can transform a 5 dimensional DCCD model to its binary factorized form shown by Figure 2.6.

In the following sections, we will assume all DCCD models have already undergone a BF transform and inference is carried out on the *Binary Factorized Graph* (BFG) rather than the original BN. We have implicitly assumed all these DCCD models are *BF decomposable*, which means CPDs in the DCCD are either factorizable, e.g. for continuous variable case, or can be represented in a BFG form, i.e. discrete variables case.

The BFG makes our discrete inference feasible to perform DCCD models but under the BF decomposable assumption. Beside that we will guarantee the DCCD to BFG model is *I-equivalent,* which means network structures' encode the same independent statements. So that the BFG is sufficient to represent any BF decomposable distribution $p$ implied by DCCD. Such a BFG is unique and can be obtained by extending the BF process with an additional property. We define a *full-BFG,* $G'$:

> **Definition:** A full-BFG, $G'$, is a BFG where for any $i < j < k$ where $k > 2$ it is possible to find paths from $X_i$ to $X_j$ and $X_i$ to $X_k$ through intermediate variables $E_t$ in $G'$, where the set of intermediate variables are disjoint in the two paths [1].

Following conversion to a full-BFG each original variable $X_i \in G$ now has at most two parents in $G'$, and each new intermediate variable $E_t$ in $G'$ has one and only one child in $G'$. The disjoint property is important because it guarantees that the resulting full-BFG is an I-map of the original joint distribution, $p$.

To create a full-BFG we perform binary factorization on all nodes in the order of the node parents numbering in the original BN $G$. So, if a node has three parents $X_1$, $X_2$, $X_3$ then the binary factorization will be in the order $(X_1, X_2)$ and $X_3$ rather than $X_1$ and $(X_2, X_3)$ or $(X_1, X_3)$ and $X_2$. In Figure 2.6 $X_1, X_2, X_3$ have no more than two parents in their CPD expressions and so need no factorizations, and the paths $X_1 \rightarrow X_4$ and $X_1 \rightarrow X_5$, pass through the disjoint sets of intermediate variables.

A BFG $G'$ that does not satisfy the above constraints will not be I-equivalent to its DCCD, and hence it will be insufficient to represent an arbitrary (BF decomposable) distribution, $p$. Example 2.2 illustrates this point using a counter example.

[1] We require $k > 2$ since root nodes $X_1$ and $X_2$ are directly connected to $X_3$.

**Example 2.2**

Consider the full-BFG in Figure 2.7 which we present again in Figure 2.7 as $G'$ (with the graph reorganized). If $c_1 = d_1$ and $c_2 = d_2$, then the CPD expressions for $E_1$ and $E_2$ become identical. We could therefore create a binary factorized graph $G''$ in which these two intermediate variables are merged into one variable, $E_1$ (Figure 2.7). But then the disjoint sets defined by the full-BFG no longer hold because the paths $X_1 \to X_4$ and $X_1 \to X_5$ have a shared intermediate variable, $E_1$. The graph $G''$ is not I-equivalent with its DCCD for an arbitrary distribution $p$ since it cannot represent the distribution $p$ when $c_1 \neq d_1$ and/or $c_2 \neq d_2$.

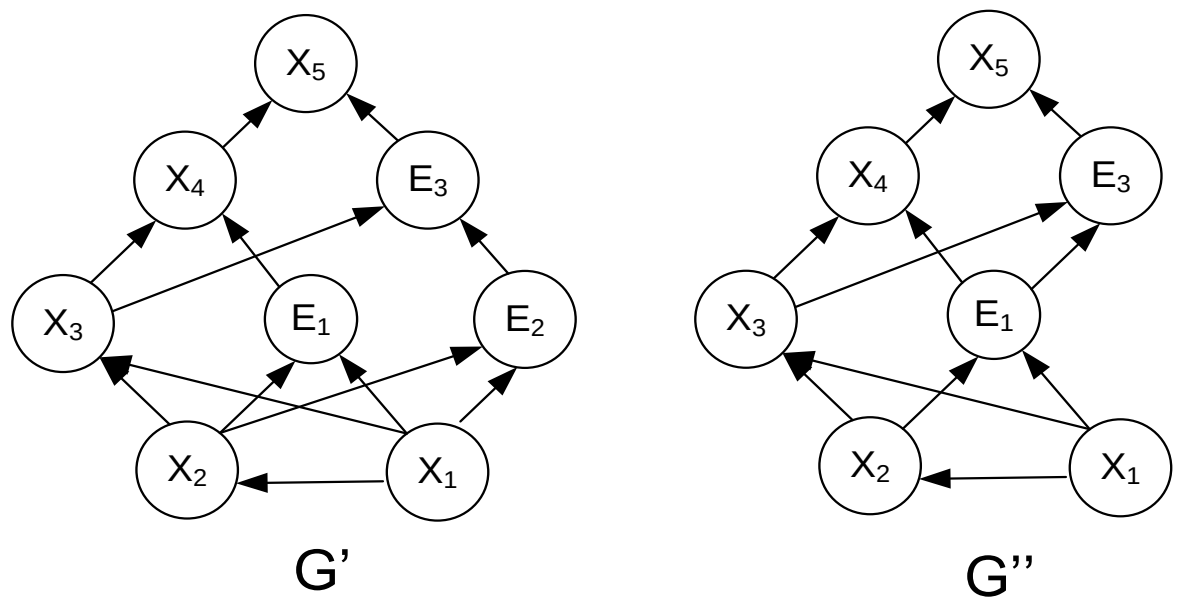

Figure 2.7 Counter example of complete BFG

In the case of $G''$ it is easy to see that if, for each intermediate variable, we add edges connecting its parent variables to its child variables and then remove the intermediate variables and all connecting edges, the resulting graph is not a DCCD. For $G'$ applying the same process results in a DCCD.

Except notations, we will assume all models discussed in the following sections are full-BFGs. So in general, the BF transform of an $n$-dimensional DCCD model results in a $\kappa_n$ dimensional full BFG model, where the number of nodes $\kappa_n$, can be determined by the sum of $n$ original variables and the inserted number of intermediate variables, as in Equation 2.6.

$$\kappa_n = n + \frac{(n-2)(n-3)}{2} = \frac{n^2 - 3n + 6}{2} \tag{2.6}$$

The complete BF notation introduced in this section will be applied to chapter 5 for developing TRC algorithm.

## 2.5. Converting a sparse graph into a DCCD

A continuous sparse graph model, $G$, can be converted to a DCCD, $G'$, while preserving associated CPDs by:

1. Ordering all original variables from parent to child, with children allocated higher valued labels than their parents;
2. Adding edges for each variable to all of its higher labelled descendant variables (Each variable is then connected via a path to its descendants);
3. Assigning each original variable the new CPD defined on each variable and its ancestors, by blocking those unrelated ancestors {*ancestors\parents*} by setting zeros, as weights, to the new conditionally deterministic expression, with respect to the variables in {*ancestors\parents*}.

If a CPD is conditionally deterministic: $X_5 = X_1 + X_2 + X_4 = X_1 + X_2 + 0 \times X_3 + X_4$, the expression for $X_5$ is defined on three parents, but it also can be defined on all of its four ancestors to satisfy a DCCD, with a block on one ancestor, in this case $X_3$, as encoded in the DCCD with a zero. Once we have converted the sparse graph to a DCCD we can then convert it to a full-BFG.

For discrete and hybrid BNs, the conversion is feasible, but is subject to ongoing research.

**Example 2.3**

Here we transform a sparse BN, $G$, into a DCCD and then a $\kappa_5$ full-BFG.

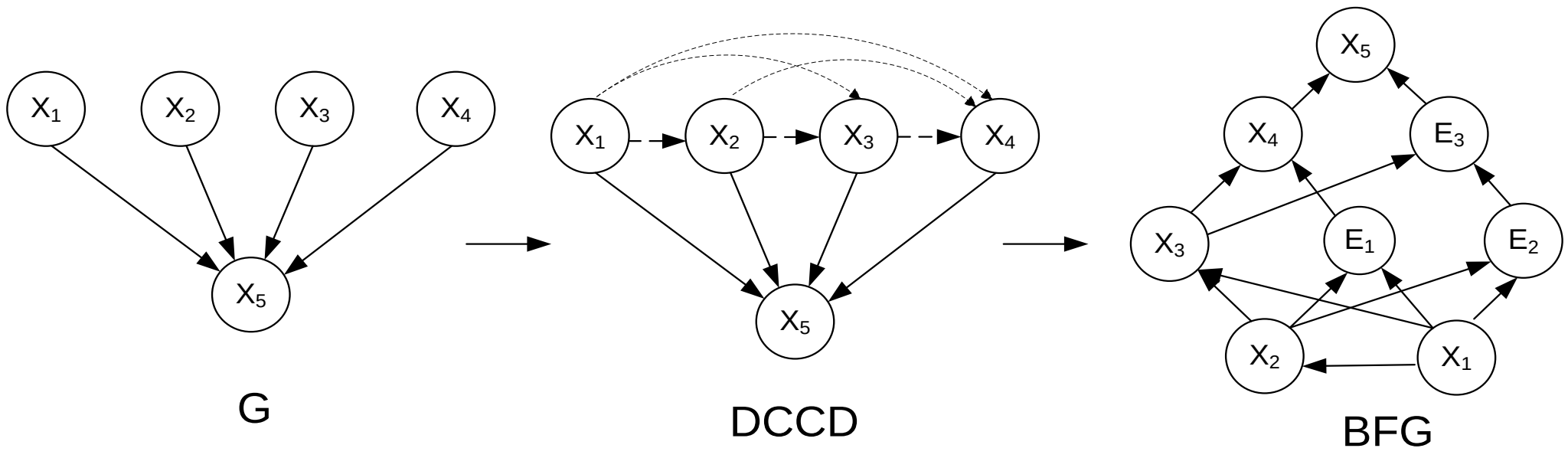

Figure 2.8 conversion of a sparse 5 dimensional BN to its full BFG

We consider a sparse original BN $G$ in Figure 2.8. The variables are labelled in parent to child order. We start from lowest labelled variable, $X_1$, so it adds edges to all its higher labelled variables, $X_2, X_3, X_4$ (dashed lines in DCCD in Figure 2.8). Then $X_2$ adds edges to all its higher labelled variables, $X_3, X_4$. This process repeats until all variables are connected to its higher labelled variables by directed edges. Next each variable needs a new CPD to accommodate its new parents. In the case of $X_3$, if the original CPD was defined as $f(X_3) \sim N(\mu = 5, \sigma^2 = 10)$ then its new CPD is changed to $f(X_3 \mid X_1, X_2) \sim N(\mu = 0 \times X_1 + 0 \times X_2 + 5, \sigma^2 = 10)$. After we include the intermediate variables we have the full-BFG shown in Figure 2.8.

# 3. Inference in Belief Networks

This chapter discusses the popular exact and approximation inference of BN algorithms. Section 3.1 deals with exact inference, 3.2 with approximate inference and 3.3 and 3.4 are particularly focused on developments related to the TRC and DDBP algorithms.

## 3.1. Exact Inference

There are different types of inference that can be performed in BNs, such as marginal inference, max-product inference and inference through finding the most probable states  (Barber, 2012). In this thesis we focus our discussion solely on marginal inference. Marginal inference is concerned with the computation of the distribution of a subset of variables, with possible conditioning on another subset.

Consider a 5 dimensional DCCD, with joint distribution:

$P(X_1 = x_1, X_2 = x_2, X_3 = x_3, X_4 = x_4, X_5 = x_5)$ and evidence $X_1 = e$ . We can perform marginal inference to query a subset joint distribution $P(X_2 = x_2, X_1 = e)$ as equation 3.1.

$$P(X_2 = x_2, X_1 = e) = \sum_{x_3, x_4, x_5} P(X_1 = e, X_2 = x_2, X_3 = x_3, X_4 = x_4, X_5 = x_5) \tag{3.1}$$

The right term in equation 3.1 can be expressed by the chain rule of CPDs implied by the DAG $G$ in equation 3.2.

$$\begin{aligned} P(X_2 = x_2, X_1 = e) = \sum_{x_3, x_4, x_5} & P(X_1 = e) \cdot P(X_2 = x_2 \mid X_1 = e) \cdot P(X_3 = x_3 \mid X_1 = e, X_2 = x_2) \cdot \\ & P(X_4 = x_4 \mid X_1 = e, X_2 = x_2, X_3 = x_3) \cdot \\ & P(X_5 = x_5 \mid X_1 = e, X_2 = x_2, X_3 = x_3, X_4 = x_4) \end{aligned} \tag{3.2}$$

Equation 3.2 is a discrete form of marginalization over a set of variables $\{X_3, X_4, X_5\}$. Rather than proceeding the marginalization over all three variables at the same time, a more efficient way is to push the summation over each variable in the set $\{X_3, X_4, X_5\}$.

$$P(X_2 = x_2, X_1 = e) = P(X_1 = e) \cdot P(X_2 = x_2 \mid X_1 = e) \cdot \sum_{x_3} P(X_3 = x_3 \mid X_1 = e, X_2 = x_2) \cdot \sum_{x_4} P(X_4 = x_4 \mid X_1 = e, X_2 = x_2, X_3 = x_3) \cdot \underbrace{\sum_{x_5} P(X_5 = x_5 \mid X_1 = e, X_2 = x_2, X_3 = x_3, X_4 = x_4)}_{\gamma_{X_5}(X_1, X_2, X_3, X_4)} \tag{3.3}$$

In Equation 3.3 variable $X_5$ only appears in one term of factors, so it can be marginalized out individually to obtain the remaining factor $\gamma_{X_5}(X_1, X_2, X_3, X_4)$. Then $X_4$ is marginalized out and we repeat the process by marginalizing one variable out at a time. This procedure is called *variable elimination* (VE) (Barber, 2012), because each time a single variable is eliminated from the joint distribution.

We can view the VE process as passing a *message* to a neighbouring node on a tree (singly connected graph). One can calculate a univariate marginal of any tree by starting at a leaf node of the tree, eliminating the variable there and then obtain a subtree of the original tree to perform next VE process. If an original BN graph is not singly connected we can group a set of nodes into a single *clique*, by which we can obtain a *join/junction tree* to perform massage passing. To convert an original BN into junction tree, involves the use of tree-decomposition operations. Tree decomposition is the basis for many well-known algorithms, i.e. junction tree clustering (JT) (Jensen, Lauritzen, & Olesen, 1990) (Shafer & Shenoy, 1990) (Lauritzen & Spiegelhalter, 1988) (Jensen & Nielsen, 2009) and cluster-tree elimination (CTE) (Kask et al, 2005). JT is an exact inference algorithm. The decomposition is achieved by using a triangulation algorithm by embedding the BN’s moral graph into a chordal graph, and using its maximal cliques as nodes in the junction tree. JT and other tree decomposition algorithms are all based on *local computation* where we compute the joint distribution by calculating the marginal for one or more clusters in the tree and then using message passing to update the other clusters in the same tree.

The JT algorithm consists of the following steps (Barber, 2012):

1. Moralization: add edges for parent nodes, this is only required for DAG.
2. Triangulation: this process produces a graph for which there exists a VE order that introduces no extra links in the graph, where every ‘square’ (loop of length 4) must have a ‘triangle’ with edge added to satisfy this criterion.
3. Junction Tree: Form a junction tree from the triangulated graph, removing any unnecessary links in a loop on the clique graph. There is exactly one path between each pair of cliques. This is called *singly connected*.
4. The resulting junction tree’s cliques are all connected by intersections/*separator*, which are common variable sets between neighbouring cliques. This is also called the *running intersection property*.
5. Factor Assignment: factors for each clique are assigned by multiplying all its associated NPTs induced by the variables contained. Separators are assigned by uniform factors.
6. Message Propagation: perform absorption (sending messages to neighbours) and distribution operations (receiving messages from neighbours) to pass update messages to all cliques and separators in the junction tree. The resulting marginal for each variable in different cliques are now consistent.

The efficiency of the JT architecture depends on the size of the cliques in the associated tree and the inference for both exact and approximate is NP-hard (Cooper & Herskovits, 1992). Although JT based algorithms are intended to produce junction trees with minimum clusters size for many models the complexity of JT algorithm grows exponentially with clique size as the number of discrete variables and states increases, thus making the computation of the marginal distribution very costly or even impossible.

Consider the full-BFG $G'$ in Figure 2.7, and compare it with the original BN, $G$. The largest factor size of a single NPT is reduced from $m^5$ to $m^3$, where $m$ is the number of partitions. However, the largest clique size is $m^5$ because of the triangulation performed during JT construction. The BF procedure has only reduced the computation complexity for the NPTs but not of the cliques, i.e. to compute the marginal for $X_1$ from the joint distribution the minimal clique (containing variables $X_1$, $X_2$, $X_3$, $E_1$ and $E_2$) size is still five. Besides that, to find the most efficient and optimal triangulation is NP-hard, thus an alternative clustering algorithm is required. Section 3.4 introduces Generalized Belief Propagation (GBP), which offers the potential to be a credible alternative.

We will use the VE process in Chapter 4 for developing BFE algorithm. The JT algorithm will be discussed again in the context of dynamic discretization (Neil, Tailor, & Marquez, 2007) in Section 3.3.

# 3.2. Approximate Inference

The most well-known approaches to approximate inference in BNs include Markov Chain Monte Carlo (MCMC) samplers, Variational Inference (VI), Dynamic Discretization with Junction Tree (DDJT), (Generalized) Belief Propagation (G/BP) and its variants, and BP-based continuous domain algorithms: expectation propagation (EP), Non-parametric Belief Propagation (NBP), Particle Belief Propagation (PBP). This section briefly reviews MCMC, VI, DDJT, BP/GBP, EP, NBP and PBP. GBP will be discussed in further in section 3.4.

1. **MCMC**: MCMC is a stochastic simulation based method. There are a large number of specifically designed MCMC samplers (Gamerman & Lopes, 2006) that are widely applied to Bayesian inference. MCMC samples from probability distributions based on constructing a Markov chain that has the desired distribution as its equilibrium distribution and then the target distribution is obtained from the sample states of the chain. MCMC is not restricted in the number of model variables and so scales up very well, but there are still some open problems to be solved when using MCMC. For example, handling multimodal distributions[2] is difficult and guaranteeing convergence for arbitrary models can be a major problem. When designing an MCMC algorithm there is always a balance to be found between exploiting information to adjust the parameters and searching for new regions of the sample space (Ceperley et al., 2012). When the models are complex and hybrid the necessary MCMC algorithms have to be specially tailored, thus involving significant labour and testing.

[2] For example, multi-modal distributions with multiple peaks and narrow variance around each peak are difficult to handle using MCMC.

2. **VI**: Analytical approximation methods have been developed to achieve much higher computational efficiency compared to simulation based approaches such as MCMC. The mainstream methods include 'variational' approaches (Beal, 2003) and the mean field algorithm [17], a simplified variational algorithm (Jordan, et al, 1998), which involves choosing a family of distributions over the latent (as opposed to observable) variables with their own variational parameters. Then, it searches for the parameters settings that make the chosen parameter distributions closer to the true posterior of interest. Recent work on variational inference (Wang & Blei, 2013), has explored a generic algorithm for non-conjugate models, involving extending variational inference to a broader range of models, but despite this the analytical effort can be prohibitive.

3. **DDJT**: The Dynamically Discretized Junction Tree (DDJT) algorithm (Neil et al., 2007) is a combination of the JT algorithm and Dynamic Discretization (DD), where JT performs inference over a set of discrete variables and DD transforms the continuous variables in the model using discrete approximations. The advantage of using DDJT over the other approaches is that it can easily perform discrete inference on a hybrid BN. The approximation error is only introduced by discretization and DD performance is much superior to static discretization. DDJT performance is robust and accurate; it has been applied in numerous domains: systems reliability modelling (Neil, et al, 2008) (Marquez, Neil, & Fenton, 2010) , operational risk modelling in finance (Neil & Fenton, 2008) and others (Fenton & Neil, 2010) (Neil & Marquez, 2012) (Fenton & Neil, 2012). However, the curse of dimensionality is a major barrier in DDJT since the Node Probability Tables (NPTs) needed in JT is exponential to the number of parent node states. Likewise, the triangulation operation in the JT algorithm means that the cluster sizes (the basic data structures in JT) also grow exponentially.

4. **BP**: Belief Propagation (BP) (Pearl, 1988) (Kschischang, Frey, & Loeliger, 2001), also known as sum-product message passing, is a message passing algorithm for performing inference on undirected joint tree/graph models. Unlike most sampling schemes, BP does not suffer from high variance and is often much more efficient (Welling, 2004). When the graphical model represents a BN whose corresponding factor graph[3] forms a tree, the inference is equivalent to the exact inferences obtained by the JT algorithm (Yedidia et al., 2005). If the factor graph contains loops it is an approximation. Performing BP iteratively on networks that contain cycles yields loopy belief propagation (LBP, also called iterative BP) (Murphy, Weiss, & Jordan, 1999). However the convergence of LBP is not guaranteed, although LBP has achieved success on coding networks (Mateescu et al, 2010). Survey Propagation (SP) is a version of LBP, which is an improvement of BP and has been successfully applied to NP-complete problems like satisfiability (Braunstein, Mézard, & Zecchina, 2005), but SP does not guarantee convergence either. Exact inference, i.e. JT, maybe introducing large clusters during belief updating, Mini-Clustering belief updating (MC-BU) algorithms (Kask, 2001) (Mateescu et al, 2002) (Dechter & Rish, 2003) partition large clusters into bounded (user specified) mini-clusters, and use max/mean operators in all of its mini-clusters to derive a strict upper bound of the joint belief. The performance of MC-BU algorithm is still limited on specific networks (Mateescu et al., 2010). The Iterative Join Graph Propagation (IJGP) algorithm (Mateescu et al., 2010) combines the idea of mini-clustering and iterative BP and produce a join graph decomposition based on bounded mini-cluster size. BP is carried out on the resulting cyclic join graph. IJGP introduces cycles in the join graph, so it also uses edge separation algorithms to avoid over counting each variable. It enables trade-offs between accuracy and complexity based on user defined mini-cluster size. Mateescu (Mateescu et al., 2010) shows that IJGP surpasses other join graph based BP algorithms, i.e. Edge Deletion Belief Propagation (EDBP) (Choi & Darwiche, 2006), Truncated Loop Series Belief Propagation (TLSBP) (Gómez, Mooij, & Kappen, 2007), and surpasses or is equal to iterative BP and MC-BU for specific networks.

[3] A factor graph is a bipartite graph representing the factorization of a function. So in the case of a BN the function is the associated joint probability distribution function.

5. **EP, NBP and PBP**: Expectation propagation (EP) (T. P. Minka, 2001), non-parametric belief propagation (NBP) (Sudderth, Ihler, Freeman, & Willsky, 2003) and particle belief propagation (PBP) (Ihler & McAllester, 2009), characterize messages in the continuous domain, where additional approximations are developed using BP decompositions. EP uses an iterative approach that leverages the factorization structure of the target distribution $p$, where large joint factors are factorized into more compact factors, where the resulting joint distribution $\tilde{p}$ is tractable and one sets free parameters that minimize the Kullback-Leibler distance, $KL(p \| \tilde{p})$ (Barber, 2012). EP is limited to exponential families. NBP and PBP are both MCMC sampling based algorithms. NBP uses Gaussian mixtures to represent BP messages and it needs to smooth the sample set when taking products of messages, which is a further approximation. PBP sidesteps the shared collection of samples used in NBP by using separate message sampling strategies, which shown improved performance over NBP (Ihler & McAllester, 2009). Because they are both sampling based algorithms, performance is sensitive to the choice of proposed sampling distribution and sampling strategy and, in practice, the iterative message sampling procedure for NBP and PBP can make it difficult to achieve convergence (Ihler & McAllester, 2009).

6. **GBP**: Generalized Belief Propagation (GBP) (Yedidia et al., 2005), is also an improvement on BP and provides the generalized form for all message passing algorithms. GBP perform message passing on a region graph, which is a directed graph produced using flexible clustering schemes. There are connections between the region graph and join graph formalisms (Mateescu et al., 2010); if the edge directions in a region graph are removed it can be viewed as join graph with a particular clustering scheme. GBP is best explained with the context of statistical physics, where the simplest version of GBP is called Bethe approximation. A more general version is called Kikuchi (Kikuchi, 1951) approximation, where variables are grouped into clusters. GBP algorithms are applications of Kikuchi approximation. Yedidia (Yedidia et al., 2005) shows that GBP is flexible and can achieve accurate results provided that it converges. The most widely applied GBP algorithm would be Cluster Variation Method (CVM). There are numerous lines of GBP based algorithms, i.e. join graph based algorithms (i.e. IJGP) can be converted to region graph based algorithms (i.e. CVM). However, the GBP algorithm has rarely been applied to factorized models, the exception being some research on its application to continuous Gaussian linear models (Bickson et al., 2008) (Shental et al., 2008), where the emphasis is on analytical solutions to the Gaussian case only. In practice general purpose inference using GBP has to use heuristics to convert a factor graph into a region graph, where a good construction of such a region graph can be difficult to find and is non-generalizable.

## 3.3. Dynamically Discretized Inference Algorithm

This section introduces dynamic discretization (Neil et al., 2007), alongside the Junction Tree algorithm, to perform approximate inference on hybrid models. Different with conventional approximation inferences, dynamic discretized inference algorithm uses exact inference JT algorithm. The approximation error is introduced by discretization, which is more accurate than static discretization methods.

### 3.3.1 Dynamic Discretization

If continuous variables are represented by mixtures of constant uniform distributions, the overall BN variables can be represented universally using only discrete variables. Conventional static discretization has historically been used to approximate the domain of the continuous variables in a BN using predefined, fixed piecewise constant partitions. This approximation will be valid when the posterior high density region remains in the specified domain during inference. However the analyst has no advanced knowledge about which areas of the domain require the greater number of intervals thus resulting in an inaccurate posterior estimate. Dynamic Discretization (DD) is an alternative discretization approach that searches for the high density region during inference. It dynamically adds more intervals where they are needed, and removes intervals where they are not (by merging or deletion). The algorithm iteratively discretizes the target variables by the convergence of relative entropy error threshold (Kozlov & Koller, 1997).

Kozlov (Kozlov & Koller, 1997) proposed a non-uniform dynamic discretization algorithm for hybrid networks. The discretization of continuous variables is done at the level of join tree cliques, where they use a recursive binary split partition (BSP) tree data structure to decompose the multidimensional hierarchical space generated by the clique. Leaves of the BSP tree store the average constant value of the function over a sub-region. The discretization interleaves the join tree propagation algorithm, and iteratively improves accuracy by calculating the weighted KL distance between multivariate distributions and splitting regions that improve this. The approach is accurate since it is performed over multidimensional cliques directly, rather than subsets of them, but it can be slow because the number of samples needed to determine where the next optimal split should take place is determined by the size of the clique. Likewise, implementing the algorithm is challenging since the algorithms for manipulating BSP tree are more complex than those needed for plain junction trees. Compared to Kozlov's approach, DD discretizes each variable separately based on the marginal distribution and uses the standard junction tree message passing algorithm. This is less accurate than Kozlov's approach, demands more space, but converges to the same result and is faster.

In addition to the above splitting strategies for representing CPD tables, Friedmen (Friedman & Goldszmidt, 1996) describes a method that uses local structures in CPD tables during the learning process and this can reduce the tabular size for learning. The local structures used employ knowledge about context specific independence, described by either asymmetrical tables or decision trees (Friedman & Goldszmidt, 1996). This structure reduces the number of parameters required in CPDs. In contrast with DD, Friedman's approach limited to discrete case only and is rarely implemented.

In outline the DD algorithm is introduced below:

Suppose $X$ is a continuous random node in a BN. The range of $X$ is denoted by $\Omega_X$, and the probability density function of $X$, with support $\Omega_X$, is denoted by $f_X$. The idea of discretization is to approximate $f_X$ as follows:

1. Partition $\Omega_X$ into a set of interval $\Psi_X = \{w_j\}$,
2. Define a locally constant function $f_x$ on the partitioning intervals.

We estimate the relative entropy error induced by the discretized function using an upper bound of the KL metric between two density functions $f$ and $g$:

$$D(f \parallel g) = \int f(x) \log \frac{f(x)}{g(x)} dx \tag{3.4}$$

Under the KL metric the optimal value for the discretized function $\tilde{f}_x$ is given by the mean of the function $f_x$ in each of the intervals of the discretized domain. The discretization task reduces then to finding an optimal partition set $\hat{\Psi}_x$.

DD searches $\Omega_X$ for the most accurate specification of the high-density regions given the model and the evidence, calculating a sequence of discretization intervals in $\Omega_X$ iteratively. At each stage in the iterative process, a candidate discretization, $\Psi_x = \{w_j\}$, is tested to determine whether the relative entropy error of the resulting discretized probability density $\tilde{f}_x$ is below a given threshold, defined according to some stopping rule. After each variable in the model is discretized, the underlying inference algorithm, e.g. Junction Tree, is performed to calculate the joint posterior distributions for all variables in the model. This produces a new posterior probability density for all variables and these are then re-discretized in the next iteration. This process continues until a stopping rule is triggered, i.e. where the KL distance reaches some target.

### 3.3.2 Dynamically Discretized Junction Tree Algorithm

DD offers the potential to be constructed with any inference engine, e.g. JT algorithm, or other types of inference algorithms. The discretization strategy is determined by the relative entropy error resulted from inference, and then new factors can be generated for NPT tables in order to perform the next propagation. The NPT generation is determined by discretized partitions and deterministic or statistical functions, where the child and its parents' partitions are combined to form an NPT, either using a set of uniform mixtures, in the case of conditionally deterministic functions, or directly from a statistical distribution.

For example for a conditionally deterministic distribution, $X = Y + Z$, the NPT for $p(X \mid Y, Z)$ can be determined by (Neil et al., 2007):

$$p(X \mid Y \in [y_u, y_v], Z \in [z_u, z_v]) \\ = Uniform(\min(y_u z_u, y_u z_v, y_v z_u, y_v z_v), \max(y_u z_u, y_u z_v, y_v z_u, y_v z_v)) \tag{3.5}$$

The new generated NPT is converted to factors for the JT tree to perform next iteration. We summarize the dynamic discretized junction tree (DDJT) algorithm as (Neil et al., 2007):

***Algorithm 1*** *Dynamic Discretized Junction Tree (DDJT)*

**Input**: original BN $G$

**Output**: original BN $G$ with marginals

1. **Build** a Junction Tree to determine the cliques, $\mathbf{R}$, and separators
2. **Assign** the clique's potentials/factors $\varphi(R_k)$ by multiplying the NPTs/factors $P(X \mid pa\{X\})$ for all variables into the relevant cliques
3. **Assign** uniform potentials/factors for all separators
4. **Initialize** each clique's discretization $\mathbf{\Psi}_k^{(0)}$, by its support $\mathbf{\Omega}_k$
5. **Do Compute** the approximate NPTs, $P^{(0)}(X \mid pa\{X\})$, on $\Psi_X^{(l-1)}$ for all variables $X$ and initialize the clique's factors
6. **Query** the BN from node $\mathbf{X}_E = \mathbf{e}$
7. **for** $l = 1$ to $\text{max_num_ite}$ **repeat**
8. **Create** a new discretization $\mathbf{\Psi}_k^{(l)}$ for clique domain $\mathbf{\Omega}_k$
9. **Perform** *absorption and distribution of messages* on the JT
10. **Compute** the new discretized potential $\varphi^{(l-1)}(R_k)$
11. **Compute** the approximate total relative entropy error
12. **end for**
13. **until** convergence of the posterior marginal for each clique by stable entropy error stopping rule and low entropy error stopping rule
14. **Extract** marginal for each node from the relevant clique
15. **return** $G$

Algorithm 1 DDJT algorithm

**Example 3.1** Convolution $X = Y + Z$

Consider the example BN model in Figure 3.1

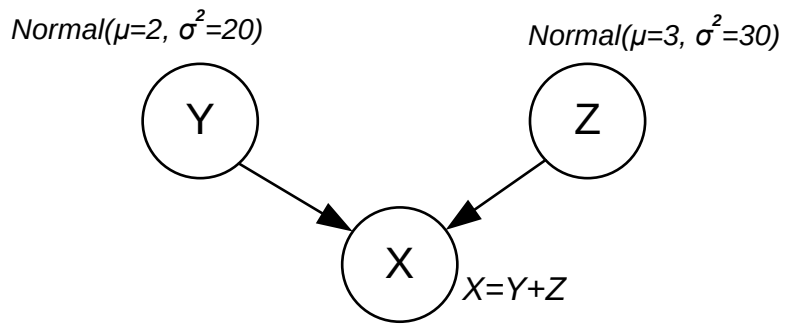

Figure 3.1 BN model where $X$ is conditionally deterministic sum of normal distribution parents

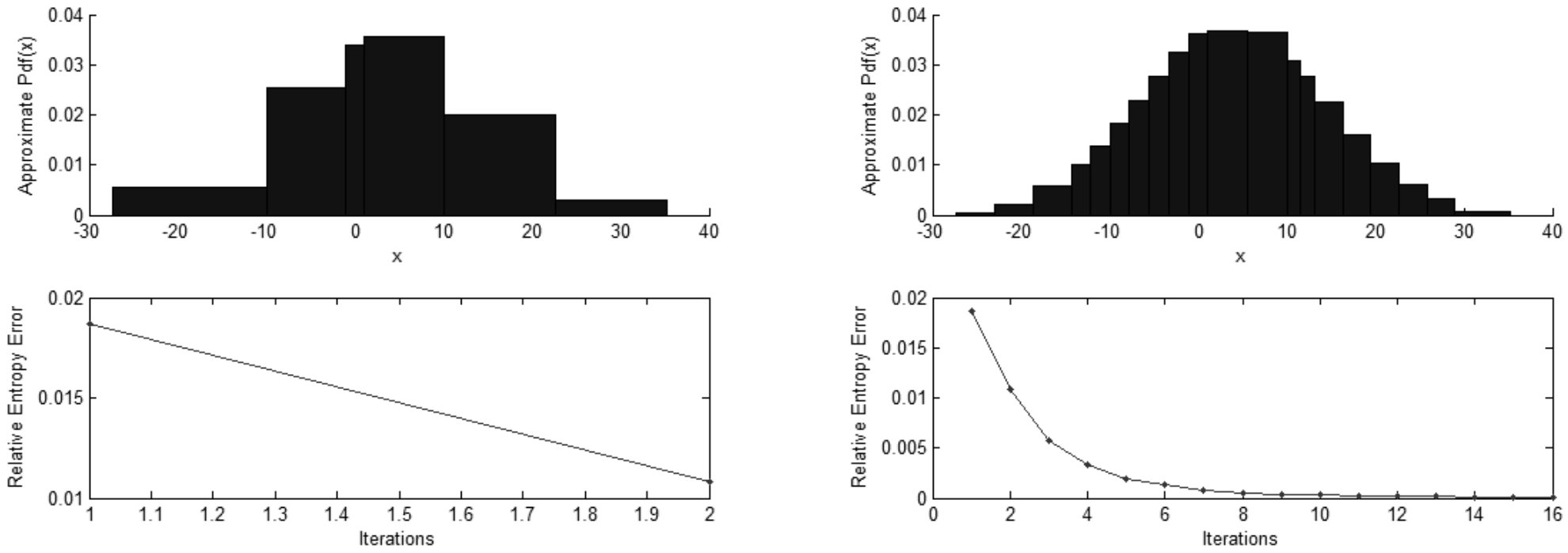

Figure 3.2 DDJT approximation of posterior marginal with relative entropy error convergence plot

Figure 3.2 illustrates the results from applying the DDJT iteratively to approximate the posterior marginal of $X$ . Figure 3.2 left shows the result after two iterations with a relative entropy error for $X$ of approximately 0.01, and after 16 iterations (Figure 3.2 right) the model converges to an acceptable marginal distribution with a relative entropy error for $X$ of 0.001. After 25 iterations the marginal distribution of $X$ is approximated with a mean 5 and variance 51.

This thesis discusses the DDBP algorithm in chapter 5, where the JT inference engine is swapped out and replaced by the GBP inference engine.

## 3.4. Generalized Belief Propagation

Generalized Belief Propagation (GBP) is a generalization of all message passing algorithms, including JT, and allows flexible clustering, where cluster size can be adjusted. In all standard BP algorithms (Kschischang et al., 2001), messages are sent from one node to a neighbour node in a factor graph. Propagation is exact when the factor graph has no cycles, but for factor graph containing cycles we can only perform approximate propagation. Although the BP algorithms are well defined when factor graph have cycles, convergence is sometimes not achieved (Yedidia et al., 2005).

(Yedidia et al., 2005) generalized the BP algorithm and demonstrated that BP convergence is equivalent to stationary points of the Bethe approximation of the free energy of a factor graph. They demonstrated that the Generalized BP (GBP) algorithm obtained from the region based free energy approximation, improved the Bethe approximation and achieved better accuracy than ordinary BP. These gains are achieved by constructing a region graph (with factor graph nodes being clustered) as an alternative to a factor graph. The region based free energy corresponds to the difference of variational average energy of region beliefs and region entropies and when the region beliefs are the same as the joint probability distribution the free energy is minimized. Furthermore, they also demonstrated that a valid construction of the corresponding region graph can be specified by the Bethe Method (BM), Junction Graph Method (JGM) and the Cluster Variation Method (CVM). As we have noted in section 3.2, the join graph based algorithm, i.e. IJGP can be converted to CVM and vice versa. These connections provide some justification for the development of the TRC algorithm in this thesis, as described in Chapter 5.

### 3.4.1 Converting a BN to a Markov Network

GBP is defined on an undirected graph and this requires the conversion of a BN to a *Markov Network* (MN). For a set of variables $\mathbf{X} = \{X_1, ..., X_n\}$, a MN is an undirected graph $G$, and is defined as a product of factors, $\phi_c(\mathbf{X}_c)$, on subsets of the variables $\mathbf{X}_c \subseteq \mathbf{X}$:

$$p(X_1, ..., X_n) = \frac{1}{Z} \prod_{c=1}^{C} \phi_c(\mathbf{X}_c), \tag{3.6}$$

where $c = 1, ..., C$ are the maximal clusters of $G$, and where $Z$ is a normalization constant.

The conversion from BN to MN involves two steps:

1. Convert BN parameterization to MN parameterization:

$$p(X_i \mid pa\{X_i\}) = \phi_{\{X_i\} \cup pa\{X_i\}}(X_i, pa\{X_i\}).$$

2. Connect all parent nodes that have the same child node, and convert all directed edges into undirected edges.

The converted MN graph is also called a *moral graph*, since the parents are connected.

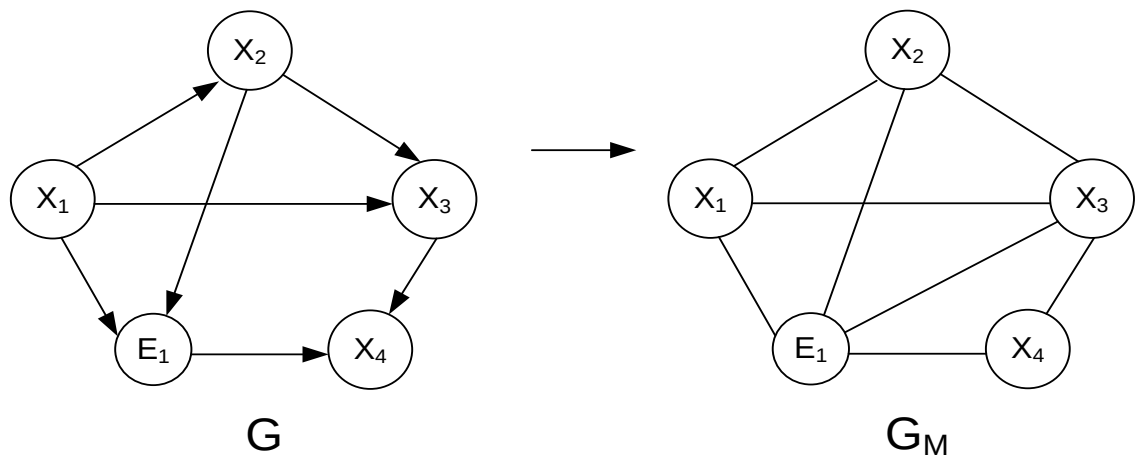

Figure 3.3 Convert a $\kappa_4$ full BFG $G$ to moral graph $G_M$

Figure 3.3 shows the results of a conversion of a BN $G$ to its corresponding moral graph. The CPD $p(X_i \mid pa\{X_i\})$ in $G$ for each variable is re-parameterized to $\phi_{\{X_i\} \cup pa\{X_i\}}(X_i, pa\{X_i\})$.

## 3.4.2 Factor Graphs

Both BNs and MNs can be represented by a unifying representation called a Factor Graph (FG) (Kschischang et al., 2001). FGs explicitly express the factorization structure of the corresponding probability distribution. Standard BP performs message passing on factor graphs.

An FG is a particular type of graphical model with applications in Bayesian inference that enables efficient computation of marginal distributions through the sum-product algorithm[4] (Koller & Friedman, 2009) , (Kschischang et al., 2001). It is a bipartite graph representing the factorization of a function. Given a factorization of a function $g(X_1, X_2, ..., X_n) = \prod_{j=1}^{m} f_j(S_j)$ , where $S_j \subseteq \{X_1, X_2, ..., X_n\}$ , the corresponding factor graph $G = (X, F, E)$ consists of variable vertices

[4] The sum-product algorithm defined as to compute marginal we need to distribute the sum over variable states over the product of factors.

$X = \{X_1, X_2, ..., X_n\}$, factor vertices $F = \{f_1, f_2, ..., f_m\}$, and edges $E$. The edges depend on the factorization as follows: there is an undirected edge between factor vertex $f_j$ and variable vertex $X_k$ when $X_k \in S_j$. The function is tacitly assumed to be real valued. FGs can be combined with message passing algorithms to efficiently compute certain characteristic of the function $g(X_1, X_2, ..., X_n)$, such as the marginal distribution.

In Figure 2.5 $G'$, the joint distribution of BN representation is factorized by the FG in Figure 3.4, where $\{X_1, X_2, X_3, E_1, X_4\}$ are variable nodes, $\{f_1X_1, ..., f_5X_3E_1X_4\}$ are factor nodes.

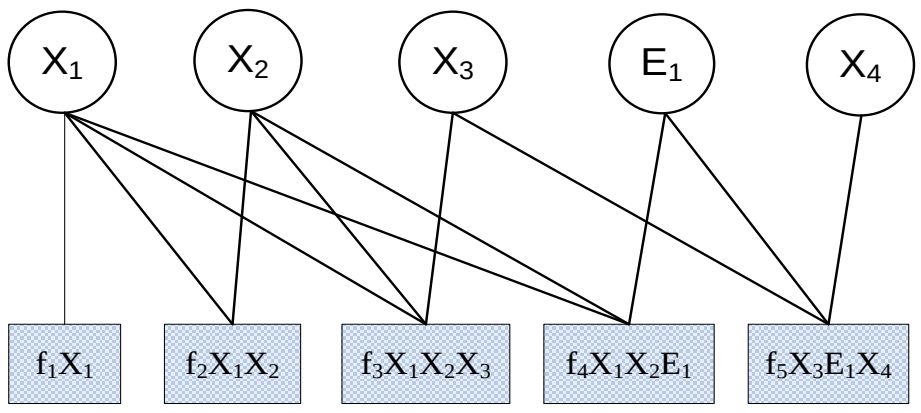

Figure 3.4 FG representation for the joint distribution:

$$P(X_1, X_2, X_3, E_1, X_4) = \frac{1}{Z} f_1(X_1) f_2(X_1, X_2) f_3(X_1, X_3, X_4) f_4(X_1, X_2, E_1) f_5(X_3, E_1, X_4)$$

### 3.4.3 Belief Propagation on Factor Graph

For a given FG with variables $X_1, ..., X_N$, the joint probability mass function is

$$P(X_1 = x_1, ..., X_N = x_N) = p(\mathbf{x}) = \frac{1}{Z} \prod_a f_a(\mathbf{x}_a) \tag{3.7}$$

Where $\mathbf{x}$ is he set $\{x_1, ..., x_N\}$. Generally we are interested in computation of estimating marginal $p_S(\mathbf{x}_S) = \sum_{\mathbf{x} \backslash \mathbf{x}_S} p(\mathbf{x})$.

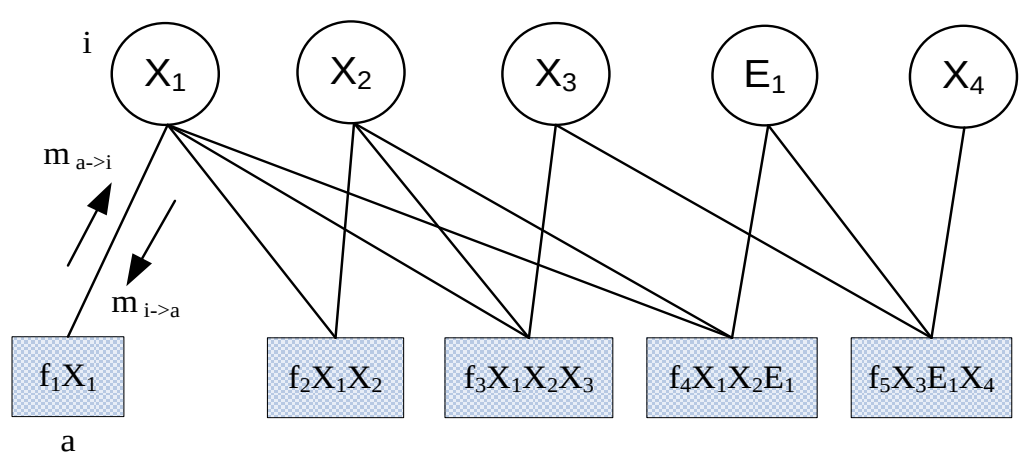

Figure 3.5 Messages passing in the FG in Figure 3.4

Depicted in Figure 3.5, the message passing in a FG involves two steps:

1. Messages $m_{a\to i}(x_i)$ from factors to variables: what values does factor $a$ like variable $X_i$ to take on.
2. Messages $m_{i\to a}(x_i)$ from variables to factors: what values $X_i$ likes based on information from all but $a$.

Refer to information theory, we define variable and factor nodes' *Beliefs* (Yedidia et al., 2005):

1. $b_i(x_i) \propto \prod_{a\in N(i)} m_{a\to i}(x_i)$, are based on all pieces of information coming into $X_i$, the product is independent pieces of information.
2. $b_a(\mathbf{x}_a) \propto f_a(\mathbf{x}_a) \prod_{i\in N(a)} n_{i\to a}(x_i)$, are based on product of local factors and messages coming from variables.

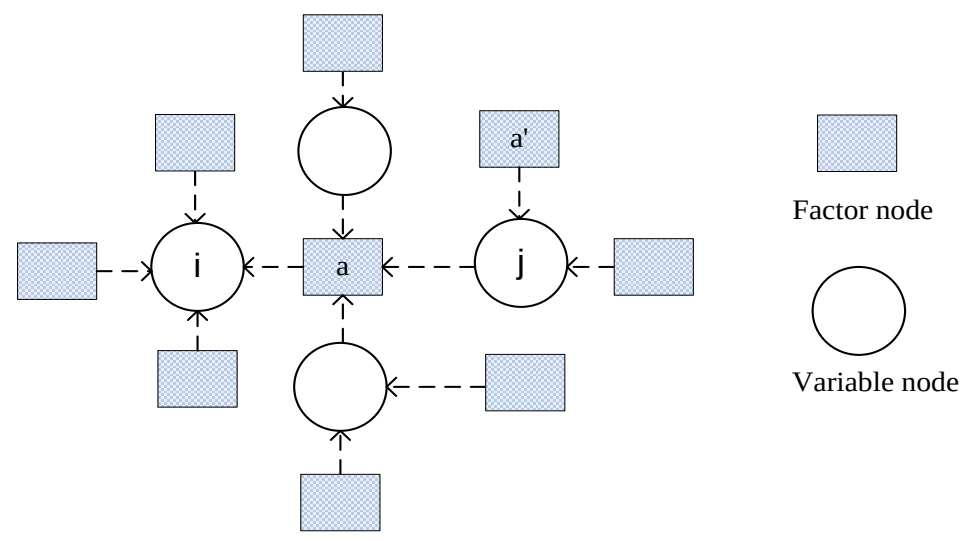

Figure 3.6 Message updating in a FG with dashed lines are message passing directions

The message updating rule is defined in the following three steps and illustrated in Figure 3.6.

1. The belief of every variable is $b_i(x_i) = \sum_{\mathbf{x_a}\backslash x_i} b_a(\mathbf{x}_a)$,
2. The product of all the messages coming from variable $i$'s neighbour factor nodes that are used for calculating the local beliefs:

$$\prod_{a\in N(i)} m_{a\to i}(x_i) = \sum_{\mathbf{x}_a \backslash x_i} f_a(\mathbf{x}_a) \prod_{j\in N(a)} n_{j\to a}(x_j) = \sum_{\mathbf{x}_a \backslash x_i} f_a(\mathbf{x}_a) \prod_{j\in N(a)} \prod_{a'\in N(j)\backslash a} m_{a'\to j}(x_j)$$
$$= \prod_{a'\in N(i)\backslash a} m_{a'\to i}(x_i) \sum_{\mathbf{x}_a \backslash x_i} f_a(\mathbf{x}_a) \prod_{j\in N(a)\backslash i} m_{a'\to j}(x_j)$$

(3.8)

3. Messages for each factor to variable nodes are computed by marginalization constraint:

$$m_{a\to i}(x_i) = \sum_{\mathbf{x}_a \backslash x_i} f_a(\mathbf{x}_a) \prod_{j\in N(a)\backslash i} n_{j\to a}(x_j) \tag{3.9}$$

When BP is performed on the factor graph messages are sent between factor nodes and variable nodes. The BP convergence is equivalent to stationary points of the Bethe approximation (Yedidia et al., 2005) of the free energy of this FG.

### 3.4.4 Region Free Energy and Region Graph

In all standard BP algorithms, messages are sent from one node to a neighbour node in a graphic representation. Propagation is exact when the graphic model has no cycles, but for models with cycles we can only perform propagation in an approximate manner.

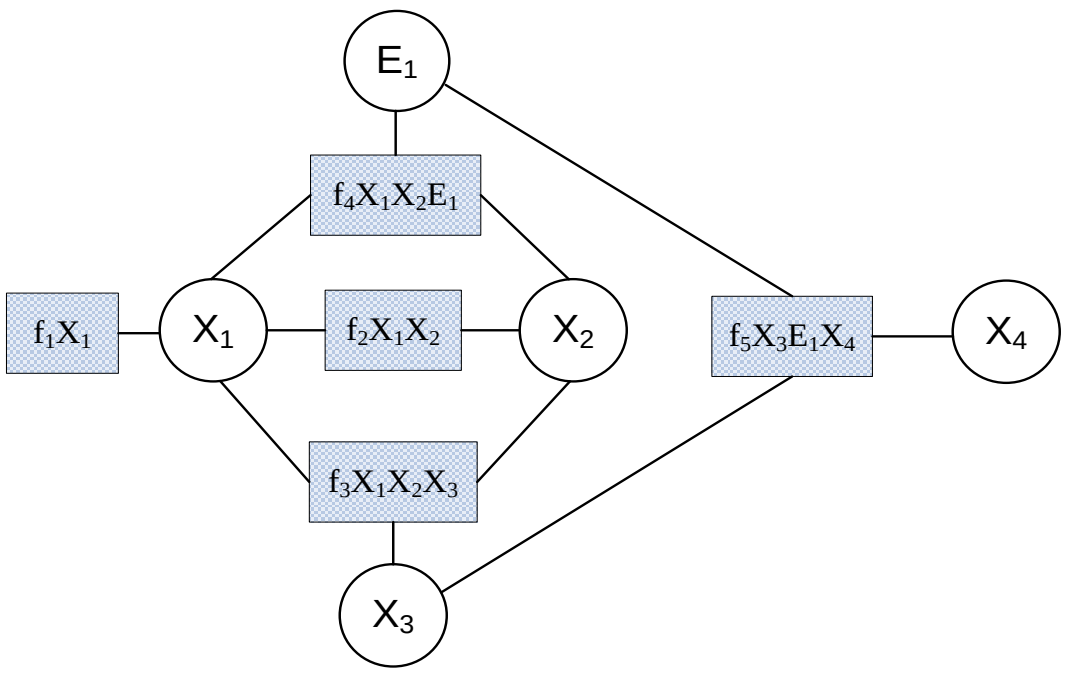

Figure 3.7 Factor graph containing cycles

Figure 3.7 is identical to Figure 3.5 but illustrates the cycles contained in the factor graph. The standard BP performance on this graph is an approximation.

Yedidia, et al (Yedidia et al., 2005) generalized the BP algorithm and demonstrated that BP convergence is equivalent to constructing the free energy of a system. Rather than propagate messages among nodes in a FG GBP operates on a Region Graph (RG) which is a graphical formalism for generating free energy approximations.

There are several ways to define the regions in a graph to support message exchange. Yedidia, et al has shown that a valid construction of the corresponding RG can be specified by several clustering algorithms: Bethe Method (BM), Junction Graph Method (JGM) and the Cluster Variation Method (CVM).

A RG is constructed as follows. Let $I$ be the set of indices for the factor and variable nodes in a factor graph. A RG is a labelled, directed graph $G=(V,E,L)$ in which each vertex $v \in V$ (corresponding to a region) is labelled with a subset of $I$. We denote the label of vertex $v$ by $l(v) \in L$. A directed edge $e \in E$ may exist directed from vertex $v_p$ to vertex $v_c$ if $l(v_c)$ is a subset of $l(v_p)$. If such an edge exists, we say that $v_c$ is a *child* of $v_p$, that $v_p$ is a *parent* of $v_c$. If there exists a directed path from vertex $v_a$ to vertex $v_d$, we say that $v_a$ is an *ancestor* of $v_d$, and $v_d$ is a *descendant* of $v_a$. For a graph $G$ to qualify as a RG, we require the following *region graph constraints* (Yedidia et al., 2005):

1. Regions can be organized as a directed acyclic graph, $R_1 \to R_2$ only if $R_2 \in R_1$, where all factors $a$ and all variables $i$ are included.
2. The marginalization constraint for region beliefs: $\sum_{\forall x_{R_1} \notin x_{R_2}} b_{R_1}(x_{R_1}) = b_{R_2}(x_{R_2})$ (where $b_{R_2}(x_{R_2})$ are probabilities/beliefs).
3. The subgraph of all regions containing each factor, $a$, or variable, $i$, is connected so that the region graph gives consistent beliefs about them.
4. $c_R = 1 - \sum_{ancestors\ R' \in A(R)} c_{R'}$, where $c_{R'}$ is degree of freedom (number of parent regions) and $c_R$ corresponds to counting number of each region.
5. For every $i \in I$ (whether it is the index of a factor node or a variable node), the sub graph $G(i)=(V(i),E(i),L(i))$ formed by just those vertices whose labels include $i$ is a connected graph that satisfied the condition: $\sum_{v \in V(i)} c_v = 1$.

Maintaining the counting number at one is important to ensure that variables are not under or over counted during inference. Note that BP produces exact results if the resulting region graph forms a tree and satisfies all of the constraints above.

The BM is always an exemplar of the JGM and is only a special case of the CVM if the factor graph does not contain any pairs of factor nodes that share more than one variable node. CVM is more flexible and easy to use than other methods and we will use CVM as reference algorithm for the TRC algorithm in later chapters. Informally, the steps in the CVM algorithm are (see Example 3.2):

1. Define the first level regions, $R_0$, such that every factor node, $a$, and every variable node, $i$, in our factor graph is included in at least one region $R \in R_0$. There must be no region $R \in R_0$ that is a sub-region of other regions in $R_0$.
2. Construct second level regions $R_1$ by forming all possible intersections between regions in $R_0$, but discard from $R_1$ any region that is a sub-region of other regions in $R_1$.
3. If possible, repeat step 2 for $R_0 \cup R_1$ to form $R_2$, resulting in a final set of regions, $\mathbf{R} = R_0 \cup R_1 \cup \ldots R_k$.

Example 3.2 shows the construction of a region graph using CVM.

**Example 3.2**

Consider the original BN in Figure 3.8, $G$. The conversion of this BN to its moral graph involves changing all directed edges to undirected edges and adding an undirected edge between each pair of parents (dashed lines in $G_M$, Figure 3.8). All CPDs in $G$ then need to be re-parameterized to factors in $G_M$. For example, the CPD $p(E_1 | X_1, X_2)$ is re-parameterized to $\phi(E_1, X_1, X_2)$, and other smaller factors like $\phi(X_1)$, $\phi(X_2)$ can be multiplied together to give $\phi(E_1, X_1, X_2)$. The resulting factors are then $\phi(E_1, X_1, X_2)$, $\phi(E_2, X_3, X_4)$ and $\phi(E_1, E_2, X_5)$. we can construct the region graph based on its moral graph $G_M$, and use these factors we obtained above. These are shown in Figure 3.8.

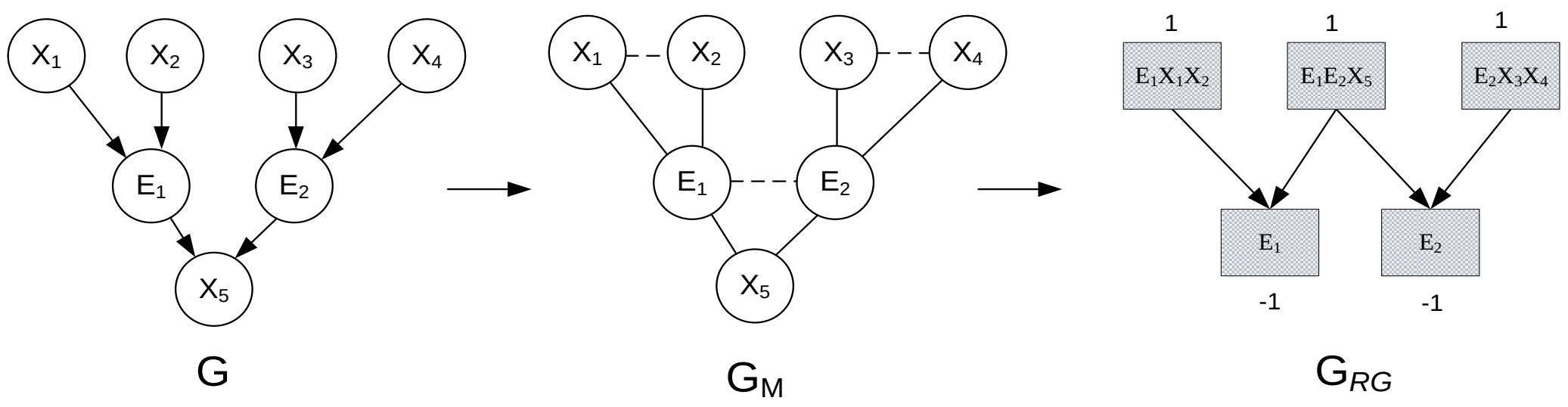

Figure 3.8 Generate region, $G_{RG}$, graph for a BN, $G$

Figure 3.8 illustrates the RG using the CVM construction method. The CVM algorithm requires that all of the factors and variables appear in the first level regions. The first level regions derived from $G_M$ are: $E_1X_1X_2$, $E_2X_3X_4$ and $E_1E_2X_5$. The resulting region graph, also show in Figure 3.8, is $G_{RG}$, which is a two-level acyclic graph. Moreover, it is singly connected[5] so it is equivalent to a JT.

If the region graph contains cycles, the approximation is not guaranteed and sometimes it can fail to converge at all, so it is preferable to construct a region graph that is a tree.

The choice of which region graph to use with GBP is an open research question, since it is not well understood which region graph topologies result in good approximations and which do not. Welling et al. (Welling, Minka, & Teh, 2005) (Welling, 2004) (Gelfand & Welling, 2012) discuss ways to structure region graphs based on graphic topology and offer guidance based on structural information criteria, a sequential approach where new regions are added bottom-up to the region graph, and tree-robustness. In practice, most region design is guided by constructing good approximations to the free energy of the problem.

There is very little research on how to ensure the *interaction strength* (refers to what and how many regions should be chosen) between regions is approximated, in terms of maintaining the interdependencies of the joint distribution; instead most research concerns structural information about the graph. Welling (Welling, 2004) has mentioned the impact of interaction strength by adding extra candidate regions to the existing region graph, where approximation can be improved by testing and choosing additional external regions from a candidate pool. We will discuss the region interaction strength in chapter 5 when develop the TRC algorithm. Below we briefly discuss one basic criterion, called Maxent-Normal constraint with reference (Yedidia et al., 2005) to free energy in statistical physics, to evaluate an already built region graph, shown below.

[5] There is only one path between pair of clusters.

In the Section 3.4.3 in Equation 3.7 we have shown that the probability for state $\mathbf{x}$ is $p(\mathbf{x}) = \frac{1}{Z}\prod_a f_a(\mathbf{x}_a)$ . An *energy* term in statistical physics is defined as: $E(\mathbf{x}) = -\sum_a \log f_a(\mathbf{x}_a)$ . So by substituting $E(\mathbf{x})$ we produce Boltzmann's Law, $p(\mathbf{x}) = \frac{1}{Z(T)} e^{-E(\mathbf{x})/T}$, where $Z(T) = \sum_{\mathbf{x}\in S} e^{-E(\mathbf{x})/T}$ is a *partition function* with $S$ , the space of all possible states $\mathbf{x}$ of the system and the temperature, $T$ (Boltzmann, 1884).

Using Gibbs free energy (Yedidia et al., 2005):

$$F(b) = \sum_{\mathbf{x}\in S} b(\mathbf{x})E(\mathbf{x}) - H(b) = U(b) - H(b) \tag{3.10}$$

where $U(b)$ is the *variational average energy*, and $H(b) = -\sum_{\mathbf{x}\in S} b(\mathbf{x})\log b(\mathbf{x})$ the *variational entropy*. The belief $b(\mathbf{x})$ is an estimate probability and we need to attain $b(\mathbf{x}) = \frac{1}{Z(T)} e^{-E(\mathbf{x})/T}$ when $b(\mathbf{x}) = p(\mathbf{x})$, and substitute $H(b) = -\sum_{\mathbf{x}\in S} b(\mathbf{x})\log b(\mathbf{x})$ we get $F(b) = -\log Z(T) = F_H$ , where $F_H$ is the Helmholtz free energy that can be used to recover our target function $Z(T)$ to calculate $p(\mathbf{x})$. Thus we obtain the following:

$$F(b) = F_H + KL(b \,||\, p) \tag{3.11}$$

Since $KL(b \,||\, p)$ is non-negative, $F(b) \geq F_H$ , and with $F(b) = F_H$ only when $b(\mathbf{x}) = p(\mathbf{x})$. So the task is to minimize the free energy $F(b)$. When minimising $F(b)$ computation of $H(b)$ is expensive, so a better solution is to approximate $H(b)$ as a function of marginal beliefs. In the region graph representation, we usually replace $b$ by a family of marginal region beliefs and introduce a set of *constraints* (Yedidia et al., 2005) (i.e. ensuring global consistency by connections and local consistency) on these beliefs.

We can derive the similar procedure based on region beliefs $b_R(\mathbf{x}_R)$ to approximate the true probability $p_R(\mathbf{x}_R)$. The free energy for each region is:

$$\begin{aligned} F_R(b_R) &= \sum_{\mathbf{x}_R} b_R(\mathbf{x}_R) E_R(\mathbf{x}_R) - \sum_{\mathbf{x}_R} b_R(\mathbf{x}_R) \log b_R(\mathbf{x}_R) \\ &= U_R(b_R) - H_R(b_R) \end{aligned} \tag{3.12}$$

The region based free energy is:

$$F(\{b_R\}) = \sum_R c_R U_R(b_R) - \sum_R c_R H_R(b_R) \tag{3.13}$$

Where $c_R$ corresponds to counting number of each region. The constrained region free energy $F_R(b_R)$ must be minimized. The relationship between minimizing a system free energy and maximizing the system entropy at the system equilibrium is well known in the theory of thermodynamics (Callen, 2006), which proves that for a closed system with fixed internal energy the entropy is maximized at equilibrium (fixed point of GBP), and the free energy (i.e. Gibbs) is minimized at equilibrium given fixed entropy.

So we are most interested in the accuracy of the constrained region based entropy near its *maximum* (Yedidia et al., 2005). The maximum of the true entropy occurs when the joint probability distribution is uniform. Yedidia (Yedidia et al., 2005) introduced a similar property to hold for constrained region based entropies,

*Maxent-normal*: a constrained region based free energy approximation is maxent-normal if it is valid and the corresponding constrained region based entropy $H_R(\{b_R\})$ achieves its maximum when all the beliefs $b_R(\mathbf{x}_R)$ are uniform.

We aim to maximize the overall region entropy $H_R(\{b_R\}) = \sum_R c_R H_R(b_R)$ when all beliefs are uniform. As pointed out by Yedidia (Yedidia et al., 2005), if the region based approximation is not maxent-normal one cannot expect a good result, because it will always produce the wrong answer even when there is no energy term. We will use the maxent-normal property as a requirement to our TRC algorithm in Chapter 5.

### 3.4.5 GBP Message Passing

There are several ways of message passing in GBP: parent to child, child to parent and two way message passing. All message passing algorithms are derived from belief equations. Each algorithm has its advantages and disadvantages but here the two way message algorithm is used as the basis for the TRC algorithm, since TRC guarantees to provide a region graph to be a DAG. Therefore here our discussion will focus on two way message passing. The two way message algorithm is particularly elegant when each region and its sub-regions forms a tree, and all factors appear in the first level of region graph (Yedidia et al., 2005).

In the two way message algorithm the belief equation in a region is a product of local factors and messages arriving from all the connected regions, whether they are parents or children, as in Equation 3.14.

$$b_R(\mathbf{x}_R) = \tilde{f}_R(\mathbf{x}_R) \prod_{C\in c(R)} n_{C\to R}(\mathbf{x}_C) \prod_{P\in p(R)} m_{P\to R}(\mathbf{x}_P) \tag{3.14}$$

The two way message algorithm uses the region counting numbers $c_R$ and parent numbers of region $R$ as input parameters to compute a message parameter $\beta_R$, then it defines a set of pseudo-messages for all regions $R$ and their parents $P$ and children $C$, as pseudo-messages is computed prior to real message updating. Then the real messages are a mixture of these pseudo messages.

The set of pseudo-messages for all regions $R$ and their parents $P$ and children $C$:

$$n^0_{R\to P}(\mathbf{x}_R) = \tilde{f}_R(\mathbf{x}_R) \prod_{P'\in p(R)\backslash P} m_{P'\to R}(\mathbf{x}_R) \prod_{C\in c(R)} n_{C\to R}(\mathbf{x}_C) \tag{3.15}$$

And

$$m^0_{R\to C}(\mathbf{x}_C) = \sum_{\mathbf{x}_R\backslash \mathbf{x}_C} \tilde{f}_R(\mathbf{x}_R) \prod_{P\in p(R)} m_{P\to R}(\mathbf{x}_R) \prod_{C'\in c(R)\backslash C} n_{C'\to R}(\mathbf{x}_{C'}) \tag{3.16}$$

Where $\tilde{f}_R(\mathbf{x}_R) \equiv \left(\prod_{a\in A_r} f_a(\mathbf{x}_a)\right)^{c_R}$. The real message used in the two directions of a link is the mixture of these pseudo-messages:

$$n_{R\to P}(\mathbf{x}_R) = \left(n^0_{R\to P}(\mathbf{x}_R)\right)^{\beta_R} \left(m^0_{P\to R}(\mathbf{x}_R)\right)^{\beta_R - 1} \tag{3.17}$$

And

$$m_{P\to R}(\mathbf{x}_R) = \left(n^0_{R\to P}(\mathbf{x}_R)\right)^{\beta_{R-1}} \left(m^0_{P\to R}(\mathbf{x}_R)\right)^{\beta_R} \tag{3.18}$$

For simplicity we omit the parameter $\beta_R$'s definition equations here, which can be referred to (Yedidia et al., 2005).

Once the RG is produced the two way message passing is performed using a depth first search algorithm for the updating order of region edges. When each propagation has completed the old messages are replaced by new messages, and the region belief is calculated by the product of these messages with local factors. Each propagation may be composed of multiple iterations updating the region edges.

# 4. BFE Risk Aggregation

This chapter covers Bayesian risk aggregation algorithms for hybrid models. Section 4.1 provides an overview of popular methods for risk aggregation. Section 4.2 illustrates the *n-fold* convolution using BNs. The BFE risk aggregation algorithm is described in Section 4.3 showing how it builds and extends on the standard BN algorithms. Section 4.4 presents a version of BFE that performs deconvolution and Section 4.5 presents experimental results showing the performance of BFE. Section 4.6 concludes the chapter.

## 4.1. Risk Aggregation and BNs

An encyclopaedic overview of the current state of the art in risk aggregation is presented in (McNeil, Frey, & Embrechts, 2010). The general aggregation formula for fixed, $n$, assets, is:

$$T = S_0 + S_1 + ... + S_n \tag{4.1}$$

where $T$ is the sum of $n$ asset valuations and each $S_i$ is from the same common continuous distribution $f_x$, which can be thought of as a return (severity) distribution $S$. This is called an *n-fold* convolution. If $S \sim f_x$ and if we have a variable number of assets, $N$, then Equation 4.1 can be rewritten as an *N-fold* convolution:

$$f_T(x) = \sum_{j=0}^{\infty} f^{*j}(x) P(N = j) \tag{4.2}$$

where $f^{*j}(x) = \int_0^{\infty} f^{*(j-1)}(x-y) f(dy)$ is a recursive *n-fold* convolution on $S$. We can therefore rewrite Equation 4.2 in a discrete form: $P(N = j) = a_j$, for $j = 0,1,...,L$, where $L$ is the length of discretized frequency $N$. The following expressions hold:

$$P(T = t) = a_0 P(T_0 = t_0) + a_1 P(T_1 = t_1) + ... + a_L P(T_L = t_l) \tag{4.3}$$

$$T_0 = S_0, T_1 = S_0 + S_1, ..., T_L = S_0 + S_1 + ... + S_L \tag{4.4}$$

where each $T_j$ is a constant *n-fold* convolution. The Equation 4.3 represents a mixture distribution where the mixture components consist of mutually exclusive variables, themselves composed using the conditionally deterministic functions stated in Equation 4.4.

For the sake of clarity in insurance, and similar, applications $N$ is interpreted as a frequency distribution, and $S$ is defined as a severity (loss) distribution whereas, for some other financial applications, the interpretation of the parameters differs: we could, equally validly, interpret, $N$, as a count of assets and $S$ as the financial return from each asset.

General numerical solutions to computing the aggregate distribution include Panjer's recursion (Panjer, 1981), Fast Fourier transform (Heckman & Meyers, 1983). and Monte Carlo (MC) simulation (Meyers, 1980).

In this chapter severity variables can depend on discrete explanatory variables with dependencies expressed via conditioning in Bayesian networks. This contrasts with the classic approach for dependency modelling among severity variables using copulas. Rather than use dependency and conditioning the copula approach models the dependency structure independently with marginal functions, which supports the construction of high dimensional models.

In the context of copula based risk aggregation Bruneton (Bruneton, 2011) proposes the use of hierarchical aggregation using copulas. Also, Arbenz (Arbenz & Canestraro, 2012) proposes hierarchical risk aggregation based on tree dependence modelling using step-wise low dimensional copulas, and also gives a sample reordering algorithm for numerical approximation. Brechmann (Brechmann, 2014) suggests hierarchical Kendall copulas to achieve flexible building blocks, where risk aggregation is supported by the Kendall function. These approaches capture the joint dependencies from a hierarchical structure and exploit use of small building blocks. In contrast to correlation modelling, our work assumes causality and dependency, where joint dependency is decomposed by conditional dependencies using the Bayesian network framework.

BNs have already been employed to address financial problems. For example, in (Cowell, Verrall, & Yoon, 2007) BNs were used for overall loss distribution and making predictions for insurance; in (Neil & Fenton, 2008) BNs were used for modelling operational risk in financial institutes, while the work in (Politou & Giudici, 2009) combines Monte Carlo simulation, graphic models and copula functions to build operational risk models for a bank. Likewise, (Rebonato, 2010) discusses a coherent stress testing approach using BNs.

We have chosen to use BNs because the latest algorithms can model causal dependencies between discrete and continuous variables during inference, to produce approximate posterior marginal distributions for the variables of interest. Also, by virtue of Bayes' Theorem they are agnostic about causal direction and can perform inference from cause to effect and vice versa (or convolution to de-convolution, as is the case here). Until very recently BN tools were unable to properly handle non-Gaussian continuous variables, and so such variables had to be discretized manually, with inevitable loss of accuracy. A solution to this problem is DDJT algorithm described in section 3.3.2. The result of inference is a set of queries on the BN in the form of univariate or multivariate posterior marginal distributions. This allows the approximate solution of classical Bayesian statistical problems, involving continuous variables, as well as hybrid problems involving both discrete and continuous variables, without any restriction on distribution family or any requirement for conjugacy.

We have used AgenaRisk (AgenaRisk, 2014), a BN package and extended it to incorporate the new BFE algorithm and carry out the experiments described in Section 4.5.

## 4.2. *n-fold* Convolution BF Process

The cost of using off-the-shelf BN algorithms to calculate *N-fold* convolution can be computationally expensive. The conditional probability density expression of node $T$ is defined by all of its parent nodes by Equation 4.1: $T_n = S_0 + S_1 + ... + S_n$.

If each node has a node state of size $m$ and the total number of parents is $n$, then the NPT for $T$ has a total size of $m^{n+1}$ given the intervals computed under DD. To help reduce the NPT size we employ binary factorization to factorize the BN graph according to the statistical and deterministic functions declared in it.

To illustrate the BF process, we consider constant *n-fold* convolution models for both the independent and common cause case, as represented by BNs $G1$ and $G2$ respectively in Figure 4.1. This is just Equation 2.1.

After employing binary factorization, the BNs $G1$ and $G2$ are transformed into $G1'$ and $G2'$ respectively as shown in Figure 4.1. In Figure 4.1 $G1$ shows the *n-fold* convolution when severities are independent and identically distributed. $G2$ denotes the *n-fold* convolution when severities are dependent on a discrete common cause random vector, $\mathbf{C}$.

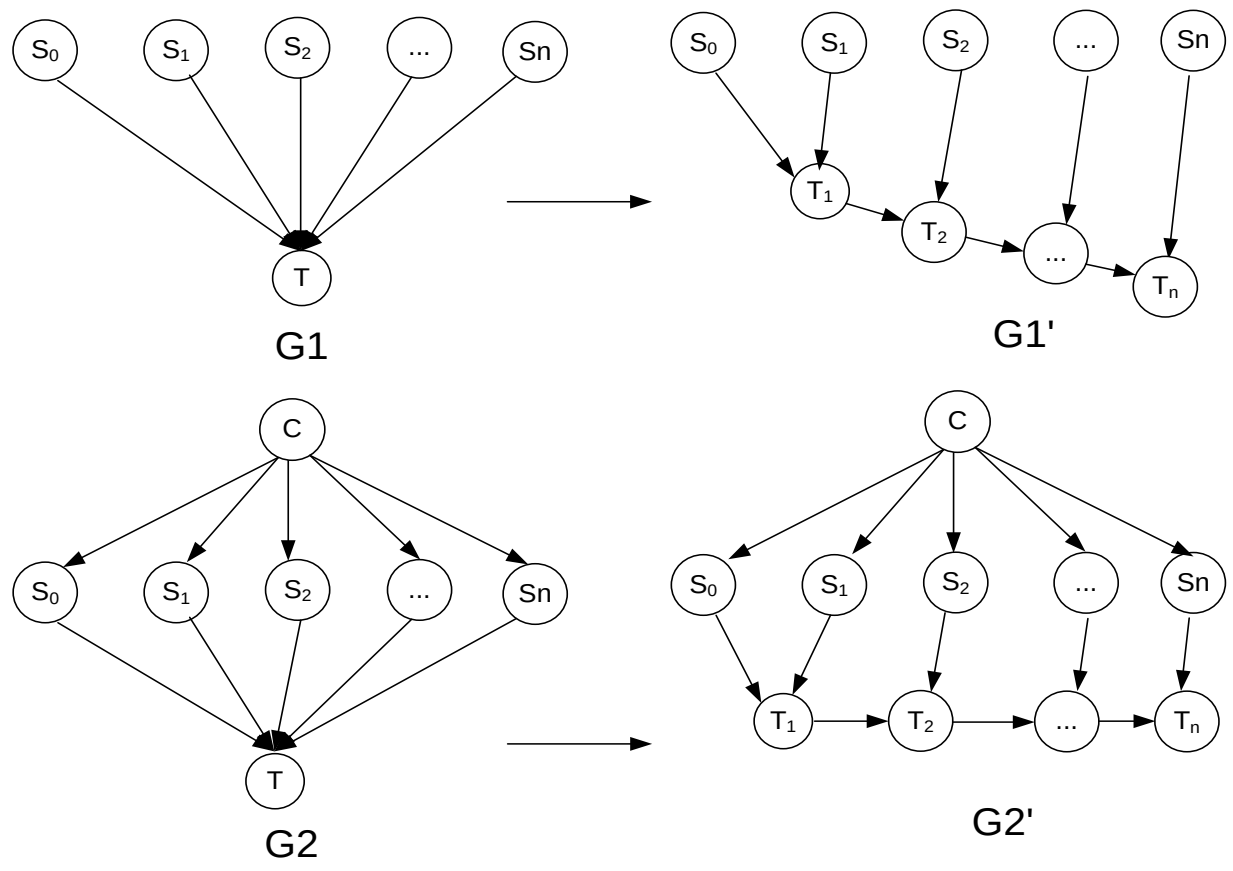

Figure 4.1 BN models of n-fold convolution of i.i.d. severity variables ($G1$) and of common cause version ($G2$) with accompanying binary factorized versions ($G1'$ and $G2'$)

BF ensures that, in the transformed BN, each variable's NPT expression involves has a maximum of two continuous parent variables in the transformed BN. This produces a maximal discretized NPT of size $m^3$. Consider the two BNs, $G1$ and $G2$, shown in Figure 3. In each we can factorize the variable $T$ by creating new variables $\{T_1, T_2, ..., T_{n-1}\}$ where each is binary factorized by pair sum blocks (Equation 4.5):

$$\{T_1 = S_0 + S_1\},\ \{T_2 = T_1 + S_2\}, \ldots,\ \{T_n = T_{n-1} + S_n\} \quad (4.5)$$

Theoretical equivalence of $G1$ and $G1'$ with the resulting BN models $G2$ and $G2'$ is given in (Neil et al., 2012).

# 4.3. Bayesian Factorization and Elimination (BFE)

To solve the *N-fold* convolution problem using off-the-shelf BN technology is not possible because we cannot compute $G1$ and $G2$ effectively from the conditional dependency structures defined in Figure 4.1. This is because, even with binary factorization, either the model size is prohibitively large (in the case of $G1$) or the junction trees clique sizes would be exponential in size (as with $G2$). Therefore, the original contribution of this thesis is to produce an iterative factorized approach to the computation that scales up to arbitrary sized models. This approach is called *Bayesian Factorization and Elimination* (BFE). This algorithm performs convolution on the hybrid models required to aggregate risk in the presence (or absence) of causal dependencies. This algorithm exploits a number of advances from the field of BNs already described in chapter 2. We refer to these advances as the BN engine and they are shown in the overall algorithm architecture in Figure 4.2.

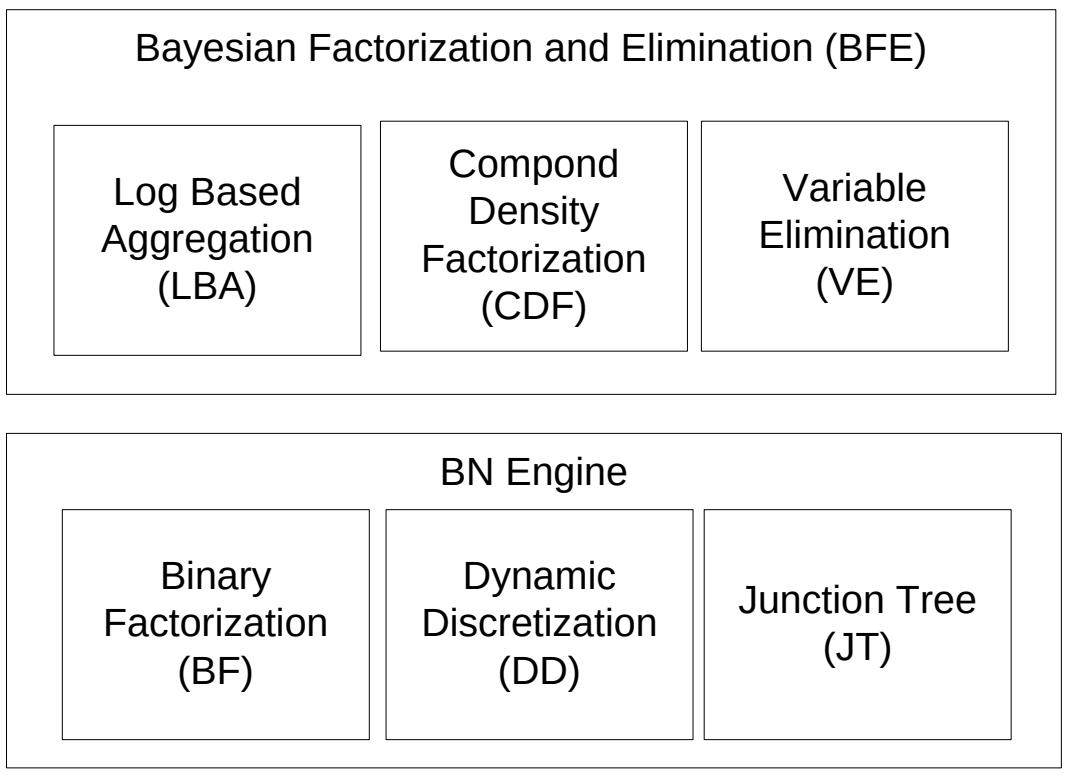

Figure 4.2 Architecture of BN algorithms

The BFE algorithm contains three separate steps, each performing specific optimisations:

1. Log Based Aggregation (LBA): this algorithm computes Equation 4.4, the *n-fold* convolution, in a log based pattern that can be more efficient than aggregation by straight summation.

2. Variable Elimination (VE): variables are iteratively eliminated during LBA process, by which we can achieve greater computation efficiency for calculating arbitrary constant *n-fold* convolutions.

3. Compound Density Factorization (CDF): the compound sum Equation 4.3 can be factorized by this algorithm in order to reduce large node probability tables into smaller ones. CDF is similar to binary factorization except that in CDF we introduce one more intermediate variable (a Boolean node) for weighting the compound density combination at each step in the aggregation process.

## 4.3.1 Log Based Aggregation (LBA)

In Equation 4.3 each $T_i, i = 1,...,n$ is the sum of its parent variables $T_{i-1}$ and $S_i$, the aggregation process simply involves repeated summations of the same variable $S_i$. As binary factorization proceeds intermediate variables $F_j$ are created to aggregate every two parents, creating a hierarchy until the total aggregate, $T$, is computed. An example, in the presence and absence of common cause vector is shown in Figure 4.3 (for convenience we have assumed the hierarchy has depth three and the other level contains intermediate variables labelled $F_j$). The computational efficiency of this process is $O(n)$.

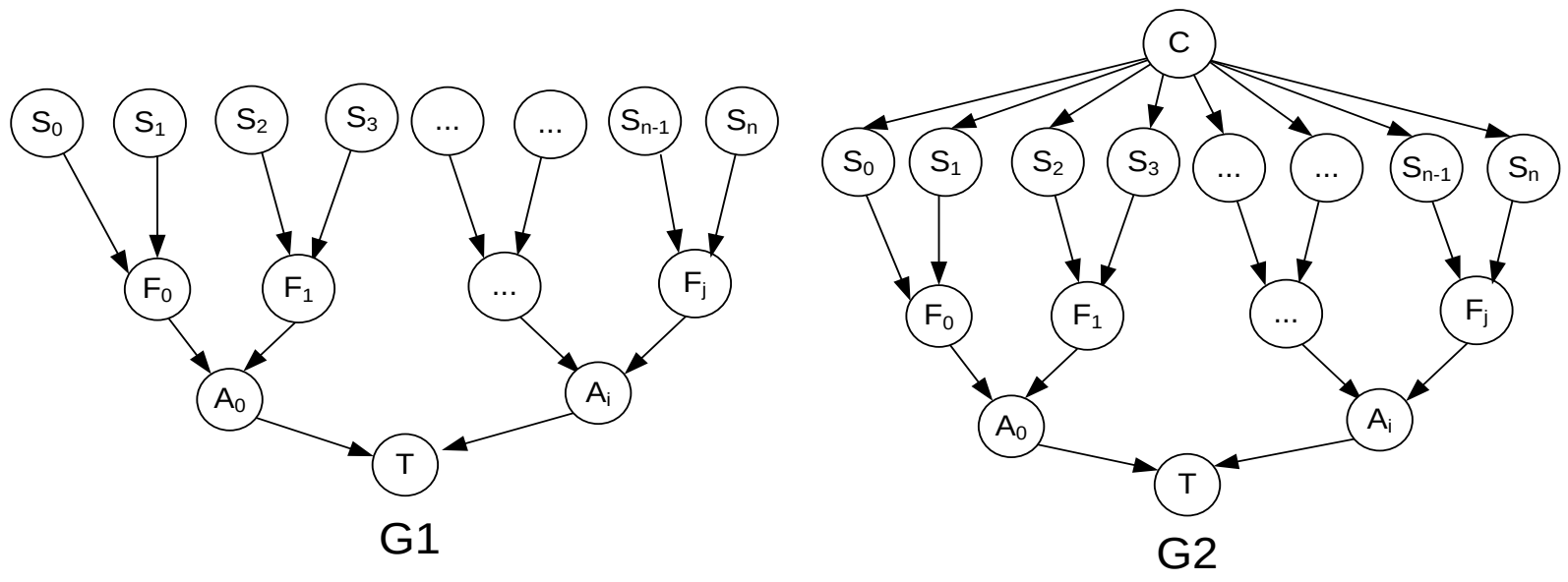

Figure 4.3 $G1$ and $G2$ BNs binary factorized for aggregation

This approach to aggregation is computationally expensive since all the variables are entered and computed in the BN explicitly. Log based aggregation simply computes and subsequently reuses prior computed results recursively, so that in each subsequent step we can reuse results from previous steps, without creating the whole

BN. Thus rather than create and compute the BN as a whole, we create and reuse BN fragments and then remove (prune) those fragments of the BN we do not need. For instance, to sum four i.i.d. variables we would sum two variables and then add the result of this summation to itself to get the total; thus achieving the aggregate total in two rather than three steps. The resulting process is $O(\log_2 n)$, hence the name log-based aggregation.

### 4.3.2 Variable Elimination (VE)

The aim of Variable Elimination (VE) is to remove nodes from a BN, $G$, that do not belong to a query set, $Q$, containing only the variables of interest, by a process of marginalization. For simple uncorrelated aggregations this process is simple and obvious but in the presence of common causes it requires some care, since the nodes being eliminated will not be leaf nodes. Here we use variable elimination to reduce the number of variables we handle but add additional steps to exploit repeated structure in the binary factorized model. We do not need, therefore, to explicitly manipulate the whole BN, nor do we create a large junction tree or use the junction tree algorithm over this large tree, because we are not interested in setting arbitrary query variables or conditioning evidence. Instead we iterate through the binary factored model, progressively creating subsets of the aggregation hierarchy that can be reused recursively, eliminating nodes and reusing parts as we go (assuming i.i.d. severity variables).

We first consider a full binary factorized BN and use this to identify variables that can be eliminated and query sets necessary during VE. In the simple case for an *n-fold* convolution for independent i.i.d. severity variables, Figure 4.1 $G1'$ denotes the binary factorized form of the computation of $T_n = \sum_{j=0}^{n} S_j$ after introducing the intermediate binary factored variables $\{T_1, T_2, ..., T_{n-1}\}$. The marginal distribution for $T_n$ has the form:

$$P(T_n = t_n) = \sum_{(s_0 \ldots, s_n, t_1, \ldots, t_{n-1})} P(S_0 = s_0, S_1 = s_1, \ldots, S_n = s_n, T_1 = t_1, T_2 = t_2, \ldots, T_{n-1} = t_{n-1}, T_n = t_n)$$
$$= \sum_{(s_0 \ldots, s_n, t_1, \ldots, t_{n-1})} P(T_n = t_n \mid T_{n-1} = t_{n-1}, S_n = s_n) P(T_{n-1} = t_{n-1} \mid T_{n-2} = t_{n-2}, S_{n-1} = s_{n-1}) \ldots$$
$$P(T_1 = t_1 \mid S_0 = s_0, S_1 = s_1) P(S_0 = s_0) P(S_1 = s_1) \ldots P(S_n = s_n)$$

(4.6)

(Exploiting the conditional independence relations in Figure 4.1)

Notice that every pair of parent variables $T_i$ and $S_{i+1}$ is independent in this model and we can marginalize out each pair of $T_i$ and $S_{i+1}$ from the model separately. Equation 4.6 can be alternatively expressed as predefined ‘query blocks’:

$$P(T_n = t_n) = \sum_{t_{n-1}, s_n} P(T_n = t_n \mid T_{n-1} = t_{n-1}, S_n = s_n)$$
$$\left\{ \cdots \left\{ \sum_{t_1, s_2} P(T_2 = t_2 \mid T_1 = t_1, S_2 = s_2) \left\{ \sum_{s_0, s_1} P(T_1 = t_1 \mid S_0 = s_0, S_1 = s_1) P(S_0 = s_0) P(S_1 = s_1) \right\} P(S_2 = s_2) \right\} \cdots \right\}$$
$$\cdot P(S_n = s_n)$$

(4.7)

So using Equation 4.7 we can recursively marginalize out, i.e. eliminate or prune, each pair of parents $T_i$ and $S_{i+1}$ from the model. For example, the elimination order in Equation 4.7 could be: $\{S_0, S_1\}, \{T_1, S_2\} \ldots \{T_{n-1}, S_n\}$. The marginal distribution of $T_n$, i.e. the final query set, is then yielded at the last elimination step.

In order to illustrate the recursive BN graph operations, required during VE, consider Figure 4.1 and BN $G1$. The first few steps involved are shown in Figure 4.4. We start by taking the first pair of severity variables $S_0$ and $S_1$ and calculate the sum, $F_0$, shown as the graph $K_1$. Once we have computed $K_1$ we can reuse this calculation in graph $K_2$, for severities $S_2$ and $S_3$ (of course if the severities are i.i.d we can simply reuse the result at $F_0$ and substitute this for $F_1$). We are now interested in using the marginal distributions of $F_0$ and $F_1$ in the next step so these are added to the query set and the nodes $S_0$ and $S_1$ are eliminated, thus reducing graph $K_1$ to $L_1$. Similarly for graph $K_2$ we eliminate $S_2$ and $S_3$ to get a graph $L_2$.

Next we can reuse the original structure in $K_1$ and substitute $K_1$'s leaf nodes with $F_0$ and $F_1$, and then compute $A_0$. The resulting $A_0$ now becomes the query set and we eliminate $F_0$ and $F_1$ to achieve graph $L_3$. At each stage we reuse the same graph structures and expressions for graphs $\{K_1, K_2, K_3\}$ and $\{L_1, L_2, L_3\}$. We can proceed through the binary factorized BN, computing the marginal distributions for the query set, removing elimination sets and repeating the process until we exhaust the variable list.

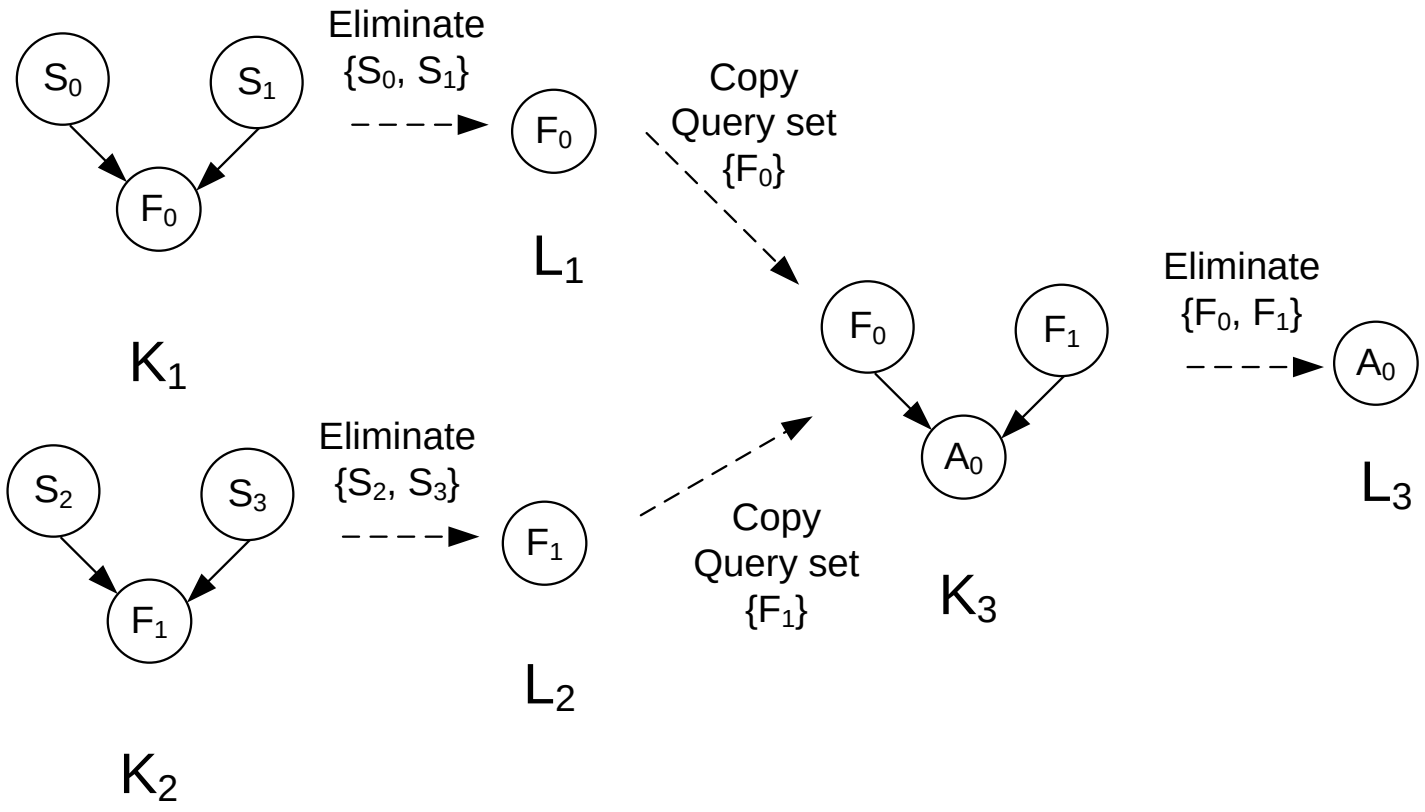

Figure 4.4 VE process applied to part of BN $G1$

However, in the case where common cause dependencies are present in the BN, as illustrated by $G2$ shown in Figure 4.1, additional care is needed during VE. Here the elimination set does not simply consist of leaf nodes that can be eliminated directly since we have a common parent node, $C$, that we want to preserve in the query set at each step. To help highlight how the VE process operates in the presence of common cause variables consider BN $G'$ and compute the posterior marginal distribution for $T_2$. The marginal distribution for $T_2$ has the form (4.8):

$$
\begin{aligned}
P(T_2 = t_2) &= \sum_{c, s_0, s_1, s_2, t_1} P(T_2 = t_2 \mid T_1 = t_1, S_2 = s_2) P(T_1 = t_1 \mid S_0 = s_0, S_1 = s_1) P(S_0 = s_0 \mid C = c) \\
&\quad \cdot P(S_1 = s_1 \mid C = c) P(S_2 = s_2 \mid C = c) P(C = c) \\
&= \sum_{c, s_2, t_1} P(T_2 = t_2 \mid T_1 = t_1, S_2 = s_2) P(S_2 = s_2 \mid C = c) P(C = c) \\
&\quad \cdot \left\{ \sum_{s_0, s_1} P(T_1 = t_1 \mid S_0 = s_0, S_1 = s_1) P(S_0 = s_0 \mid C = c) P(S_1 = s_1 \mid C = c) \right\}
\end{aligned} \tag{4.8}
$$

We first want to eliminate $S_0$ and $S_1$ by marginalizing them:

$$P(T_1 = t_1 \mid C = c) = \sum_{s_0, s_1} P(T_1 = t_1 \mid S_0 = s_0, S_1 = s_1) P(S_0 = s_0 \mid C = c) P(S_1 = s_1 \mid C = c) \tag{4.9}$$

The marginal of $T_2$ can now be expressed along with $C$, $T_1$ and $S_3$ alone:

$$P(T_2 = t_2) = \sum_{c, s_2, t_1} P(T_2 = t_2 \mid T_1 = t_1, S_2 = s_2) P(S_2 = s_2 \mid C = c) P(T_1 = t_1 \mid C = c) P(C = c)$$

Next we eliminate $S_2$ and $T_1$:

$$P(T_2 = t_2 \mid C = c) = \sum_{t_1, s_2} P(T_2 = t_2 \mid T_1 = t_1, S_2 = s_2) P(S_2 = s_2 \mid C = c) P(T_1 = t_1 \mid C = c) \tag{4.10}$$

In general, by variable elimination, we obtain the conditional distribution for each variable $T_{n-1}$ (the sum of $n$ severity variables) with the form:

$$\begin{aligned} P(T_{n-1} = t_{n-1} \mid C = c) = \sum_{t_{n-2}, s_{n-1}} & P(T_{n-1} = t_{n-1} \mid T_{n-2} = t_{n-2}, S_{n-1} = s_{n-1}) \\ & \cdot P(T_{n-2} = t_{n-2} \mid C = c) P(S_{n-1} = s_{n-1} \mid C = c) \end{aligned} \tag{4.11}$$

Since Equation 4.11 specifies the conditional distribution for variable $T_{n-1} \mid C$, therefore the posterior marginal distribution for the target *n-fold* convolution $T_{n-1}$, the aggregate total, is obtained by marginalizing out $C$.

In order to explain the VE algorithm, in terms of graph manipulation, in the common cause case we step through a *3-fold* convolution. Figure 4.5 (a) depicts a *3-fold* convolution model, binary factorized (from $G$ to $G'$) and then subject to VE, resulting in reduced the BN $V$. The VE steps are shown in Figure 4.5 (b), which, although operating on subsets of $G$, result in the same graph i.e. $L_2 = V$.

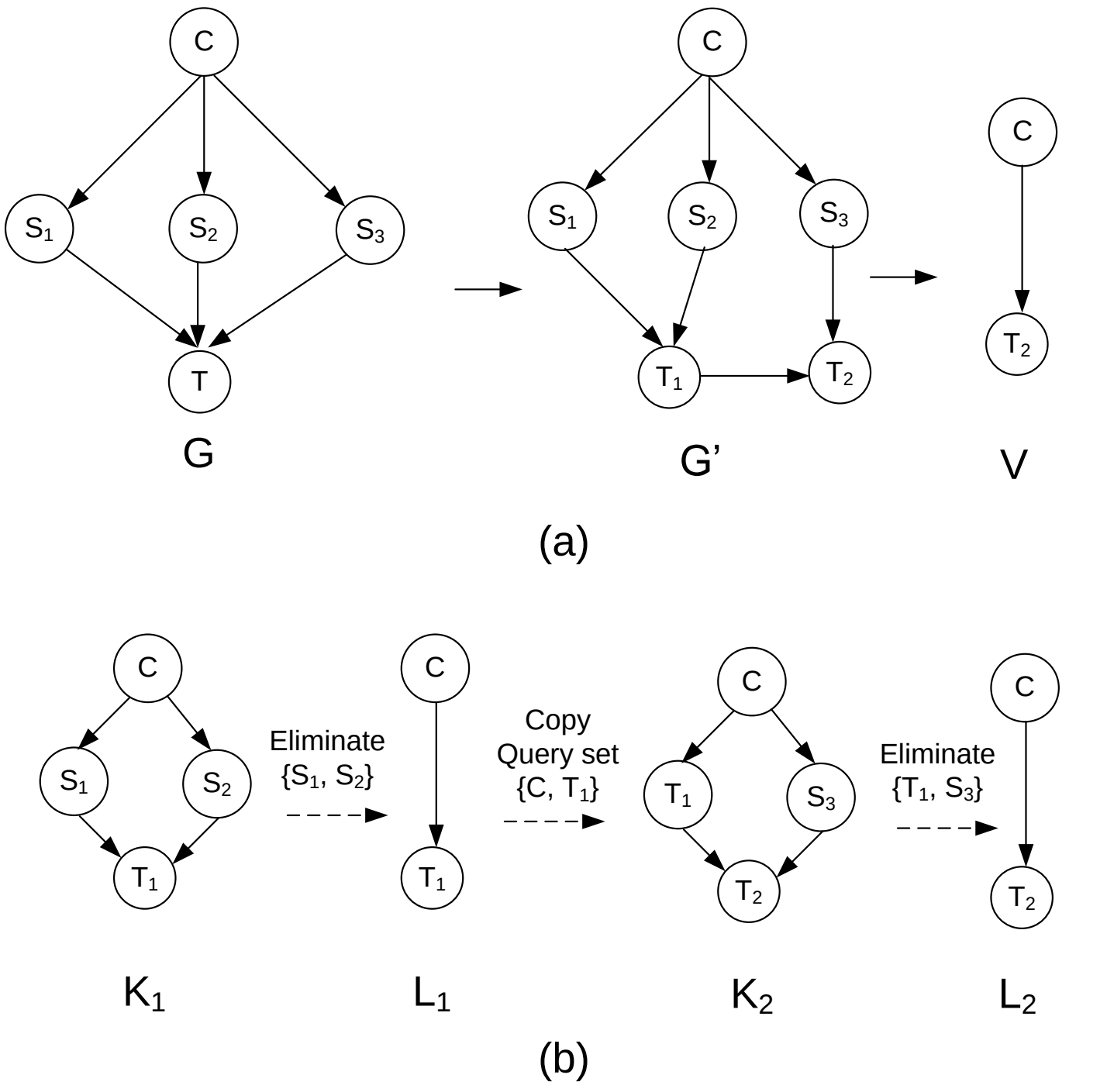

Figure 4.5 (a) Simple common cause model binary factorization and VE process, (b) VE process applied to part of BN $G$

To calculate the arbitrary *n-fold* convolution in the multiple common cause case it is essential to maintain the structure connecting the common causes in $G'$ in every elimination step so that when variables are eliminated any dependencies on common cause variables are correctly maintained. Consequently the elimination task involves generating the marginal for variable $T_j$ conditional on the set $\mathbf{C} = C_0, C_1, ..., C_m$. This more general case is shown in Figure 4.6, with multiple common cause variables $C_0, C_1, ..., C_m$, and dependent severity variables, $S_i$. It is easy to see how the scheme can be generalised to any configuration of common causes (e.g. including parent nodes of the common causes).

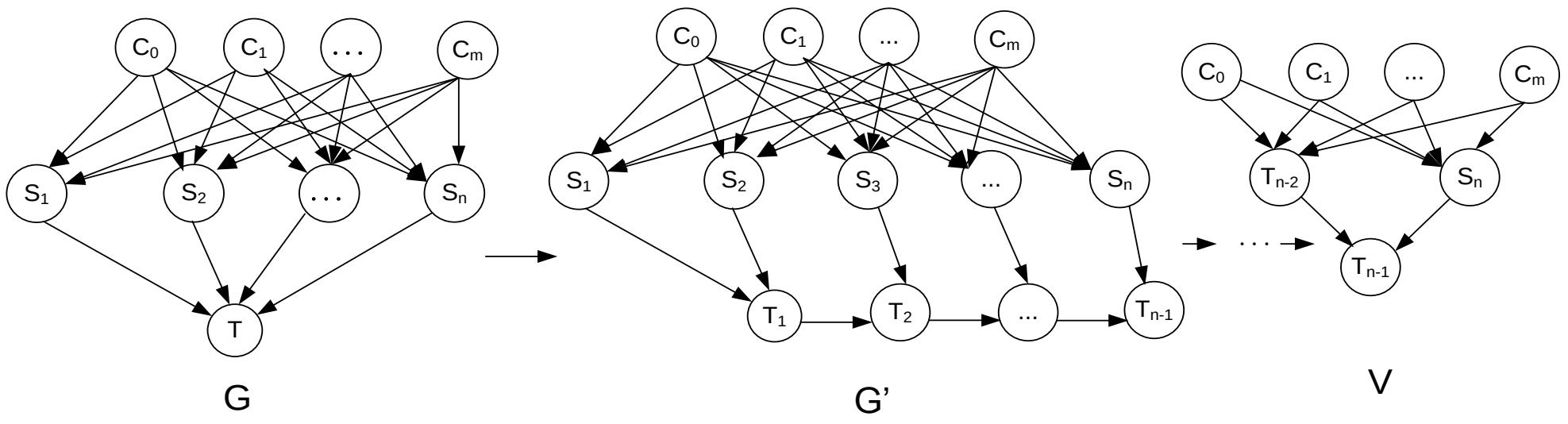

Figure 4.6 Multiple common cause model binary factorization and VE process

## 4.3.3 Compound Density Factorization (CDF)

The compound density factorization (CDF) involves the approximate inference in the context of mixture models. Similar approaches, such as Minka (Minka & Winn, 2008) proposed the "Gates" representation of mixture models in factor graph, and discussed to apply a variety of approximate inference approaches (i.e. EP, message passing, Gibbs sampling) with gate models. Our CDF algorithm uses DDJT as the approximate inference for mixture models, and is an implementation version of cut set conditioning (Pearl, 1988).

Recall the compound density expression for an *N-fold* convolution, as given in Equation 4.3, where $T_j = \sum_{i=0}^{j} S_i,\ j = 0...L\ (length\ of\ N)$ is an *i-fold* convolution with $S$ itself and $a_j = P(N = j)$ is the weighting assigned to the corresponding $T_j$. Unfortunately, the compound density expression for $P(T)$ is very space inefficient and to address this we need to factorize it. Given each component in the mixture is mutually exclusive, i.e. for a given value of $N$ the aggregate total is equal to one, and only one $T_i$, variable, this factorization is straightforward. However, we cannot use a binary factorization for Equation 4.3, therefore we factorize the compound density expression into pair block densities and combine each block density incrementally.

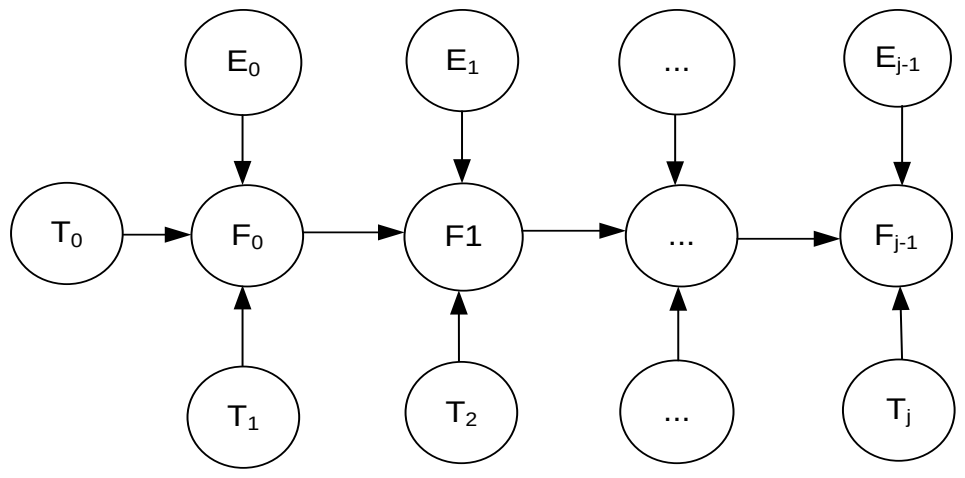

Figure 4.7 Compound density factorization

Equation 4.3 is factorized as shown in Figure 4.7, where additional Boolean variables, $E_j$ (with only two states *True* and *False*)[6], are introduced to assign weightings proportional to $a_j$, to each pair of block nodes, for example,

[6] "True" and "False" are used for convenience; any binary labelling would do equally well.

$\{T_0, T_1\}, \{F_0, T_2\}, ..., \{F_{j-2}, T_j\}$ . Factor variables, $F_j$ , are created to calculate the weighted aggregate for each step, up to the length of the *N-fold* convolution, $L$ .

The node probability table for $E_{j-1}$ is defined by the following:

$$P(E_{j-1} = True) = \frac{a_0 + a_1 + ... + a_{j-1}}{a_0 + a_1 + ... + a_j} \tag{4.12}$$

The conditionally deterministic expression for variable $F_{j-1}$ (called a partitioned node in BN parlance) is defined by:

$$F_{j-1} = \begin{cases} F_{j-2} & \text{if } E_{j-1} = True \\ T_j & \text{if } E_{j-1} = False \end{cases} \tag{4.13}$$

Since $T_0$ and $T_1$ are mutually exclusive, the marginal distribution for variable $F_0$ is:

$$P(F_0 = f_0) = P(E_0 = True)P(T_0 = t_0) + P(E_0 = False)P(T_1 = t_1) = a_0 P(T_0 = t_0) + a_1 P(T_1 = t_1)$$

which is identical to the first two terms in the original compound density expression, Equation 4.3. Similarly, the marginal for variable $F_j$ becomes:

$$P(F_{j-1} = f_{j-1}) = P(E_{j-1} = True)P(F_{j-2} = f_{j-2}) + P(E_{j-1} = False)P(T_j = t_j) \tag{4.14}$$

After applying the CDF method to Equation 4.3 we have the marginal for $F_{j-1}$ as shown by Equation 4.14, which yields the compound density, $P(T)$, for the *N-fold* convolution. Therefore by using the CDF method we can compute the compound density (Equation 4.3) more efficiently. The proof is given in the Appendix, proof A.

The CDF method is a general way of factorizing a compound density. It takes as input any *n-fold* convolution, regardless of the causal structure governing the severity variables. Note that the CDF method can be made more efficient by applying variable elimination (VE) to remove leaf nodes. Likewise we can execute the algorithm recursively reuse the same BN fragment $P(F | F, T, E)$.

### 4.3.4 The BFE Convolution Algorithm with Example

The BFE convolution algorithm is formalised, as pseudo code:

***Algorithm 2 BFE convolution algorithm***

**Input**: $S$ : Severity variable, $N$ : Frequency variable, $C$ : vector of common causes (optional)

**Output**: Compound density $T$

**Main**:

**1.** Compute the probability density function of $N$ , with sample space $Z$ by: $f_N(x) = P(N = x) = P(\{z_j \in Z : N(z) = x\}) = a_j,\ j = 0,1,...,length(Z)$

**2.** **for** $j = 0$ to (length of $Z$ ) **do**

**3.** **for** $i = 0$ to $z_j$ **do**

**4.** Compute $z_j$*-fold* convolution $T_{z_j} = \sum_{i=0}^{z_j} S_i$ by **BF** and **LBA** algorithms

**5.** Eliminate nodes (out of query set) by **VE** algorithm

**6.** **end for**

**7.** **While** $j \geq 2$ **do**

**8.** Apply **CDF** algorithm to factorize (4.3) by probability density of $N$ , Compute $F_{j-1} = P(E_{j-1} = True)P(F_{j-2}) + P(E_{j-1} = False)P(T_{z_j})$

**9.** Eliminate nodes $S_i$, $F_{j-2}$ and $T_{z_j}$ by **VE** algorithm

**10.** **end while**

**11.** **end for**

**12.** **return** $P(F_{j-1})$ {marginal distribution of $T$ }

Algorithm 2 BFE convolution algorithm

**Example 4.1**

Consider a simple example model that aggregates events with marginal frequency $N \sim Geometric(0.5)$ and with marginal severity distribution $S \sim Exponential(1)$, shown in Figure 4.8.

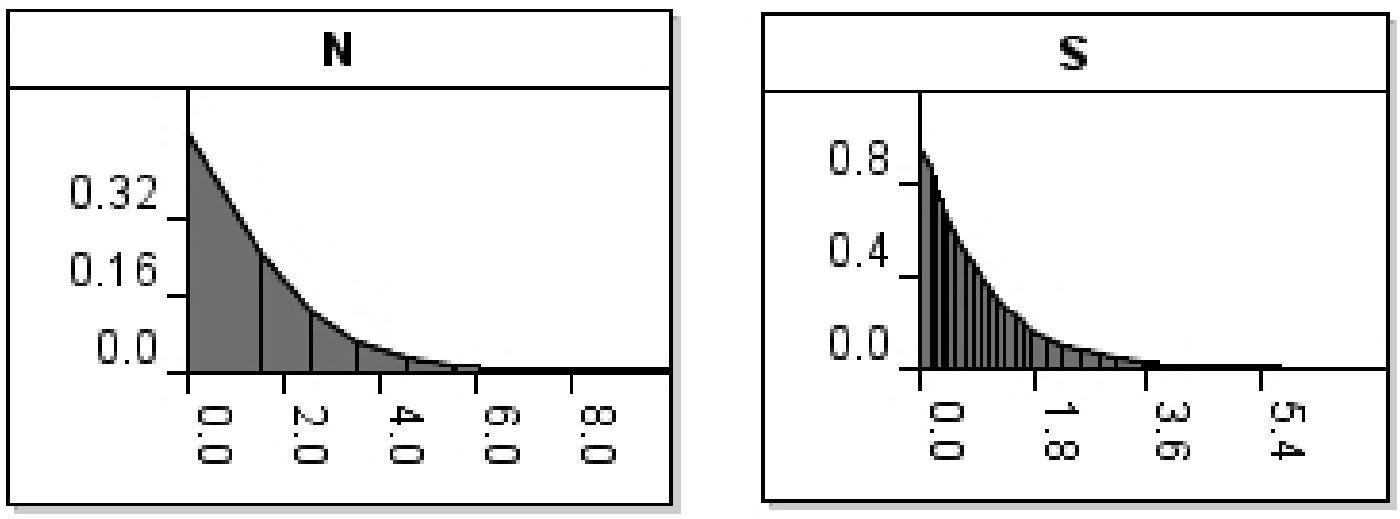

Figure 4.8 Frequency and severity distribution

The algorithm works as follows. The frequency variable $N$ is coarsened by dynamic discretization (DD) with 25 iterations into partitioned intervals, with a probability density associated with each interval. Typically, this is a single integer but where it is an interval we use the midpoint integer value from each interval to form a dataset $D_N$ that contains the numbers of constant *n-fold* convolution with their weightings, $a_0, a_1, ..., a_L$. There were 27 weights needed for this example.

The algorithm first uses LBA (performed using DD set to converge at 65 iterations) and the VE algorithm to calculate each *j-fold* convolution with $j$ from $D_N$ to generate $T_j,\ j = 0...26$. Next we generate the compound density $T$ with weightings corresponding to each $T_j$.

So $P(T) = 0.5 \times P(T_0) + 0.25 \times P(T_1) + ... + 1.16E\text{-}10 \times P(T_{26})$ in this example. The CDF algorithm is then used to factorize the compound density $T$ and compute the marginal for all variables of interest. This involves generating intermediate Boolean variable $E_j$ and sums, $F_j$. We only use the first 4 intervals of $N$ for the following demonstration. The respected dataset is:

$$P(N = 0) = 0.5; P(N = 1) = 0.25; P(N = 2) = 0.125; P(N = 3) = 0.0625$$

For $E_0$ and $F_0$ we get:

$$P(E_0 = False) = \frac{0.5}{0.5 + 0.25} = 0.6667$$

$$P(F_0) = 0.6667 \times P(T_0) + 0.3333 \times P(T_1)$$

Finally we build and execute a BN parameterised with the relevant values. Figure 4.9 shows the partial factorization steps for the first three terms in the compound density. By keeping factorization using the CDF algorithm we yield the compound density $T$ with mean 1 and variance 3.1 in AgenaRisk, while the analytically derived mean and variance are 1 and 3 respectively.

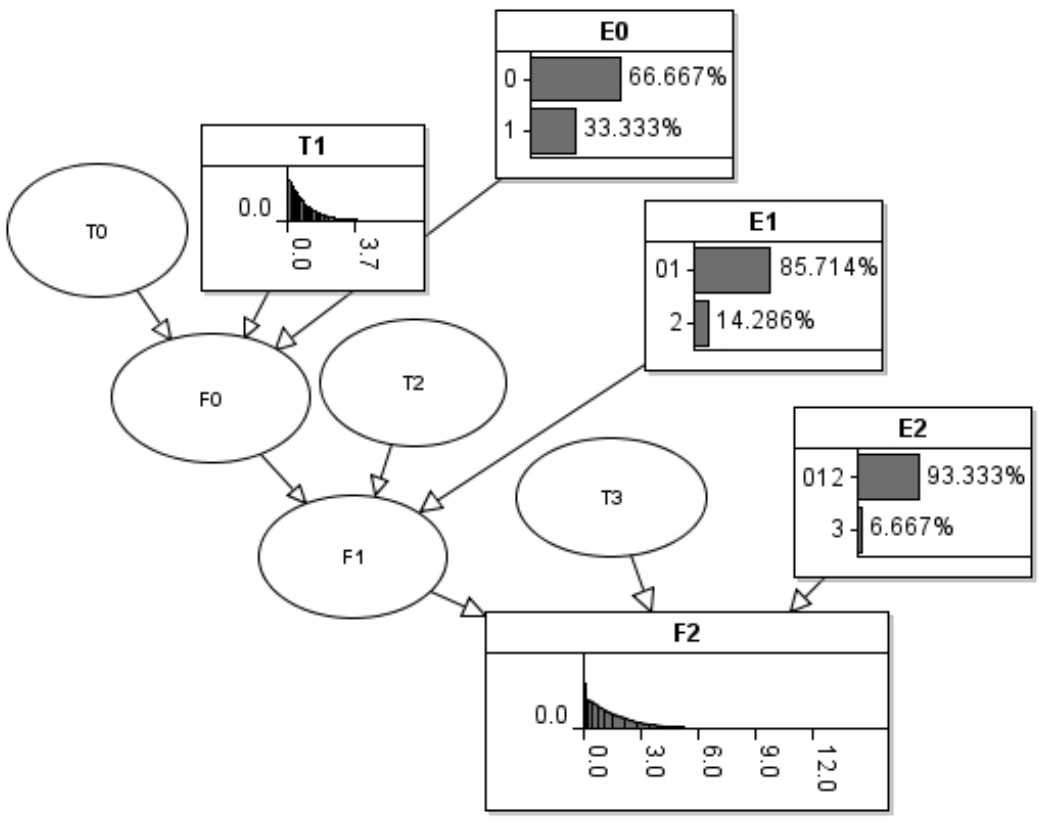

Figure 4.9 Compound density factorization for first three terms in the example

We have factorized the conditional deterministic function $P(T)$ by CDF algorithm, whereby frequency $N$ is factorized by $E_0,...,E_j$ as generating Boolean expressions. This factorization is reversible as we can recover the frequency $N$ by combining $E_0,...,E_j$, where deconvolution of $N$ is achieved.

## 4.4. Deconvolution using the BFE Algorithm

### 4.4.1 Deconvolution

Where we are interested in the posterior marginal distribution of the causes conditional on these consequences we can perform deconvolution, in effect reversing the direction of inference involved in convolution. This is of interest in sensitivity analysis, where we might be interested in identifying which causal variables has the largest, differential, impact on some summary statistic of the aggregated total, such as the mean loss or the conditional Value At Risk (cVAR), derived from $P(C \mid T > t_0)$.

One established solution for deconvolution involves inverse filtering using Fourier Transforms, whereby the severity, $S$, is obtained by inverse transformation from its characteristic function. However, it is first necessary that the density function for $S$ possess an inverse, and should this not exist or if the convolution algebra admits zero divisors, this, unfortunately, results in an infinite number of solutions (Idier, 2010). Alternative analytical estimation methods, i.e. maximum likelihood, and numerical evaluation involving Fourier transforms or simulation based sampling methods, can be attempted but none of them is known to have been applied to *N-fold* deconvolution in hybrid models containing discrete causal variables.

BN based inference offers an alternative, natural, way of solving deconvolution because it offers both predictive (cause to consequence) reasoning and diagnostic (consequence to cause) reasoning. This process is a backwards inference, whereby evidence is entered into the BN on a consequence node and then the model is updated to determine the posterior probabilities of all parent and antecedent variables in the model. A “backwards” version of the BFE algorithm offers a solution for answering deconvolution problems, in a general way without making any assumptions about the form of the density function of $S$. The approach again uses a discretized form for all continuous variables in the hybrid BN, thus ensuring that the frequency distribution, $N$, is tractable. Note that, computationally, the deconvolution process is a natural use of DDJT algorithm.

**Example 4.2** convolution and deconvolution:

To illustrate how backwards propagation works, and by extension deconvolution, let us consider a simple BN with parent variable distributions $X \sim Normal(\mu = 5, \sigma^2 = 5)$, $Y \sim Normal(\mu = 10, \sigma^2 = 10)$ and likelihood function for a child variable $P(Z \mid X, Y) = P(Z = X + Y)$. Figure 4.10 (a) shows the prior convolution effects of the backwards inference calculation, as marginal distributions superimposed on the BN graph. The exact posterior marginal for $Z$ is $Z \sim Normal(\mu = 15, \sigma^2 = 15)$. Our approximation produces a mean of 14.99 and variance 16.28 (DD performed with 25 iterations using AgenaRisk).

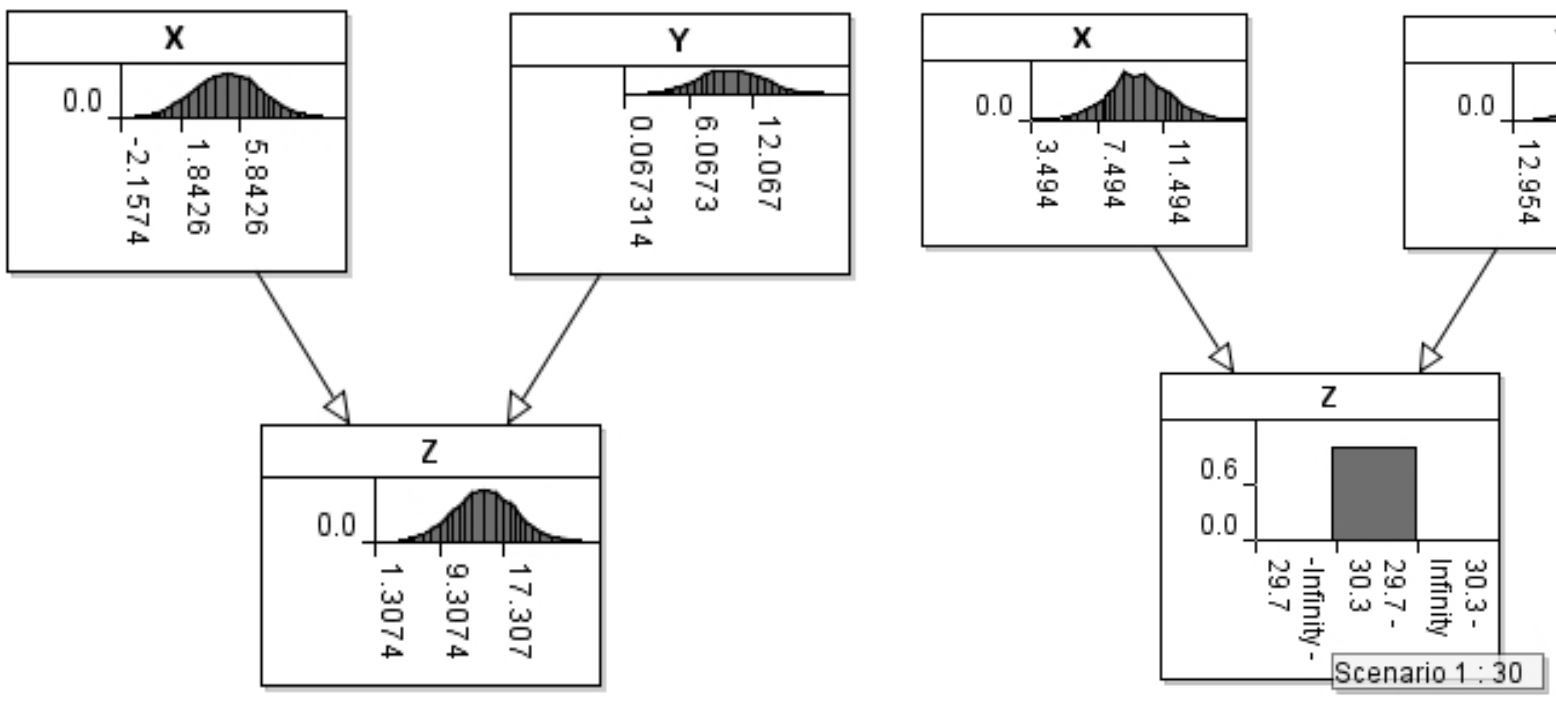

(a) (b)

Figure 4.10 (a) Convolution and (b) Deconvolution

If we set an observation $Z = z_0$ and perform inference we obtain the posterior marginal of $X$ by Bayes' rule:

$$P(X = x \mid Z = z_0) = \frac{P(X = x, Z = z_0)}{P(Z = z_0)} = \frac{\sum_{y} P(Z = z_0 \mid X = x, Y = y)P(X = x)P(Y = y)}{\sum_{x,y} P(Z = z_0 \mid X = x, Y = y)P(X = x)P(Y = y)} \tag{4.15}$$

Where our likelihood $P(Z \mid X, Y)$ is a convolution function, equation (4.15) defines the deconvolution and yields the posterior marginal distribution of $X$ given observation $Z = z_0$. In Figure 4.10 (b), the observation is $Z = 30$ (which is approximated as a discrete bin of given width), and the posterior for $X$ has updated to a marginal distribution with mean 9.97 and variance 3.477.

In the example shown in Figure 4.10 the parent variables $X$ and $Y$ are conditionally dependent given the observation $Z = z_0$. For *n-fold* convolution with or without common causes an observation on the $T_i$ variables would also make each of the severity variables dependent and we can perform *n-fold* deconvolution using the DD and JT alone for small models containing non i.i.d severity variables with query block sizes of maximum cardinality four. For large models, containing i.i.d severity variables BFE provides a correct solution with minimal computational overhead.

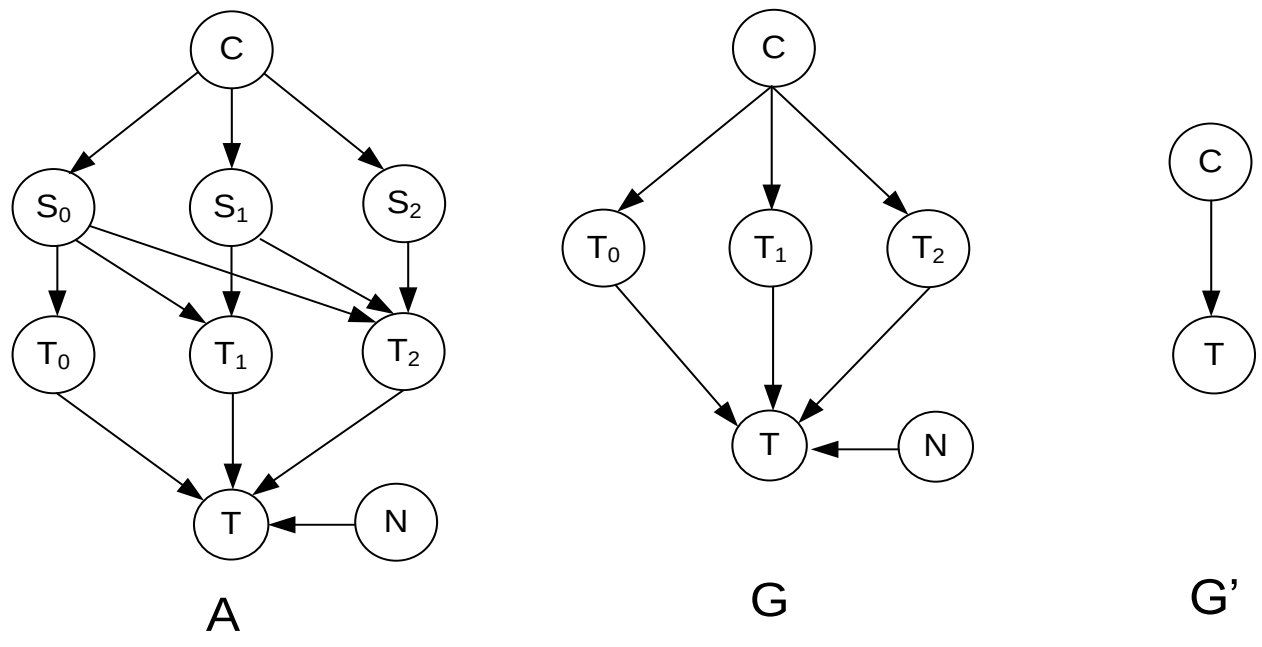

Figure 4.11 Binary factored common cause *N-fold* BN, $A$, reduced by applying the VE algorithm to $G$ and then $G'$

We have already noted that during *N-fold* convolution the $T_i$ variables are mutually exclusive, such that for a given $N = i$, if the variable $T_i$ exists, then the other variables do not. This fact can be exploited during factorization during the convolution, as we have seen, and also during the deconvolution processes.

Consider the common cause BN model shown in Figure 4.11. The fully specified model is shown in BN graph $A$. The posterior distributions for all nodes can be computed by way of the BFE convolution algorithm and we can cache any distributions and parameters we might need during this process, for subsequent use during deconvolution. The BFE deconvolution algorithm then proceeds in a simple fashion by eliminating all intermediate, frequency and severity variables until we get the reduced BN graph containing the final query set of interest. Thus in Figure 4.11 the final query set contains all variables in the common cause vector, $C$, and the aggregated total, $T$. The only significant difference between the approaches is that during convolution the DD algorithm is allowed to discretize freely in order to search for and discover the posterior high density regions, whereas during deconvolution, the same discretization as was produced during convolution is used. This approach results in a 'lower resolution' posterior but has the advantage of speed and reasonable accuracy.

Let us assume the model structure in BN $A$ of Figure 4.11. Here frequency, $N$, is discretized into three finite states $\{1, 2, 3\}$, so there are three *n-fold* convolution variables $T_i = \sum_{j=0}^{i} S_j, i = 0,1,2$ each corresponding to the sum of one, two and three severity variables. $T$ is the compound distribution defined by:

$$P(T=t)=a_0P(T_0=t_0)+a_1P(T_1=t_1)+a_2P(T_2=t_2), a_i=P(N=i), i=0,1,2$$

Given evidence $T=t_0$ the deconvolution of $C$ is achieved by:

$$P(C=c \mid T=t_0)=\frac{P(C=c,T=t_0)}{P(T=t_0)}$$
$$=\sum_{s_i,t_i,n} P(T=t_0 \mid pa\{T\})P(N=n)P(T_i=t_i \mid pa\{T_i\})P(S_i=s_i \mid C=c)P(C=c)$$

(4.16)

where $pa\{T\}$ denotes the parents of $T$.

So, once the convolution model has eliminated all irrelevant variables, in this case $S_i, T_{z_j}, E_j, F_j$ we would be left with the query set, which here is $Q=\{C,T\}$.

## 4.4.2 Reconstructing the Frequency Variable

If we are also interested in including the frequency variable, $N$, in our query set we must be careful to cache variables $E_j$, $F_{j-2}$ and $T_{z_j}$ during convolution. Recall that the prior distribution for $N$ was decomposed into the $E_j$ during compound density factorization, therefore we need some way of updating this prior using the new posterior probabilities generated on the Boolean variables, $E_j$, during deconvolution. To perform deconvolution on $N$ it is first necessarily to reconstruct $N$ from the $E_j$ variables that together composed the original $N$.

Reconstruction involves composing all Boolean variables, $E_j$, into the frequency variable $N$, in a way that the updating of $E_j$ can directly result in generating a new posterior distribution of $N$. The model is established by connecting all $E_j$ nodes to $N$, where the new NPT for $N$ has the form of combining all its parents. However, it turns out this NPT is exponential ($2^{j+1}$) in size. To avoid the drawback we use an alternative, factorized, approach that can reconstruct the NPT incrementally.

As before, we reconstruct $N$ using binary factorization where the conditioning is conducted efficiently using incremental steps. Here the intermediate variables produced during binary factorization, $N_k,(k=0,...,j-1)$, are created efficiently by ensuring their NPTs are of minimal size.

The routine for constructing the NPTs for $N_k,(k=0,...,j-1)$ from the $E_j$'s is:

1. Order parents $E_j$ and $E_{j-1}$ from higher index to lower index for $N_k$'s NPT (since $E_j$ is Boolean variable with only two states, one concatenating all $E_{j-1}$'s states and another state is single state that $E_{j-1}$ does not contain. In this example $E_1$ should be placed on top of $E_0$ in the NPT table, as it is easier for comparing the common sets)
2. As we have already generated the NPT map of $E_j$, $E_{j-1}$ and $N_k$. Next we specify the NPT entry with unit value ("1") at $N_k=\tau$, when $E_j$ and $E_{j-1}$ has common sets $\tau$ (In this example, e.g. $E_1$ and $E_0$ have common sets $\tau="0"$ and $\tau="1"$)
3. Specify NPT entry with value ("1") at $N_k=\tau$, when $E_j$ and $E_{j-1}$ has no common sets and $E_j=\tau$ ($E_j$ has one state $\tau$ that $E_{j-1}$ does not contain, so under this case $N_k$ only needs to be consistent with $E_j$ as the changes on $E_{j-1}$ won't affect the probability $P(N_k=\tau)$, E.g. in this example it is when $E_1=\tau="2"$)
4. Specify NPT entry with value ("0") at all other entries.

We repeat this routine for all $N_k,(k=0,...,j-1)$ until we have exhausted all $E_j$'s, producing a fully reconstructed $N$. Once we've built the reconstructed structure ($N_k$) for $N$, in fact the updates of $E_j$'s probabilities are directly mapping to $N_k$, and so deconvolution of $N$ is retrieved.

## 4.4.3 The BFE Deconvolution Algorithm with Examples

The BFE deconvolution algorithm, for *N-fold* deconvolution, is formalised, as pseudo code:

***Algorithm 3 BFE deconvolution algorithm***

**Input**: $S$ : Severity variable, $N$ : Frequency variable, $\mathbf{C}$ : vector of common causes and $T = t_0$

**Output**: posterior marginal of query set members i.e. $P(\mathbf{C} \mid T = t_0)$, $P(N \mid T = t_0)$

**Main**:

1. **do** convolution BFE algorithm to produce final query set
2. **if** $N$ is in query set
3. **reconstruct** $N$ from $E_j$
4. **end if**
5. **set** evidence $t_0$ on $T$ and perform inference
6. **return** posterior marginal distributions for query set

Algorithm 3 BFE deconvolution algorithm

**Example 4.3** deconvolutes frequency distribution:

Consider a simplified example for deconvoluting $N$, suppose frequency distribution $N$ is discretized as $\{0.1, 0.2, 0.3, 0.4\}$ with discrete states $\{0, 1, 2, 3\}$ and $S \sim Exponential(1)$. Figure 4.12 (a) shows these incremental steps for example 4.3. In this example there are three parents ( $E_0, E_1, E_2$ ) to $N$. The incremental composition steps of $E_j$ have introduced two intermediate variables $N_0$ and $N_1$, and we expect the frequency $N$ to be reconstructed at the end of the incremental step, which is variable $N_1$. Key to this process is how to build the NPT for each $N_k$.

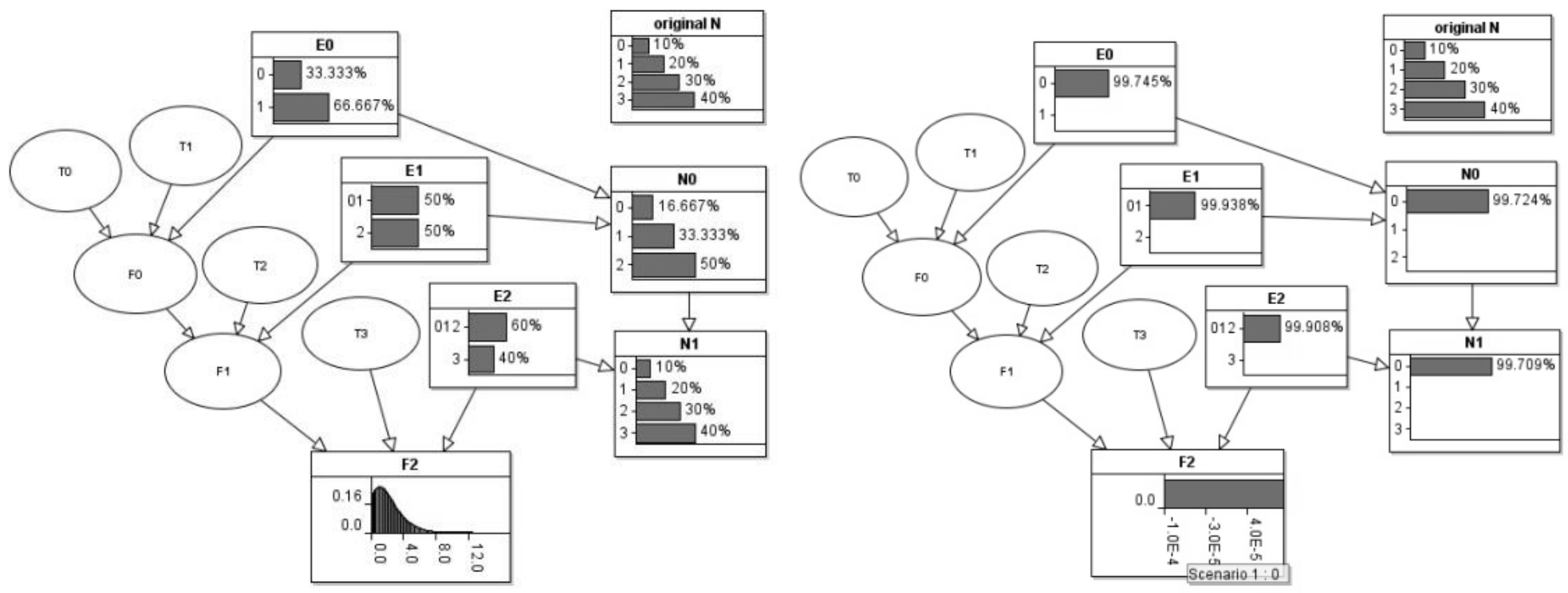

(a) (b)

Figure 4.12 (a) Reconstruct $N$ (b) Deconvoluting $N$

| $E_1$ | "01" | | "2" | |
|---|---|---|---|---|
| $E_0$ | "0" | "1" | "0" | "1" |
| $N_0 = 0$ | 1.0 | 0.0 | 0.0 | 0.0 |
| $N_0 = 1$ | 0.0 | 1.0 | 0.0 | 0.0 |
| $N_0 = 2$ | 0.0 | 0.0 | 1.0 | 1.0 |

Table 4.1 The NPT of $N_0$

Table 4.1 illustrates the NPT of $N_0$, where it composes $E_0$ and $E_1$ successively, in such a way that each $N_k$ contains all and only its parents' discrete states. So $N_0$ has the discrete distribution on "0", "1" and "2".

Figure 4.12 (b) shows the deconvolution of $N$ by our reconstruction process. The reconstructed prior distribution of $N_1$ is identical to node "original $N$" (shown in Figure 4.12 (a)) as we expected. After setting an observation value "0" at the compound sum variable $F_2$ we have queried that the posterior of $N$ is 99.7% probability at state "0", since at state zero it has all possibility of generating a zero compound sum at $F_2$.

The reconstruction algorithm is applicable to cases that $N$ has discrete parent cause variables as well, where $E_j$'s NPTs are generated directly from $N$'s parents, and the deconvolution is performed by BFE deconvolution algorithm.

**Example 4.4** de-convolutes common cause variables:

Consider another example for deconvoluting common cause variable as shown in the BN in Figure 4.13. The model has this form: Frequency distribution $N$ is discretized as $\{0.2, 0.3, 0.5\}$ with discrete values $\{1, 2, N\}$.

The common cause variable, $C$, has labelled values and probabilities $\{Normal = 0.8, High = 0.19, Extreme = 0.01\}$. There are three *i-fold* variables, one for each of the frequency states: $T_1_fold$, $T_2_fold$ and $T_N_fold$, which are all dependent on $C$. The compound density is given by variable $T$ which satisfies:

$$T = P(N = 1)P(T_1_\ fold) + P(N = 2)P(T_2_\ fold) + P(N = N)P(T_N_\ fold)$$

Figure 4.13 (a) shows the *N-fold* common cause model executed as a single BN using DD and JT algorithms alone (i.e. BFE convolution is not applied) and (b) shows the result from applying BFE convolution, where new intermediate variables are introduced by binary and compound density factorization. The marginal distributions for all variables are identical in (a) and (b).

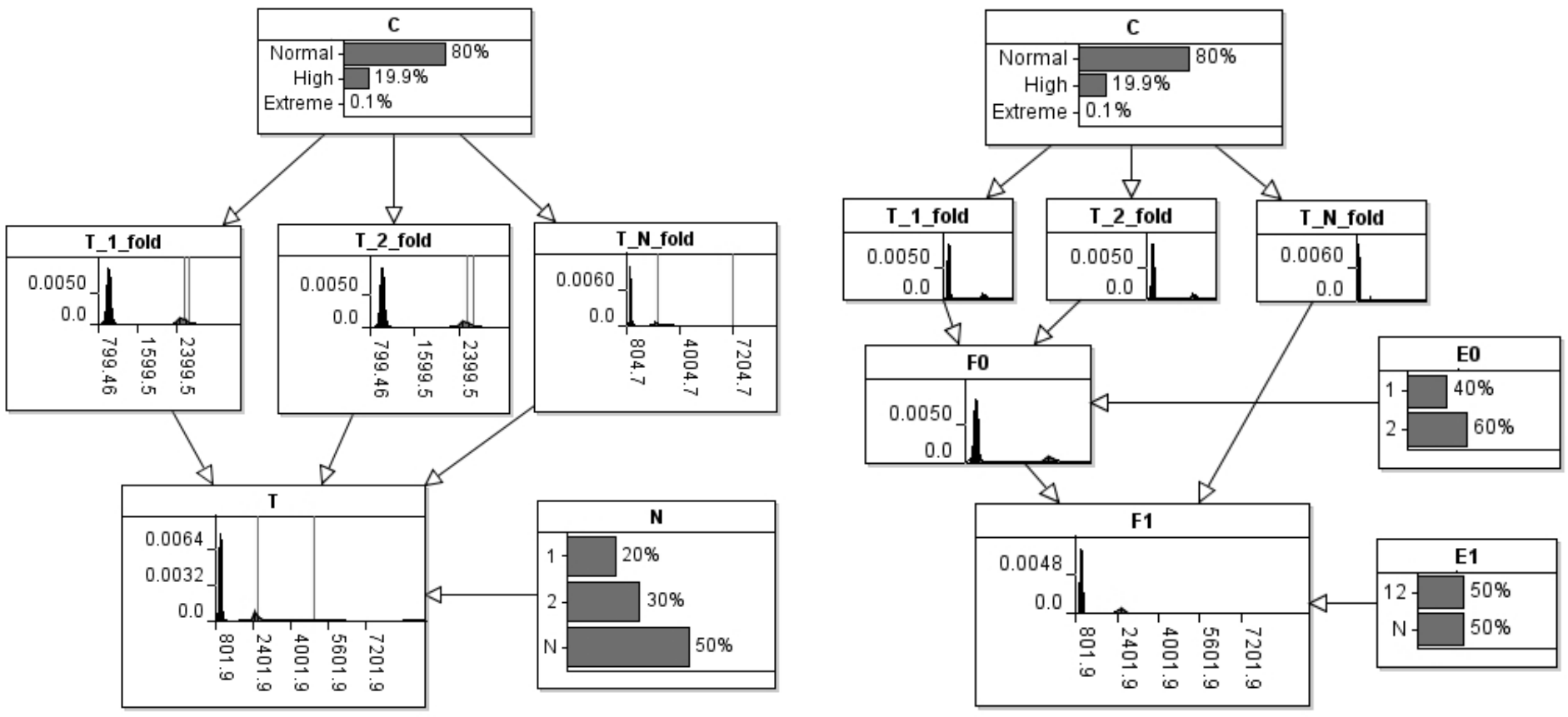

(a) (b)

Figure 4.13 (a) *N-fold* common cause model, (b) Binary factorized form of (a) with marginal distributions superimposed on the BN graph

We might be interested in the scenario where $T = 3000$ and wish to determine what the posterior probability is for the common cause given this observations. Figure 4.14 (a) shows the full BN, with evidence entered and Figure 4.14 (b) is the corresponding BN generated by BFE.

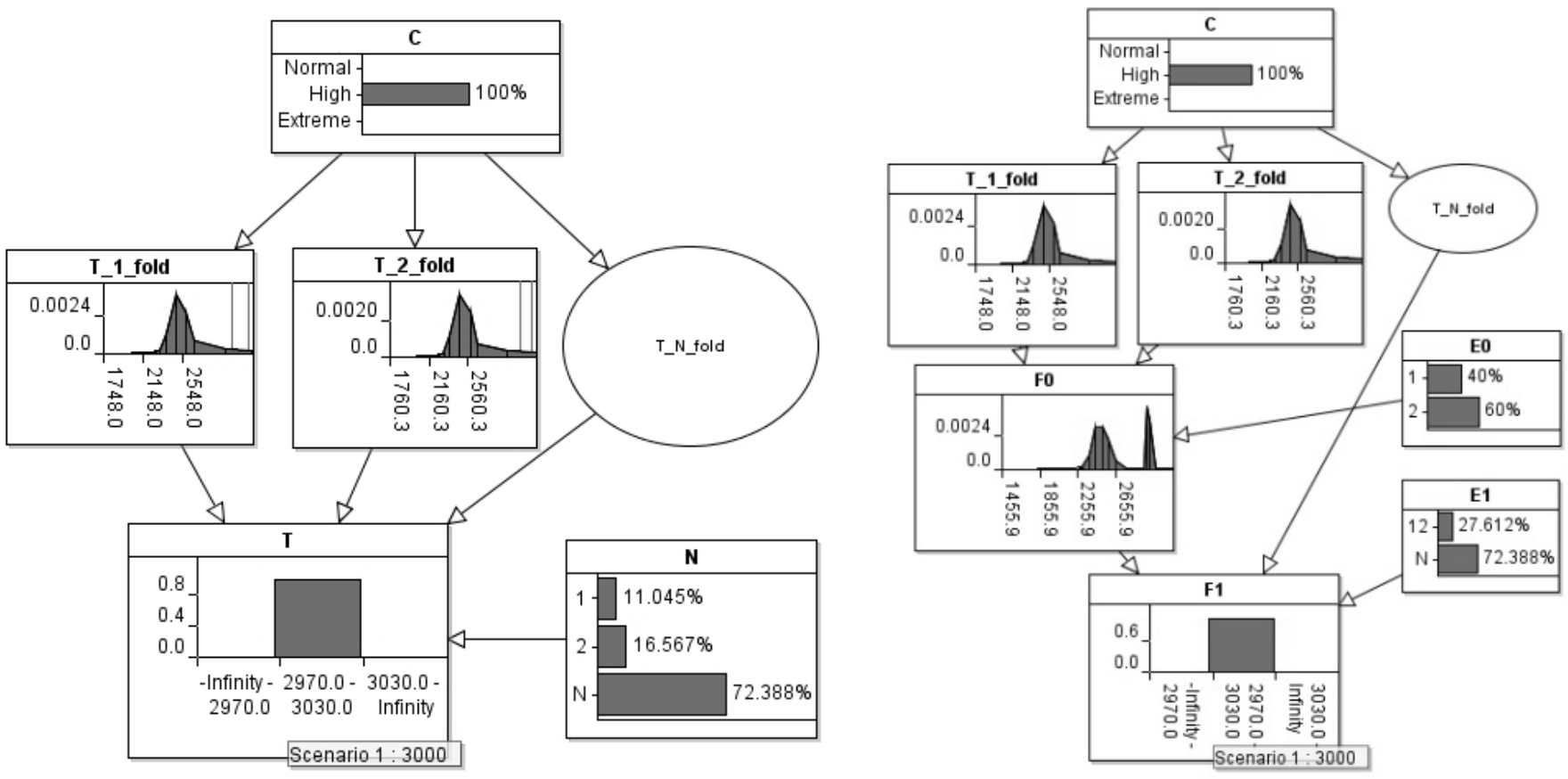

(a) (b)

Figure 4.14 (a) *N-fold* model deconvolution on full model; (b) *N-fold* model deconvolution using BFE algorithm

We can reconstruct the frequency $N$ from the $E_j$ in Figure 4.14 (b), to obtain the posterior distribution for $N \sim \{0.11, 0.166, 0.724\}$. The two models (a) and (b) produced identical posterior marginal for the query set $Q = \{C, N\}$ under BFE deconvolution.

A proof that our deconvolution algorithm produces the same result as a complete model is given in the Appendix – Proof B.

# 4.5. Experiments

We report on a number of experiments using the BFE algorithm in order to determine whether it can be applied to a spectrum of risk aggregation problem archetypes. Where possible the results are compared to analytical results, FFT, Panjer's approach and Monte Carlo (MC) simulation[7]. Throughout experiments 1-3, we use MC as a numerical reference point to determine whether the BFE algorithm's is sufficiently accurate since Panjer and FFT are not convenient or possible to calculate in some of the experiments.

The following experiments, with accompanying rationale, were carried out:

[7] In the experiments, we define the MC process as: first we draw samples $n_i$ from frequency distribution, and then draw $n_i$ samples from severity distribution and sum them. Finally we normalize all the samples to yield the compound sum.

1. *Experiment 1*: *Convolution with basic parameterisation*. This is a straightforward *N-fold* convolution, included here for easy comparison with prior art. The severity variables are assumed i.i.d.
2. *Experiment 2*: *Convolution with multi-modal (mixtures of) severity distribution*. We believe this to be a particularly difficult case for those methods that are more reliant on particular analytical assumptions. Practically, multi-modal distributions are of interest in cases where we might have extreme outcomes, such as sharp regime shifts in asset valuations.
3. *Experiment 3*: *Convolution with discrete common causes variables*. This is the key experiment in the paper since these causes will be co-dependent and the severity distribution will depend on their values (and hence will be a conditional mixture).
4. *Experiment 4*: *Deconvolution with discrete common causes*. This is the inverse of experiment 3 where we seek to estimate the posterior marginal for the common causes conditioned on some observed total aggregated value.

The computing environment settings for the experiments are as follows. Operation system: Windows XP Professional, Intel i5 @ 3.30GHz, 4.0GB RAM. AgenaRisk was used to implement the BFE algorithm, which was written in java (not generally recognised as a high performance language for numerical calculations), where typically the DD settings were for 65 iterations for severity variables and 25 iterations for the frequency variable. The reference algorithms were written in R (R, 2013) using the actuarial add-on package, actuar. For FFT the settings used were range $M = 0:95$ and span $h = 0.1$. Panjer settings: R, range $M = 0:95$ and span $h = 0.1$. A sample size of 2.0E+5 was used as the settings in R for the Monte Carlo simulation.

Comparing the speed and memory requirements of each reference algorithm can be deceptive given that solving each problem involves both the human analyst and the computer. Some algorithms require non-trivial front loaded investment of analytical effort to design and configure the exact form of the algorithm in advance of any computation (for instance the frequency variable has to be in Panjer's class for use in the Panjer algorithm (Embrechts & Frei, 2009). Others are more general purpose and simply involve the user declaring the model and setting some parameters with no mathematical preparation stage. Given this we do not compare simply choose the algorithm on computational speed but instead bear in mind the commensurate or compensating benefits of ease of use, personal productivity and ease of presentation to, and validation by, an end user.

In each experiment we compare the following summary statistics, for all aggregated distributions, for each algorithm: Mean, Standard deviation, 95$^{th}$ percentile and 99$^{th}$ percentile. We also give an indication of analytical preparation time needed $\{High, Medium, Low\}$ (of course this would include effort to formulate the model by hand and write any bespoke code needed to solve, or approximate it).

*Experiment 4.1: Convolution with basic parameterisation*

Consider the simple example model by that the frequency is $N \sim Possion(50)$ with severity $S \sim Exponential(1)$ distributions is computed.

| **Algorithm** | **Mean** | **Standard Deviation** | **95th Percentile** | **99th Percentile** | **Analysis Effort** |
|---|---|---|---|---|---|
| Exact | 50.0 | 10.0 | N/A | N/A | High |
| FFT | 49.4 | 9.9 | 66.4 | 73.5 | High |
| Panjer | 48.9 | 9.9 | 66.5 | 74.4 | High |
| MC | 49.8 | 9.8 | 66.6 | 75.4 | Low |
| BFE | 50.3 | 10.9 | 67.9 | 76.6 | Low |

Table 4.2 Results of convolution with basic parameterisation

Table 4.2 shows the accuracy of each algorithm. Clearly BFE is as accurate as other approaches. However the preparation and analysis time is commensurate with using Monte Carlo simulation.

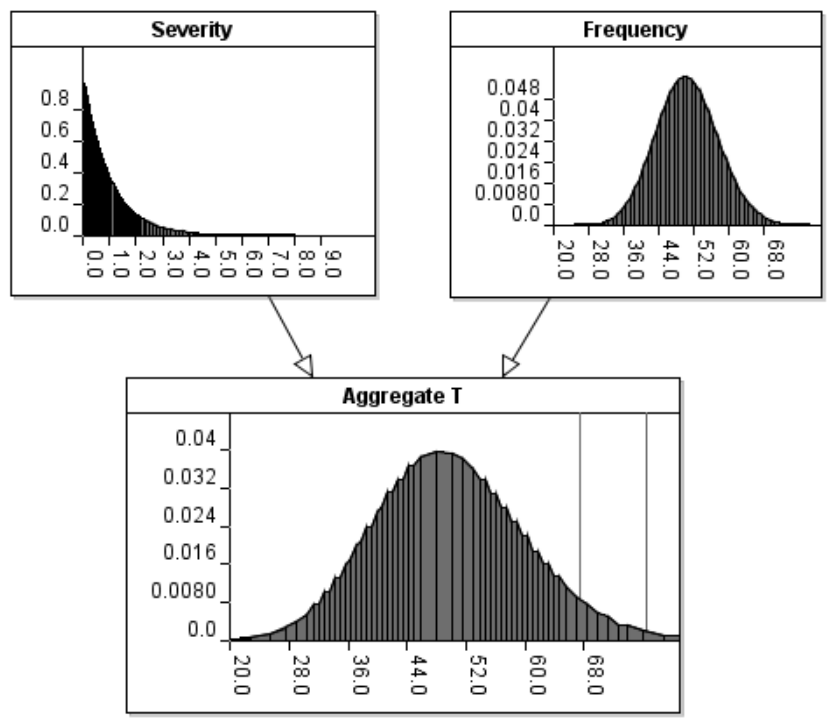

Figure 4.15 Marginal distributions for overlaid on BN graph containing query nodes for experiment 4.1

The corresponding marginal distribution for the query node set $\{T, N, S\}$ is shown in Figure 4.15.

*Experiment 4.2: Convolution with multi-modal severity distribution*

Here we set the event frequency as $N \sim Poisson(50)$ but the severity distribution is a mixture distribution, $S \sim f_S$:

$$f_S = (0.2)Gamma(5, 1.5) + (0.3)Normal(25, 2) + (0.4)Normal(50, 3) + (0.1)Gamma(100, 2)$$

In a hybrid BN a mixture distribution is modelled by conditioning the severity variable on one or more partitioning discrete variables, $C$. Assuming that that severity variables, $S_j$, are i.i.d. we can calculate the compound density using BFE.

The characteristic function of a mixture distribution is inconvenient to define (with continuous and discrete components). The analytical and programming effort needed to solve each multi-modal severity distribution for Panjer is high, so here we compare with MC only.

| **Algorithm** | **Mean** | **Standard Deviation** | **95th Percentile** | **99th Percentile** | **Analysis Effort** |
|---|---|---|---|---|---|
| MC | 2444.8 | 516.7 | 3340.0 | 3787.7 | Low |
| BFE | 2441.1 | 523.3 | 3341.5 | 3783.1 | Low |

Table 4.3 Results of convolution with multi-modal severity distribution

The corresponding marginal distribution for the query node set $\{T, N, S\}$ is shown in Figure 4.16.

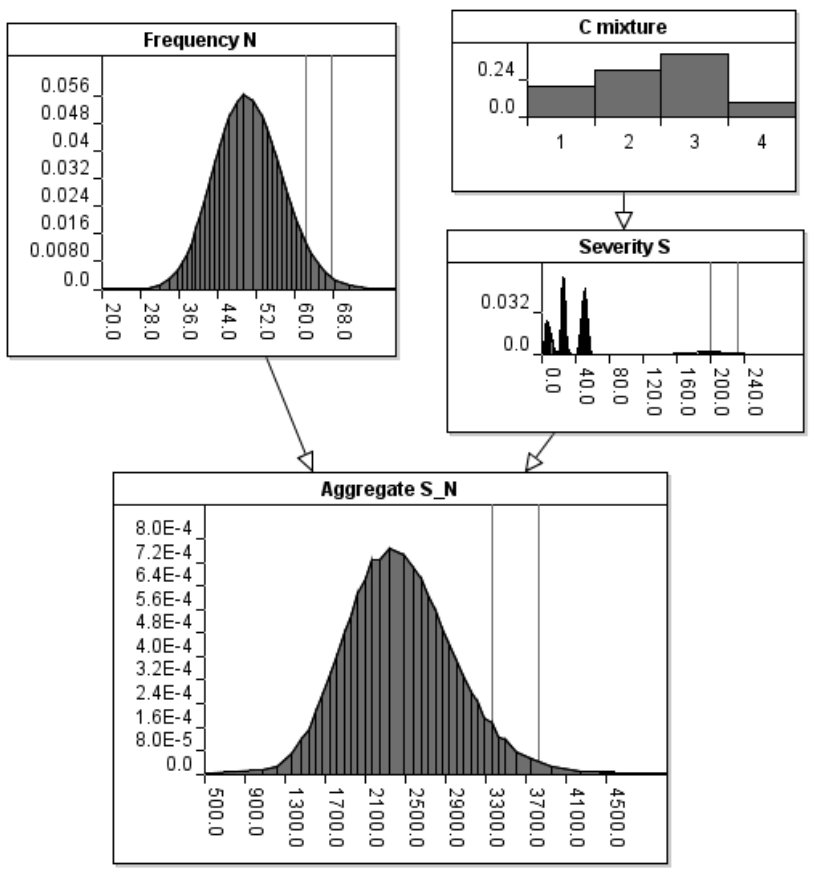

Figure 4.16 Marginal distributions for overlaid on BN graph containing query nodes for experiment 4.2

*Experiment 4.3: Convolution with discrete common causes variables*

Loss distributions from operational risk can vary in different circumstances, e.g. exhibiting co-dependences among causes. Suppose in some cases that losses are caused by daily operations and these losses are drawn from a mixture of truncated Normal distributions, whereas extreme or some unexpected losses are distributed in a more severe distribution. We model this behaviour by a hierarchical common cause combination $C_0,...,C_4$.

The severity variable $S$ is conditioning on common cause variable, $C_0, C_1, C_2$. And these common cause variables are conditioned on higher common causes $C_3$ and $C_4$. Severity NPT is shown in Table 4.4. The frequency distribution of losses is modelled as $N \sim Poisson(50)$.

| $C_0$ | High | | | | Low | | | |
|---|---|---|---|---|---|---|---|---|
| $C_1$ | High | | Low | | High | | Low | |
| $C_2$ | High | Low | High | Low | High | Low | High | Low |
| Expression | Normal (1,2) | Normal (2,3) | Normal (3,4) | Normal (4,5) | Normal (100,110) | Normal (110,120) | Normal (120,130) | Normal (130,140) |

Table 4.4 Severity NPT

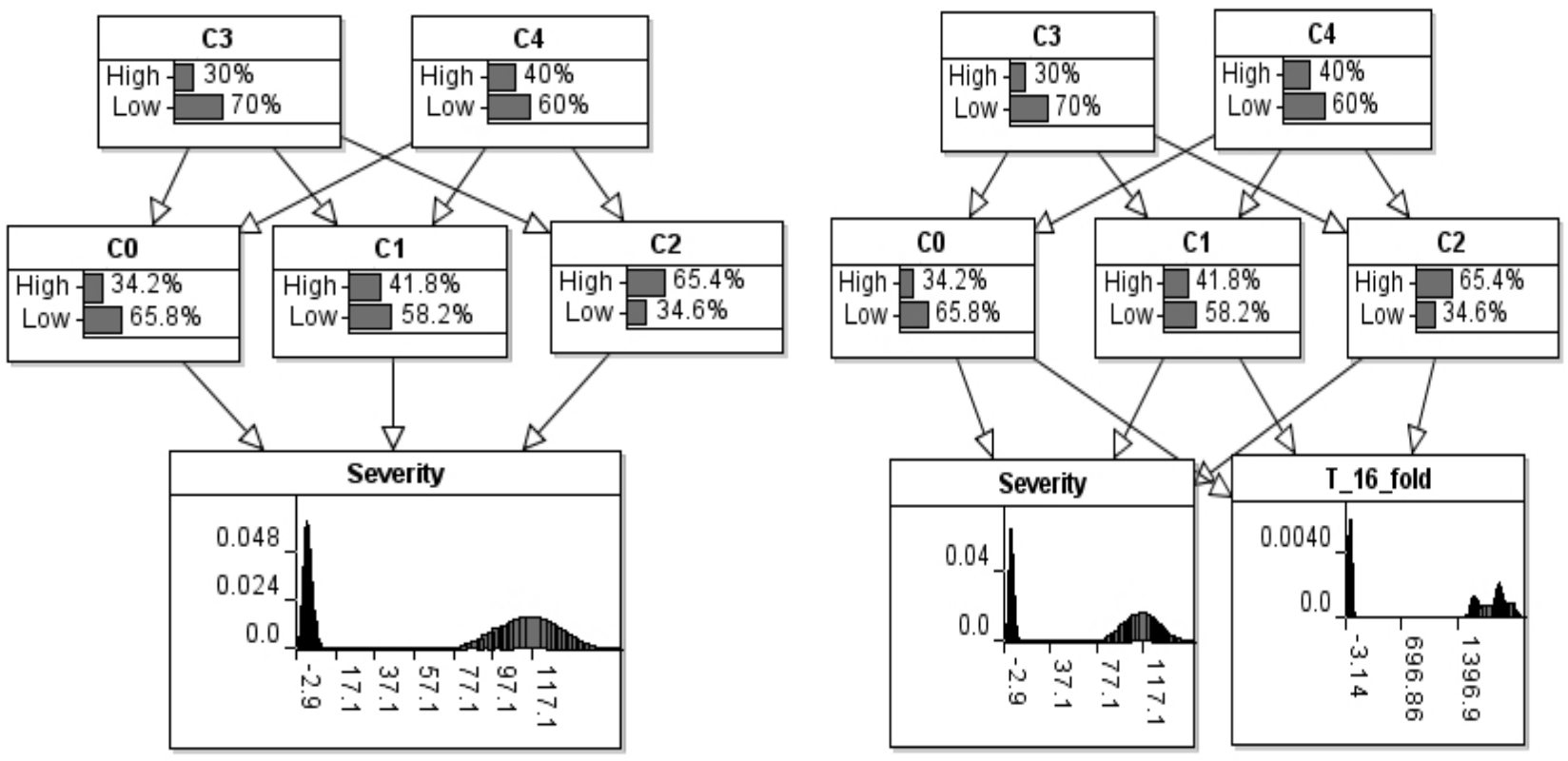

(a) (b)

Figure 4.17 (a) Common cause dependent severity; (b) *16-fold* convolution of dependent severity

In Figure 4.17 (a) the model severities with dependencies by common cause variables $C_0,...,C_4$ is introduced. Figure 4.17 (b) depicts a *16-fold* convolution of dependent severities using the variable elimination method. For any given frequency distribution, $N$ , we can apply the BFE convolution algorithm to calculate the common cause *N-fold* convolution.

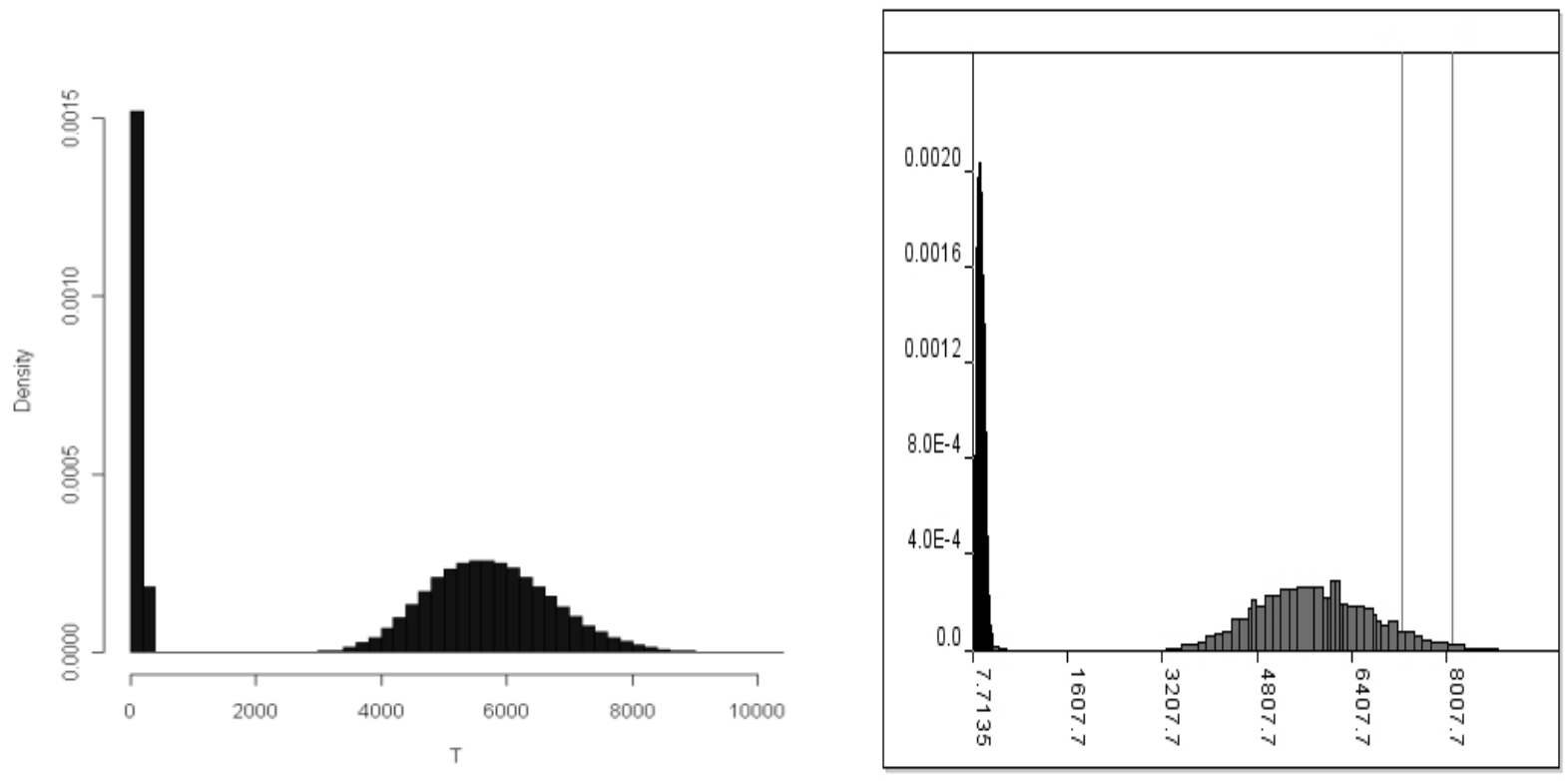

(a) (b)

Figure 4.18 Compound densities (a) MC; (b) BFE

| **Algorithm** | **Mean** | **Median** | **Standard Deviation** | **95th Percentile** | **99th Percentile** |
|---|---|---|---|---|---|
| MC | 3831 | 5017 | 2784 | 7215 | 8023 |
| BFE | 3871 | 5052 | 3255 | 7267 | 8115 |

Table 4.5 Common cause *N-fold* convolution density

Figure 4.18 illustrates the output compound densities for the compared algorithms. Table 4.5 shows the results for the two approaches are almost identical on summary statistics except the small difference on standard deviation. BFE has offered a unified approach to construct and compute such a model conveniently.

*Experiment 4.4: deconvolution with discrete common causes variables*

We reuse the convolution model from experiment 4.3 as the input model for deconvolution (Figure 4.19).

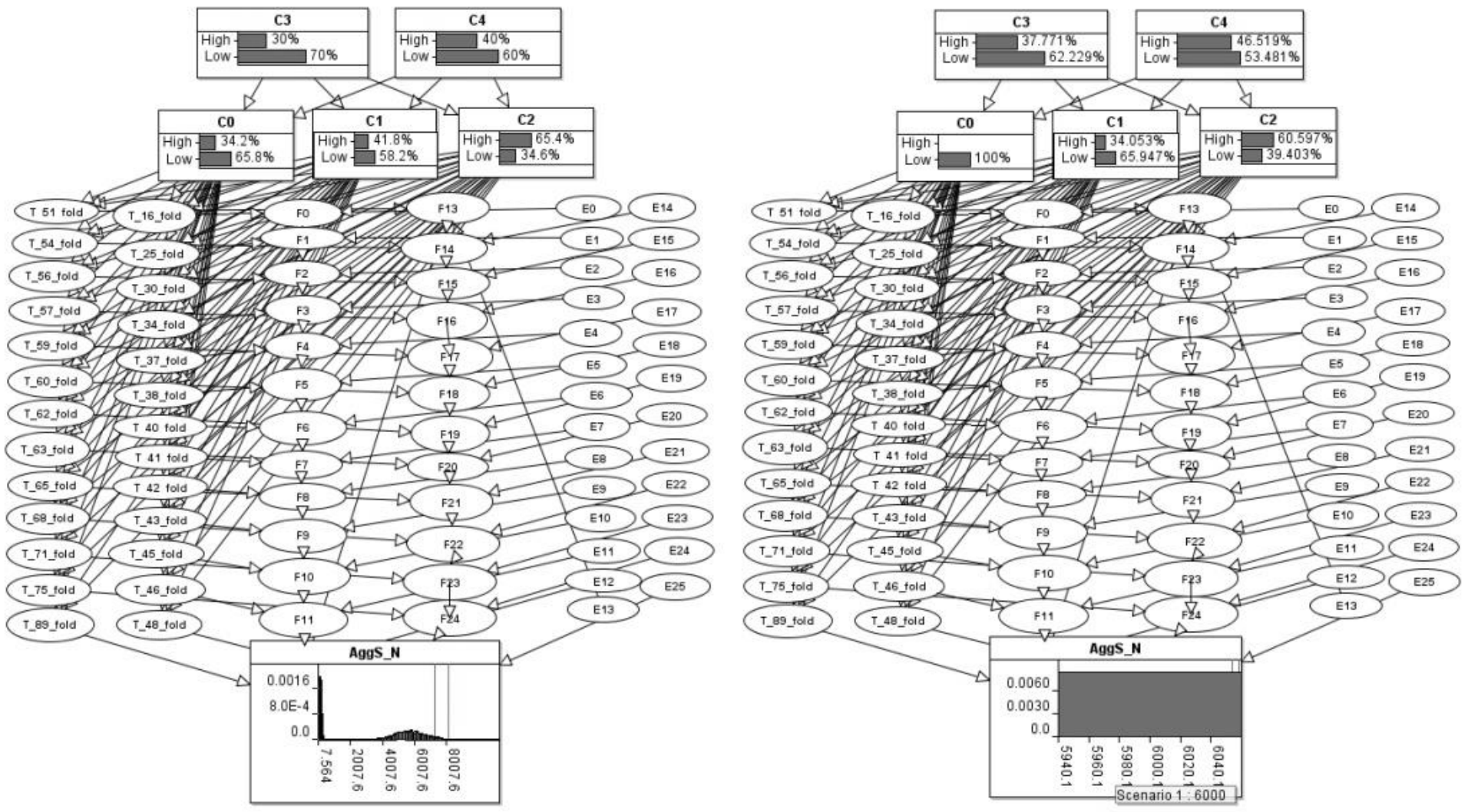

(a) (b)

Figure 4.19 (a) Common Cause *N-fold* convolution using BFE algorithm; (b) *N-fold* model deconvolution using BFE algorithm

In Figure 4.19, the intermediate variables in this example are shown for reference despite them being eliminated during the convolution process.

Figure 4.19 (b) sets an observation on total aggregation node $AggS_N$. After performing deconvolution we queried the posterior marginal of common causes and diagnose that the most likely common cause is $C_0$, which is in its “Low” state with certainty. This is easily explained by the fact that from the severity NPT, shown in Table 4.4, it is only when state of $C_0$ is “Low” that a value of 6000 can be at all probable. This deconvolution is currently only supported by BEF since the information cannot be back retrieved by other approaches.

Deconvolution is obviously useful in carrying out a sensitivity of the model results, allowing the analyst to quickly check model assumptions and identify which causal factors have the largest consequential effect on the result. This is difficult to do manually or informally in the presence of non-linear interactions. Also, without "backwards" deconvolution we can only compute such sensitivities "forwards" one casual variable at a time and this is computationally much more expensive. For example, the forwards calculation of $T$ from ten Boolean common cause variables would require $2^{10}$ calculations versus 40 in the backwards case (assuming $T$ was discretized into 40 states).

## 4.6. Summary

This chapter has reviewed historical, popular, methods for performing risk aggregation and compared them with a new method called *Bayesian Factorization and Elimination* (BFE). The method exploits a number of advances from the field of Bayesian Networks, covering methods to approximate statistical and conditionally deterministic functions and to factorize multivariate distributions for efficient computation. The objective for BFE was for it to perform aggregation for classes of problems that the existing methods cannot solve (namely hybrid situations involving common causes) while performing at least as well on conventional aggregation problems. The experiments show that our objectives were achieved. For more difficult hybrid problems the experimental results show that BFE provides a more general solution than is possible or convenient to produce with the previous methods. For example, BFE outperforms the Panjer and FFT in hybrid cases. MC sampling techniques are tailored to specific cases; thus a general solution may be difficult. MCMC, however, may be generally applicable, such as a Metropolis-Hasting sampler (Metropolis, et al, 1953) (Hastings, 1970) or Gelman's Stan toolbox (Matthew, Carpenter, & Gelman, 2012), but a general MCMC sampler may still perform poorly on a problem without bespoke design or parameter adjustment. BFE does not require specific adjustment on experimental settings except where certain problems may require an increase in discretization resolution. MC is also not suitable for constructing a deconvolution model from the convolution model. In contrast BFE provides a single unified procedure for performing Bayesian convolution, and also a convenient way to perform deconvolution or model reconstruction.

The BFE approach can be easily extended to perform deconvolution for the purposes of stress testing and sensitivity analysis in a way that competing methods cannot currently offer. The BFE deconvolution method reported here provides a low resolution result, which is likely good enough for the purposes of model checking and sensitivity analysis. However, we are investigating an alternative high resolution approach whereby variables are discretized efficiently during the deconvolution process, thus providing more accurate posterior results.

With regard to the research hypotheses outlines in Chapter 1, Section 1.2 the research results presented here positively satisfy the first three hypotheses.

# 5. Inference for High Dimensional Models

To perform Bayesian risk aggregation when there are many aggregated variables is very challenging. This includes the representation challenges and inference challenges. One general approach to model general dependencies is by using copulas, i.e. multivariate distributions are converted into local copula parameterizations, and there exists a variety of construction mechanics using copula building blocks. Under the Bayesian framework, however, the representation requires further decomposition of the existing functions into conditional forms. One way to model an arbitrary multivariate distributions using BN maybe by seeking a combination of copula and BN, e.g. copula Bayesian Network (CBN) (Elidan, 2010), where the joint multivariate distribution can be decomposed by copula functions, and in further can be factorized into conditional forms.

If a BN representation problem can be resolved, i.e. via CBN or general decomposition approach, a DCCD structure will be sufficient with respect to the graphical representation, as discussed before, the DCCD is the unique structure for representing an arbitrary distribution. One of the classic representations of DCCD models are the conditional Gaussian (CG) models, which have been used to factorize high dimensional multivariate Gaussian distributions (MGD) (McNeil et al., 2010) into CG forms. In this way an MGD model can be always represented by CG-DCCD model (discussed in section 5.1).

The inference is very challenging for a DCCD model: the analytical solutions are usually intractable. The stochastic simulation based algorithms are flexible to use, but is problem tailored and the convergence for arbitrary models is not guranteed. The standard discrete approximation approach, i.e. DDJT, can easily become computational intractable along with the rapid growth of the number of discrete space caused by high dimensions. GBP is flexible to use and can achieve accurate result if converged. However, the GBP algorithm has rarely applied to DCCD models, although there exists some research for Gaussian belief propagation solving continuous Gaussian linear models (Bickson et al., 2008) (Shental et al., 2008), these works are analytical and particularly focus on Gaussian models. General purpose inference for GBP in discrete forms requires a sophisticated clustering procedure to convert a factor graph into region graph, where a good construction of such region graph is difficult to find.

This chapter targets the inference problem of DCCD models, explores discrete approximation extending the use of discretization and approximation inference.

It develops a *triplet region construction* (TRC) algorithm based on the clustering method in GBP in discrete form, by constructing an optimal region graph and also satisfies the maxent-normal constraint (Yedidia et al., 2005). The TRC algorithm uses binary factorization algorithm to reduce the computation complexity introduced by deterministic function of parent-child NPTs in DCCD models. The size of the factors generated for region graph is then reduced. Then belief propagation is performed on TRC, compared to JT, the efficiency is achieved by replacing JT based cliques into smaller triplet clusters.

It then combines TRC based belief propagation with dynamic discretization to propose a *dynamically discretized belief propagation* (DDBP) algorithm. In such a way, inference can be carried out for general purpose, with continuous variables being discretized. We use conditional Gaussian DCCD models to show the accuracy of our algorithm, since such model has a simple analytical solution and can be used to validate the algorithms. The result of our experiment is close to the analytical solution. All experiments are implemented onto Bayesian software AgenaRisk (AgenaRisk, 2014).

# 5.1. Conditional Gaussian DCCD Model

In practice a BN with continuous CPDs (i.e. conditional Gaussian, continuous conditional non-Gaussian) are easy to convert to DCCD models, since the CPDs are deterministic likelihood functions that are easy to modify (when edges are added) using arithmetic operations. Usually discrete and hybrid models will present some difficulty because their CPDs are not conditionally deterministic, although it is feasible to use methods to ensure they can be factorized.

Let us focus on the common special case where we have a complete graph all of whose nodes correspond to Gaussian distributions. It is well known that Conditional

Gaussian (CG) models can be used to factorize MGDs. So an MGD model can be always represented by a DCCD model for a BN structure. Inference for such a model can be approximate or may use exact methods (S. L. Lauritzen, 1992) (this is not restricted to continuous but is also applicable to hybrid CG models). Our experiments in Section 5.4 are carried out using MGD models, as they have an analytical solution and so is a good basis for comparison (but note that the DDBP algorithm is designed to be general purpose and is not exclusively designed for use on MGD models).

To show how we decompose a complete MGD model consider an MGD vector $\mathbf{X} = \{X_1, ..., X_n\}$, where each univariate component is a Gaussian distribution and the pairwise correlations are encoded in a correlation matrix. Given such a vector $\mathbf{X}$, there always exists a partition $\mathbf{X} = \{\mathbf{X_1}, \mathbf{X_2}\}$, where vector $\mathbf{X_1} = \{X_1, ..., X_k\}$ and $\mathbf{X_2} = \{X_{k+1}, ..., X_n\}$, if the mean vector and covariance matrix for $\mathbf{X_1}$ and $\mathbf{X_2}$ are respectively:

$$\boldsymbol{\mu} = \begin{pmatrix} \boldsymbol{\mu}_1 \\ \boldsymbol{\mu}_2 \end{pmatrix}, \boldsymbol{\Sigma} = \begin{pmatrix} \boldsymbol{\Sigma}_{11} & \boldsymbol{\Sigma}_{12} \\ \boldsymbol{\Sigma}_{21} & \boldsymbol{\Sigma}_{22} \end{pmatrix}$$

Then $\mathbf{X_1} \sim N(\boldsymbol{\mu}_1, \boldsymbol{\Sigma}_{11})$ and $\mathbf{X_2} \sim N(\boldsymbol{\mu}_2, \boldsymbol{\Sigma}_{22})$ are also MGD vectors. Assuming that each $\boldsymbol{\Sigma}$ is positive definite, the conditional distributions of $\mathbf{X_2}$ given $\mathbf{X_1}$ may also be shown to be multivariate Gaussian, i.e., $\mathbf{X_2} \mid \mathbf{X_1} \sim N(\boldsymbol{\mu}_{2|1}, \boldsymbol{\Sigma}_{22|1})$, as in Equation 5.1, where

$$\boldsymbol{\mu}_{2|1} = \boldsymbol{\mu}_2 + \boldsymbol{\Sigma}_{21}\boldsymbol{\Sigma}_{11}^{-1}(\mathbf{X}_1 - \boldsymbol{\mu}_1) \text{ and } \boldsymbol{\Sigma}_{22|1} = \boldsymbol{\Sigma}_{22} - \boldsymbol{\Sigma}_{21}\boldsymbol{\Sigma}_{11}^{-1}\boldsymbol{\Sigma}_{12} \qquad (5.1)$$

is the conditional mean vector and covariance matrix respectively. More generally, the partition can be sequenced arbitrarily. Each univariate variable can be conditioned on its antecedent variables, thus resulting in the conditional representation of a multivariate distribution.

The MGD vector $\mathbf{X}$ can be decomposed to a CG-DCCD where there are interdependences between every pair of variables in $\{X_1, ..., X_n\}$ (i.e. the covariance between the pairs). An arbitrary variable $X_j$ is conditioned on all its antecedents $X_1,...,X_{j-1}$ but not directly conditioned on its descendants. We can then apply Equation 5.1 to express the decomposition of general multivariate Gaussian models.

**Example 5.1**

Consider an MGD vector $\mathbf{X}=\{X_1,X_2,X_3,X_4,X_5,X_6\}$ with mean vector and covariance matrix respectively:

$$\boldsymbol{\mu}=\begin{pmatrix}2\\3\\4\\5\\6\\7\end{pmatrix},\quad \boldsymbol{\Sigma}=\begin{pmatrix}4.0 & 0.6 & 0.8 & 1.0 & 1.2 & 1.4\\0.6 & 9.0 & 1.2 & 1.5 & 1.8 & 2.1\\0.8 & 1.2 & 16 & 2.0 & 2.4 & 2.8\\1.0 & 1.5 & 2.0 & 25 & 3.0 & 3.5\\1.2 & 1.8 & 2.4 & 3.0 & 36 & 4.2\\1.4 & 2.1 & 2.8 & 3.5 & 4.2 & 49\end{pmatrix}$$

After iteratively applying Equation 5.1, (McNeil et al., 2010) , the CPD expression for each univariate variable is given by:

$$\begin{aligned}
&f(X_1)\sim N(2,4)\\
&f(X_2\,|\,X_1)\sim N(2.7+.15X_1,8.91)\\
&f(X_3\,|\,X_1,X_2)\sim N(3.273+.18X_1+.12X_2,15.7)\\
&f(X_4\,|\,X_1,X_2,X_3)\sim N(3.75+.2X_1+.14X_2+.1X_3,24.375)\\
&f(X_5\,|\,X_1,X_2,X_3,X_4)\sim N(4.15+.23X_1+.15X_2+.12X_3+.09X_4,34.89)\\
&f(X_6\,|\,X_1,X_2,X_3,X_4,X_5)\sim N(4.5+.25X_1+.17X_2+.13X_3+.1X_4+.08X_5,47.25)
\end{aligned}$$

Given these CPDs for each univariate variable the equivalent BN graph for representing a MGD is shown in Figure 5.1 $G$. We see the factorization of the MGD is a six dimensional CG-DCCD model.

Figure 5.1 $G$ shows the actual BN model we have built using the expressions we derived, which is a six dimensional CG-DCCD model. As we have introduced the BF process for a DCCD model in section 2.4, we also undergo a BF process for all CG-DCCD models to reduce the NPT size. Figure 5.1 $G'$ shows the BF process for the six dimensional CG-DCCD model.

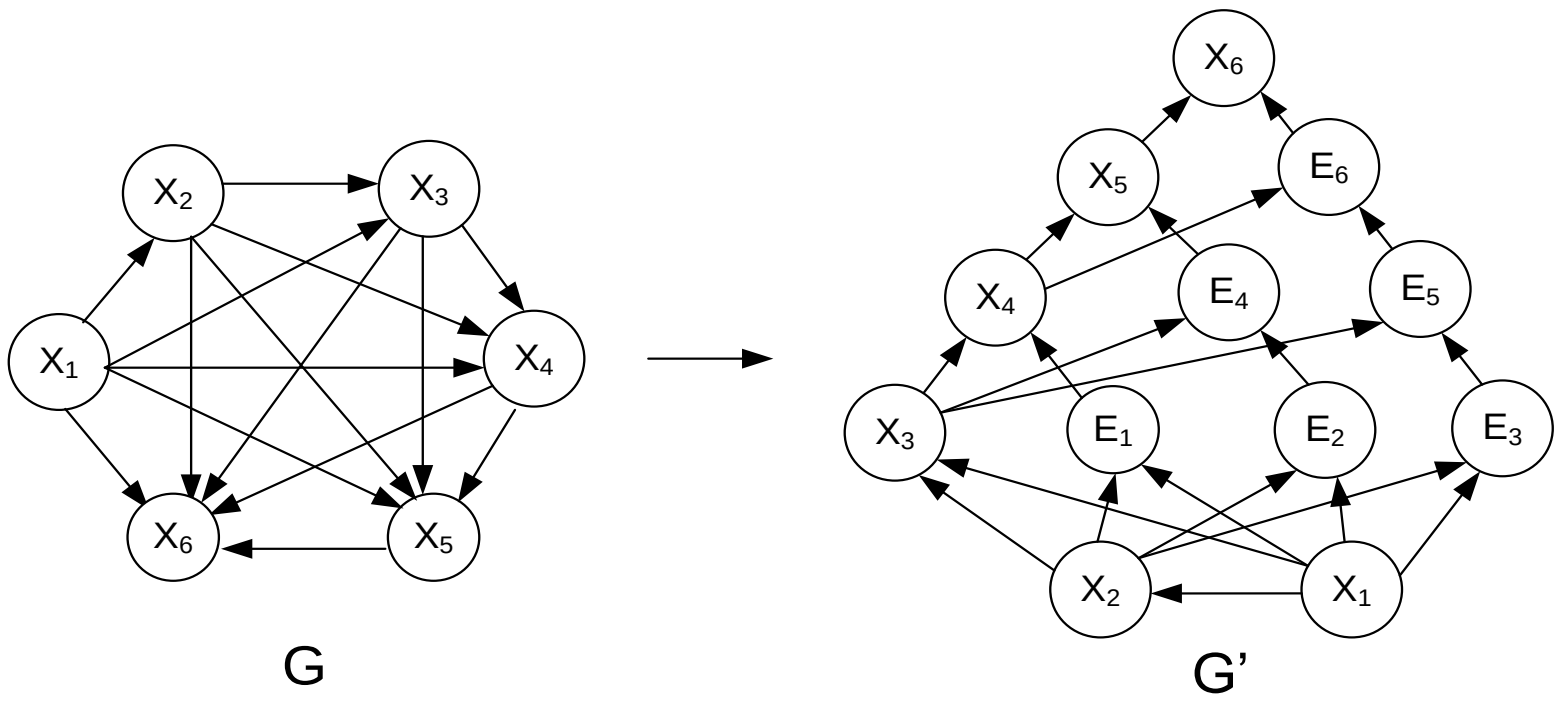

Figure 5.1 Six dimensional CG-DCCD to $\kappa_6$ full BFG

# 5.2. The Triplet Region Construction Algorithm

This section develops the *Triplet Region Construction* (TRC) algorithm along with the associated subsidiary algorithms introduced thus far. It automatically constructs an optimized region graph using a $\kappa_n$ dimensional full-BFG as input.

We have already reviewed Generalized Belief Propagation (GBP) and discussed a particular approach called the Cluster Variation Method (CVM) in chapter 3, which produces, as output, an object called a region graph, which can be used for inference.

In section 5.2.1 we outline the desirable properties of this region graph that we need to preserve in our new TRC algorithm. This involves identifying two levels of regions containing primary triplets in the first region and interaction triplets in the second region. These interaction triplets are then pruned to ensure that balance and the Maximum Entropy Normal property are maintained. Section 5.2.2 offers full definitions of all algorithmic steps. Proofs and demonstrations are provided throughout.

## 5.2.1 Identifying the TRC regions

In Chapter 3 we have introduced the region graph choice and interaction strength that are key components for good approximations. To identify TRC regions, our algorithm is similar to Welling's (Welling, 2004), where we partition the full-BFG into triplets to model interactions. The partitions chosen need to be sufficient to guarantee messages are passed between neighbouring triplets in order to minimize computational payload. However the extent to which the model sufficiently accounts for the strength of interactions can only be determined empirically.

From (Yedidia et al., 2005) and Welling et al. (Welling et al., 2005) (Welling, 2004) (Gelfand & Welling, 2012) we identify three properties necessary for guaranteeing the best approximation under GBP:

*Property 1: Acyclic* – the region graph is acyclic and contains two levels.

*Property 2: Balanced* – the region graph contains variables that are counted exactly once (i.e. the counting number for each variable is one).

*Property 3: Maximum Entropy Normal* – the maxent-normal property is met when the region based entropy is maximised when all beliefs are uniform.

From free energy theory (Yedidia et al., 2005), the free energy of the region graph achieves its minimum when the beliefs, $b$, derived from the region graph is equal to the joint probability distribution, $p$, or, equivalently, the entropy of the region graph achieves its maximum when all region beliefs are uniform (i.e. have the same non-informative value). This maxent-normal property is a necessary, but not sufficient, condition for a good approximation.

An acyclic region graph is preferable because it helps ensure convergence of messages (Yedidia et al., 2005). In our case an increase in model dimensionality would lead to an exponential increase in the amount of message passing, making message scheduling difficult and sensitive to message order. Many popular message passing algorithms, such as the tree reweighted max-product algorithm (Kolmogorov, 2006) and loopy BP (Murphy et al., 1999), do not guarantee convergence (Meltzer, Globerson, & Weiss, 2009).

The full-BFG graph $G'$ topology has the following advantages in inference:

1. Homogeneous: The two level acyclic region graph structure is maintained irrespective of the number of model dimensions.
2. Uniform factor size: except for the root NPT $p(X_1)$ and single parent NPT $p(X_2 \mid X_1)$, all other NPTs are defined on *triplets* (a group of three variables containing a child with two parents, or parent with two children). For each triplet there is always an associated factor in the factor graph and the NPTs can be multiplied into the triplet.

The region graph can be constructed from the following components (Figure 5.2 shows these superimposed on an undirected full-BFG):

- *Primary triplet*: a triplet with an NPT defined by the BFG, $G'$, i.e. a child variable and its two parents (original and intermediate).
- *Moral edge*: an undirected edge connecting the parent nodes of each primary triplet; this links an original variable and an intermediate variable.
- *Interaction triplet*: a triplet used to interact with the primary triplet through a moral edge.

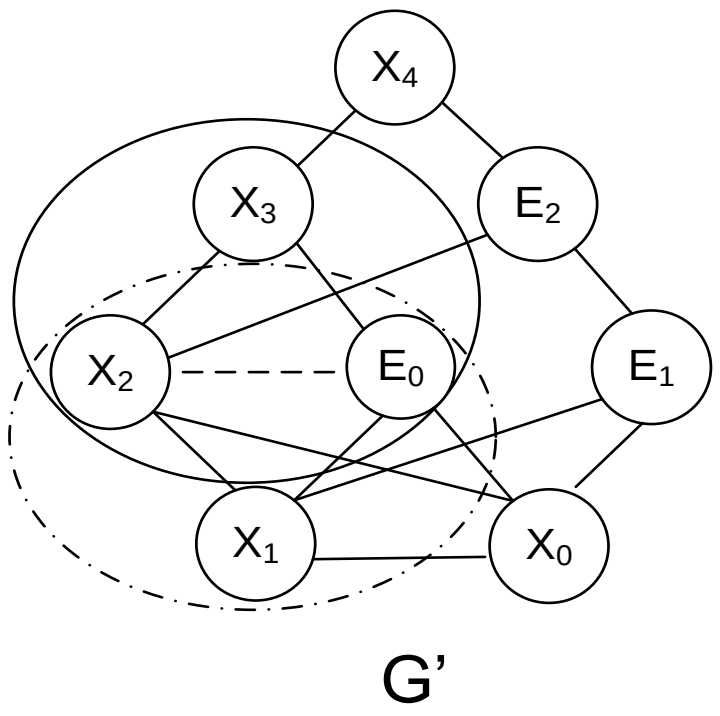

Figure 5.2 Topology of a $\kappa_5$ dimensional BFG, with solid circle a primary triplet, dashed line a moral edge and dashed circle an interaction triplet

Primary triplets are already determined by a full-BFG, since it includes child nodes with their parents. Note that there exists one "root" primary triplet $X_1X_2X_3$, as shown in Figure 5.2, where the root contains three original variables and no intermediate variables. This is explained by the fact that there is one parent-less root node in the BN.

However, for these primary triplets to communicate they need to send messages via the interaction triplets, and the number and composition of these triplets determines the strength of interaction (and the amount of dependence the approximation supports).

The only open decision then is the choice of interaction triplets and to determine the interaction triplets we can exploit the following properties:

1. Each moral edge always connects an original variable, $X_i$ , and an intermediate variable, $E_t$ , where the intermediate variable is always a common member in two primary triplets.
2. A candidate interaction triplet is always composed from the two variables connecting a moral edge (original and intermediate) and one original parent variables, $X_i$ , of either of these two variables.

When presented by a choice of interaction triplets to select we retain the one that ensures the counting number for all of the original variables, $X_i$, is balanced, and discard the others. This guarantees the “Balanced” property, and the resulting region graph is therefore balanced.

An example of the process is given in Example 5.2.

**Example 5.2**

Consider the moral graph of a $\kappa_5$ BFG $G'$ in Figure 5.3, with edge directions maintained to help identify the primary triplets. The moral edges are represented as dashed lines. All primary triplets and moral edges are listed in the table placed aside $G'$ in Figure 5.3. The root primary triplet is $X_1X_2X_3$ .

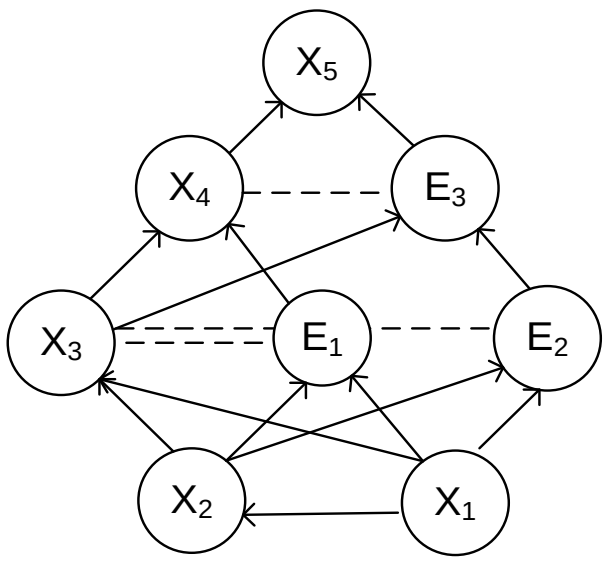

G’

| Primary Triplets | Moral Edges |
|---|---|
| $X_1X_2X_3$ | $(X_4, E_3)$ |
| $X_1X_2E_1$ | $(X_3, E_1)$ |
| $X_1X_2E_2$ | $(X_3, E_2)$ |
| $X_3E_1X_4$ | |
| $X_3E_2E_3$ | |
| $X_4E_3X_5$ | |

Figure 5.3 Identifying primary triplets using moral graph of a $\kappa_5$ full-BFG

We can identify all of the interaction triplets in Figure 5.4 $G''$ as follows. Let us first consider moral edge $(X_4, E_3)$ . The parent variables associated with variables sharing this moral edge are $X_3$ , $E_1$ and $E_2$ . The triplets we could produce by combining the parents with the moral variables, $(X_4, E_3)$, are $X_3X_4E_3$, $X_3E_1E_3$ and $X_3E_3E_2$. However, triplets $X_3E_1E_3$ and $X_3E_3E_2$ would be invalid since they do not contain any original parent variables of $(X_4, E_3)$. Therefore the interaction triplet for moral edge $(X_4, E_3)$ is $X_3X_4E_3$.

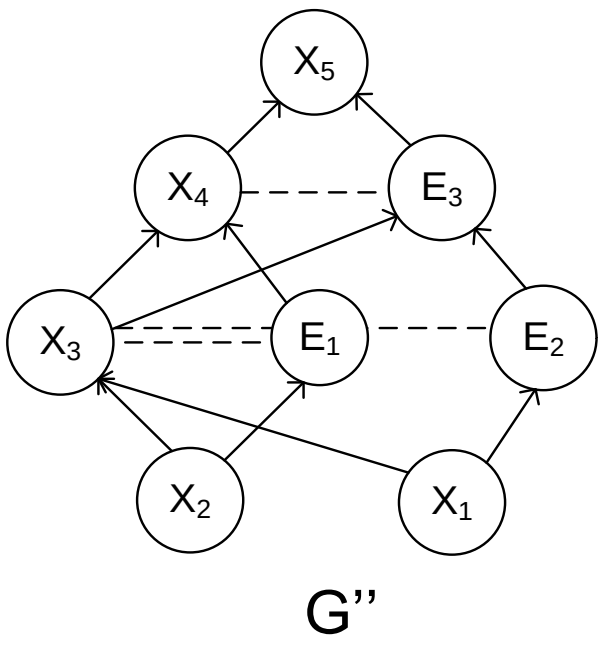

| Moral Edges | Interaction Triplets |
|---|---|
| $(X_4, E_3)$ | $X_3X_4E_3$ |
| $(X_3, E_1)$ | $X_2X_3E_1$ |
| $(X_3, E_2)$ | $X_1X_3E_2$ |

Figure 5.4 Identifying interaction triplets (with irrelevant edges removed)

Next let us consider the variables associated with moral edge $(X_3, E_1)$ and the candidate triplets can be either $X_2X_3E_1$ or $X_1X_3E_1$. Finally, for moral edge $(X_3, E_2)$ the candidate triplets would be $X_2X_3E_2$ or $X_1X_3E_2$.

To balance the number of times $X_2$ and $X_1$ appear in interaction triplets, we choose $X_2X_3E_1$ for moral edge $(X_3, E_1)$ and then should choose $X_1X_3E_2$ for moral edge $(X_3, E_2)$, or vice versa. This ensures the counting numbers for all original variables balance.

### 5.2.2 Constructing the TRC region graph

We aim to build an acyclic two-level region graph. A valid region graph for a $\kappa_5$ full-BFG model using CVM can be built using all primary triplets and all interaction triplets as first level regions, with the shared interactions between these as second level regions and finally a third level region containing variables shared amongst level two regions, as shown in Figure 5.5. Notice that Figure 5.5 has three levels, where the third level contains only marginal variables receiving message passes from parent regions spanning the width of the graph. We have found that inference in a $\kappa_5$ full-BFG model using the build of a three levels region graph can achieve convergence, but that higher dimensional models often fail to converge (for example, in Section 5.4 experiment 2 we report a $\kappa_8$ BFG model with some strong correlation factors that failed to converge by using CVM with three levels).

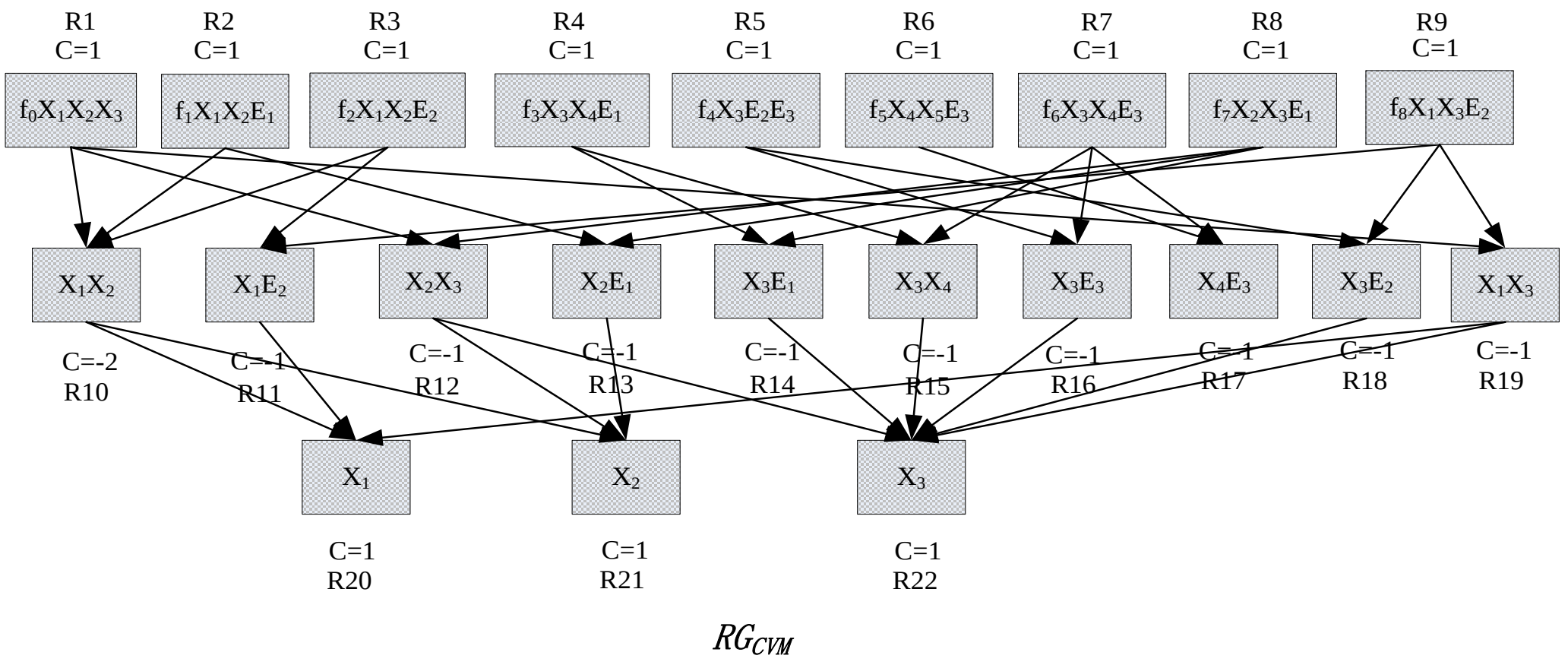

Figure 5.5 Valid region graph generated by CVM

Triplet Region Construction (TRC) starts with a similar approach to CVM. Initially we can use CVM to construct a two level region graph $RG_{Init}$ that is acyclic. The first level regions are the largest regions that contain all factors and variables, and are exclusively determined by primary triplets and interaction triplets. The second level regions are simply the intersections of the first level regions.

For example, we construct $RG_{Init}$ for a $\kappa_5$ full BFG $G'$ in Figure 5.6.

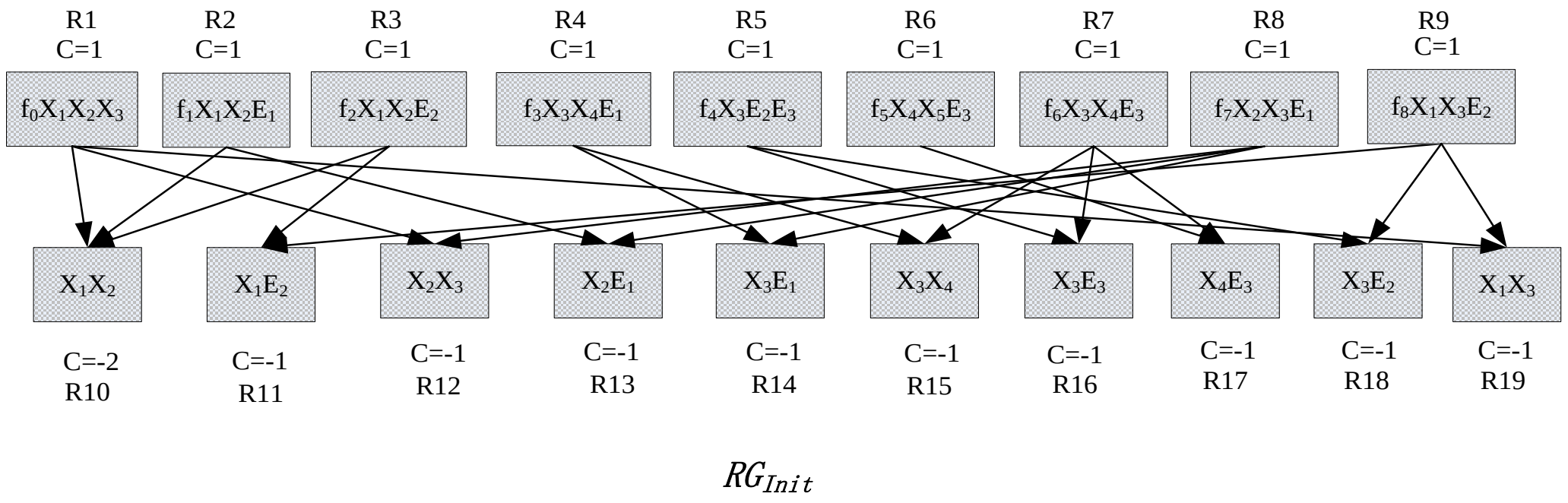

Figure 5.6 Two level initial region graph for $\kappa_5$ BFG model with region counting number listed aside

$RG_{Init}$ is not yet valid since a valid region graph requires each factor and each variable to be counted exactly once. We therefore present an optimization procedure that will guarantee the counting number by manipulating $RG_{Init}$.

From $RG_{Init}$ the second level intersection regions can be classed into two types:

- *Hybrid intersection*: contains an original variable, $X_i$, and an intermediate variable, $E_t$.
- *Cognate intersection*: contains two original variables, $X_i$ and $X_j$, except $X_1X_2$.
- *Root intersection*: is a cognate intersection $X_1X_2$ that is connected to the primary root region ( $X_1X_2X_3$ ) and contains the root variable in the full-BFG, $X_1$.

All hybrid intersections have identical counting number -1, and this is invariant with the number of dimensions. This is important since it helps to satisfy the maxent-normal property and so it would be preferable if all intersections are hybrid intersections. If these hybrid intersections are sufficient to cover all interactions among the first level regions, the region graph construction can be optimized by removing all cognate intersections.

We can test the region graph connectivity of the first level regions, by removing all cognate intersections (if we also removed the root intersection region, $R_{X_1X_2}$, the first level region $R1$ would become disconnected from the rest of the region graph, hence it is not defined as a cognate intersection).

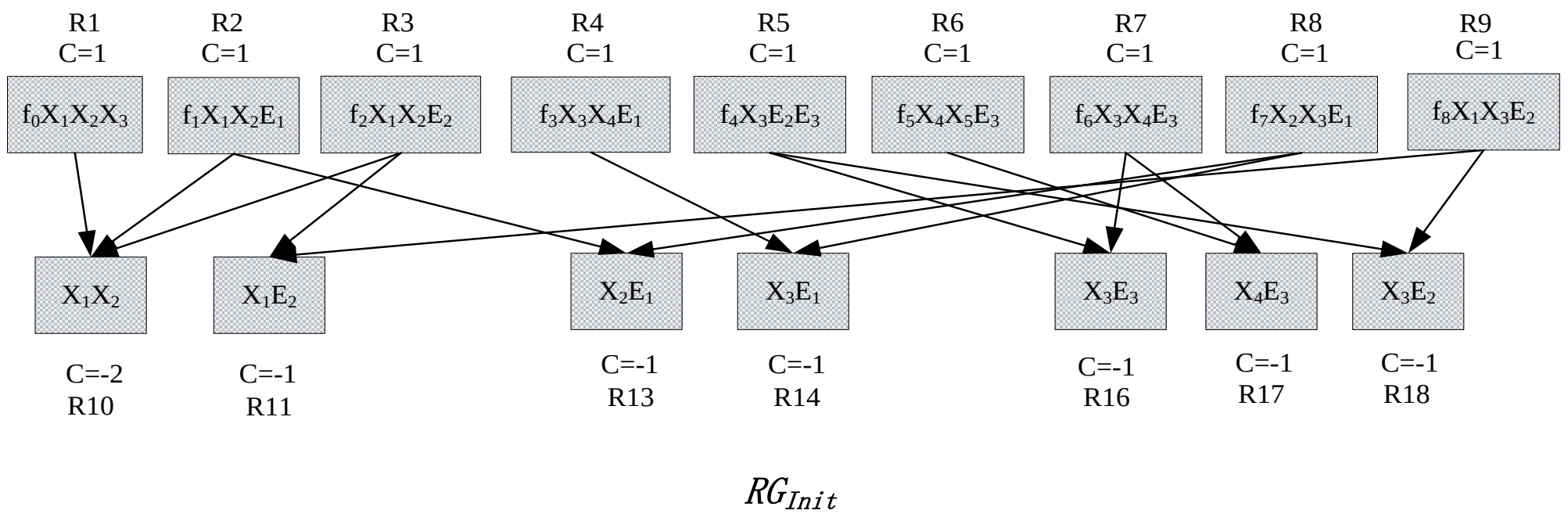

Figure 5.7 Connectivity after removing all cognate intersection regions except root intersection region $R_{X_1X_2}$

Figure 5.7 shows the result of removing all cognate intersections from Figure 5.6, demonstrating that it is sufficient to connect all first level regions using hybrid intersections and the root intersection alone. This suggests that we can optimize the region graph but to do so we need to guarantee the counting numbers. Clearly, the cognate intersections are represented as edges in the full-BFG and, given the index ordering of the original variables, $X_i, i = 1,...,n$, there is a clear pattern to the counting numbers. The absolute value of counting numbers decreases by one along the path of edges in the original full-BFG from lower to higher indexed original variables. Formally, the counting numbers, $c_{R_{X_iX_j}}$, for regions $R_{X_iX_{i+1}}$, $R_{X_{i+1}X_{i+2}}$ and $R_{X_iX_{i+2}}$ satisfy: $c_{R_{X_iX_{i+1}}} = c_{R_{X_iX_{i+2}}} + c_{R_{X_{i+1}X_{i+2}}}$. This is illustrated in Figure 5.8, where the counting number for each cognate intersection (edge) is shown. For example, the counting numbers, $c_{R_{X_iX_j}}$, for regions $R_{X_1X_3}$, $R_{X_2X_3}$ and $R_{X_1X_2}$ satisfy: $c_{R_{X_1X_2}} = c_{R_{X_1X_3}} + c_{R_{X_2X_3}}$.

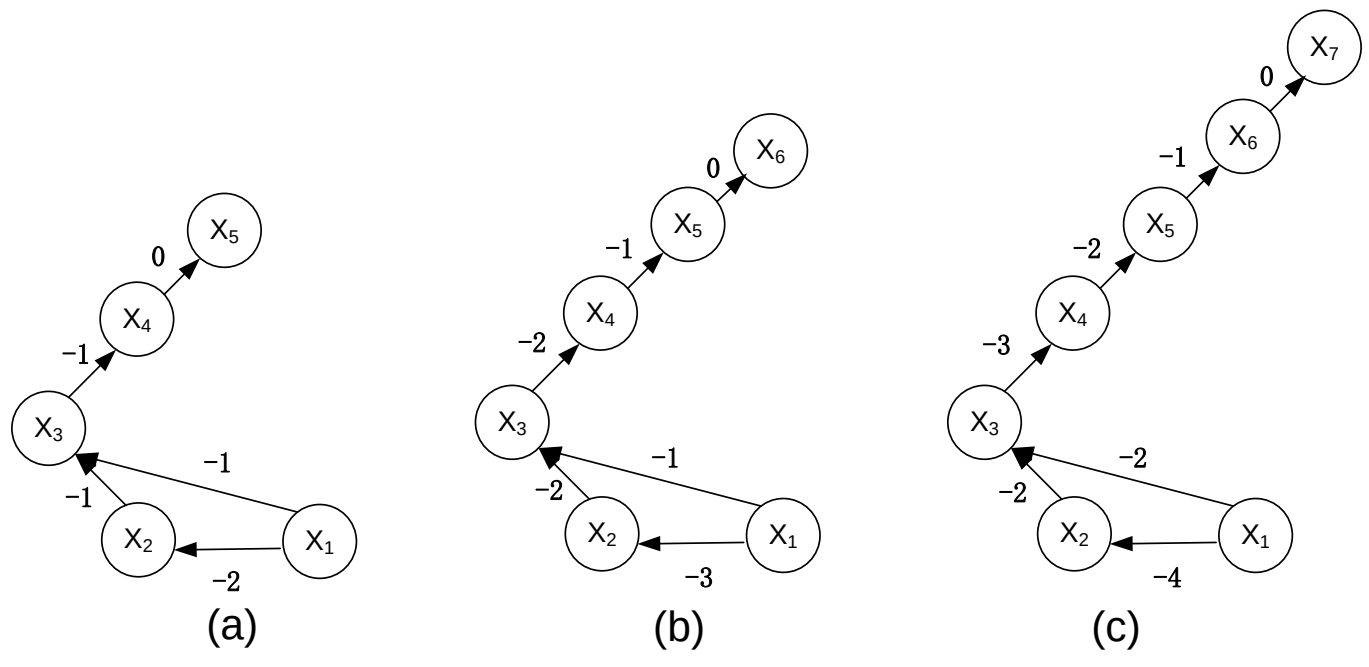

Figure 5.8 $\kappa_5$, $\kappa_6$ and $\kappa_7$ full-BFG with only original variables, and counting numbers (resulting from $RG_{Init}$) for each cognate intersection (edge) listed

We propose the *cognate intersection pruning* algorithm, to prune cognate intersections from $RG_{Init}$, which results in a new region graph $RG_{TRC}$, and guarantees optimum counting numbers and satisfies maxent-normality.

Based on the pattern of counting numbers we prune each cognate intersection $X_i X_j$ ($i < j$) by removing $X_i$ when the counting number does not equal one, $c_{X_i} \neq 1$. This will result in a region graph whose variable nodes have the counting number one, and is maxent-normal (with proof in Appendix C).

***Algorithm 4 Cognate Intersection Pruning***

**Input**: Initial region graph $RG_{Init}$, full-BFG $G'$

**Output**: TRC region graph $RG_{TRC}$

**1.** **do** each original variable $X_i$ in $G'$, $1 \le i \le N$ where $N$ includes all variables

**2.** **calculate** the counting number $c_{X_i} = \sum_{R_i \in R} c_{R_i}$

**3.** **if** $c_{X_i} \neq 1$

**4.** **for** each cognate intersection $I_j$ in $RG_{Init}$, with variables $X_0$ and $X_1$

**5.** **if** $X_0 \in pa\{X_1\}$ and $I_j \neq R_{X_1 X_2}$

**6.** **remove** $X_0$ from $I_j$ in $RG_{Init}$

**7.** **end if**

**8.** **end for**

**9.** **end if**

**10.** **until (**$c_{X_i} = 1$ for all variables**)**

**11.** $RG_{TRC} = RG_{Init}$

**12.** **return** $RG_{TRC}$

Algorithm 4 Cognate intersection pruning

After applying the cognate intersection pruning algorithm we have transformed the initial region graph $RG_{Init}$ into our target $RG_{TRC}$. Figure 5.9 shows the $RG_{TRC}$ generated by the cognate intersection pruning algorithm from the initial region graph $RG_{Init}$: regions $R12$, $R15$ and $R19$ have all been pruned from $RG_{Init}$.

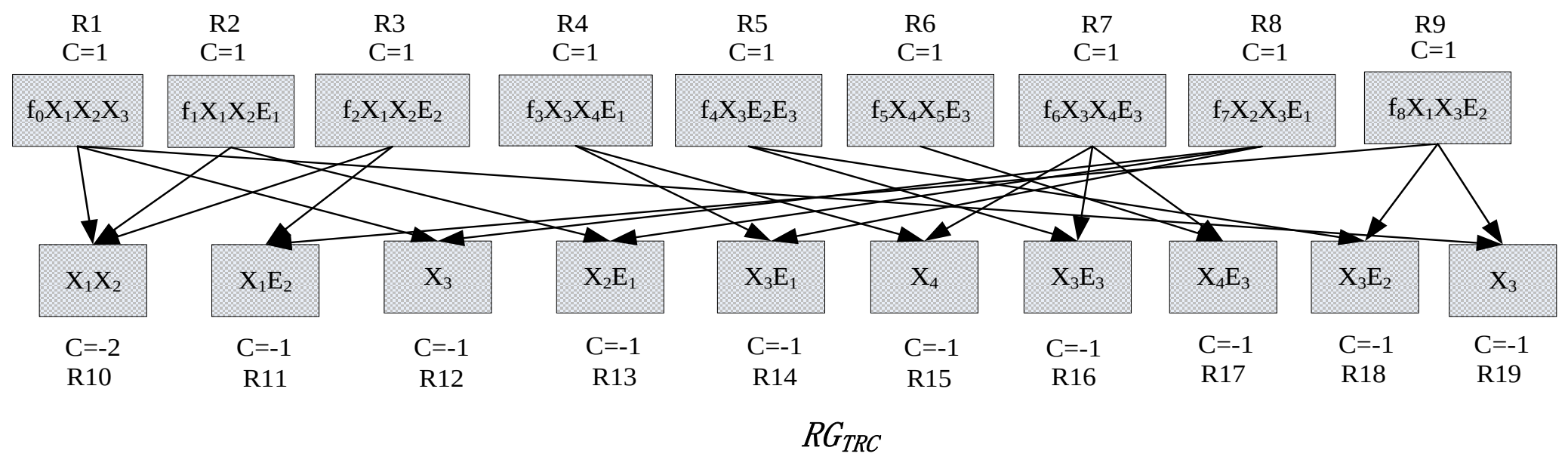

Figure 5.9 $RG_{TRC}$ generated for a $\kappa_5$ BFG

In general, if $n$ is the number of original variables, the number of cognate intersections pruned is $(n-2)$.

We formalize the TRC algorithm, by summarizing all the previous algorithms and principles for constructing an optimal region graph for BFG, as shown in Algorithm 5.

Our TRC algorithm is top down and comparable, in terms of resulting region graph structure, to bottom-up region pursuing algorithms, such as (Welling, 2004). The cognate intersection pruning algorithm is actually generating a local equivalent structure to that which would be produced by adding outer regions[8] to existing region graph shown in (Welling, 2004), resulting in a valid region graph. From this point of view, unlike CVM, TRC does not require that a second level region not be a sub-region of any other second level regions, but nevertheless TRC still satisfies the region graph constraints. In section 5.2.5 we provide a more intuitive justification of TRC given its connections with join graph approaches.

***Algorithm 5 Triplet Region Construction (TRC)***

**Input**: $\kappa_d$ full-BFG $G'$ (with $d > 4$)[9]

**Output**: $RG_{TRC}$

**1. for** each variable $X_i$ in $G'$, $1 \le i \le N$

[8] An outer region is defined to be a region with no parents.

[9] We require the minimum dimension $d > 4$ because a $\kappa_4$ full-BFG does not build a valid region graph under the TRC algorithm. In the case of $d <= 4$ we instead multiply all factors into one single region and perform BP on it.

2. **if** size of $pa\{X_i\} == 2$

3. **add** $X_i$ and $pa\{X_i\}$ to primary triplet set $F_0$

4. **find** $pa\{X_i\} = \{P_0, P_1\}$ where $(P_0, P_1)$ is a moral edge

5. **find** parent common set $\phi\{P_0, P_1\} = pa\{P_0\} \cap pa\{P_1\}$

6. **if** size of $\phi\{P_0, P_1\} > 1$ **{**choice must be between root variable $X_1$ and single parent variable $X_2$ **}**

7. **for** the size of interaction triplet set $U_0$,

8. **calculate** the number of interaction triplets $n_{X_1}$ containing $X_1$ and the number of interaction triplets $n_{X_2}$ containing $X_2$

9. **if** $n_{X_1} > n_{X_2}$ **return** the parent variable $C_0 = X_2$

10. **else return** $C_0 = X_1$

11. **end if**

12. **end for**

13. **else return** the only parent node $\phi\{P_0, P_1\} = C_0$

14. **end if**

15. **if** $C_0 \neq null$ **add** $\{P_0, P_1\}$ and $C_0$ to a new interaction triplet, and update $U_0$

16. **end if**

17. **end if**

18. **end for**

19. **compose** first level of $RG_{Init}$ by triplet sets $F_0$ and $U_0$

20. **compose** second level of $RG_{Init}$ by intersections of the first level regions

21. **call** ***Algorithm 4 cognate intersection pruning***

22. **return** $RG_{TRC}$

Algorithm 5 TRC algorithm

### 5.2.3 Proof that TRC region graph is MaxEnt-Normal and has correct counting numbers

Here we prove that $RG_{TRC}$ for an example $\kappa_5$, full-BFG model, is maxent-normal and has the correct counting number for each variable. The more detailed general proof of this, for any $RG_{TRC}$, $\kappa_n$ full-BFG is given in Appendix C.

We use the same approach as Yedidia, who gives a proof that the Bethe approximation is maxent-normal, and, given this, the entropy of the region graph, $H_{RG}$, can be written as Equation 5.2 (Yedidia et al., 2005).

$$H_{RG} = \sum_{i=1}^{N} H(b_i) - \sum_{a=1}^{M} I(b_a) \tag{5.2}$$

where $N$ is the number of variables, $X_i$, and $M$ is the number of factors, $a$, ($X_a$ are the variables defined by the factor $a$). $H(b_i) \equiv -\sum_{X_i} b_i(X_i) \ln b_i(X_i)$ is the sum of entropies from all variables $X_i$ in the region graph, and $I(b_a) \equiv \sum_{X_a} b_a(X_a) \ln b_a(X_a) - \sum_{i \in N(a)} H(b_i)$ is the entropy for a region containing factor $a$, minus the entropies of all variables contained in factor $a$. $H_{RG}$ is maximal, equalling $\sum_{i=1}^{N} H(b_i)$, when all beliefs, $b_i(X_i)$ and $b_a(X_a)$, are uniform, and under these circumstances the mutual information, $I(b_a)$, equals zero.

For the region graph in Figure 5.9 Equation 5.2 can be expanded, giving Equation 5.3.

$$\begin{aligned} H_{RG} = \sum_{i=1}^{N} H(b_i) - I(\{X_1, X_2\},\{X_3\}) - I(\{X_1, X_2\}, E_1) - I(\{X_1, E_2\}, X_2) - \\ I(\{X_3, E_1\},\{X_4\}) - I(\{X_3, E_2\}, E_3) - I(\{X_4, E_3\}, X_5) - I(\{X_3, E_3\}, X_4) - \\ I(\{X_2, E_1\}, X_3) - I(\{X_3\}, X_1, E_2) \end{aligned} \tag{5.3}$$

where each $I$ term is the mutual information for a level 1 triplet region (all three variables in $I$ term), the level 2 interaction regions it is connected to (two or single variables with bracket $\{...\}$) and the single variables belonging to the level 1 region but which are absent from the connected level 2 regions. For instance $I(\{X_1, X_2\}, E_1) \equiv -H(b_{X_1X_2E_1}) + H(b_{X_1X_2}) + H(b_{E_1})$. Therefore, for Figure 5.9 $H_{RG} = \sum_{i=1}^{N} H(b_i)$ because all of the $I$ terms have value zero, and $H_{RG}$ maximised when the beliefs are uniform it is maxent-normal.

Next we turn to the counting number of $RG_{TRC}$ for a $\kappa_5$ full-BFG model. Here the region based entropy $H_{RG}$ is defined by the sum of the entropies from all regions, weighted by their counting numbers, as shown by Equation 5.4.

$$H_{RG} = H(\sum_{t=1}^{L1} c_t R_t + \sum_{i=1}^{L2} c_i R_i) \tag{5.4}$$

We can write $H_{RG}$ for the region graph in Figure 5.9 as the sum of all region entropies, weighted by their counting numbers, as in Equation 5.5.

$$\begin{aligned}
&H_{RG} = H(\{X_1, X_2, X_3\}) + H(\{X_1, X_2, E_1\}) + H(\{X_1, E_2, X_2\}) + \\
&H(\{X_3, E_1, X_4\}) + H(\{X_3, E_2, E_3\}) + H(\{X_4, E_3, X_5\}) + H(\{X_3, E_3, X_4\}) + \\
&H(\{X_2, E_1, X_3\}) + H(\{X_3, X_1, E_2\}) - 2H(\{X_1, X_2\}) - H(\{X_3\}) - \\
&H(\{X_1, E_2\}) - H(\{X_3, E_1\}) - H(\{X_4\}) - H(\{X_3, E_2\}) - H(\{X_4, E_3\}) - \\
&H(\{X_3, E_3\}) - H(\{X_2, E_1\}) - H(\{X_3\})
\end{aligned} \tag{5.5}$$

Each region one triplet's entropy is cancelled by its connected second level regions entropies and this should result in a counting number of one for each variable. For example, the number of positively weighted occurrences for $X_1$ in Equation 5.5 is four and the number of negatively weighted occurrences for $X_1$ is three, giving a counting number of one.

## 5.2.4 TRC complexity

Here we compare the complexity of TRC with the Junction Tree algorithm. The space complexity for inference in a DCCD model using JT is, of course, exponential, being $O(m^n)$, where $m$ is the maximum number of discrete states for each variable and $n$ the number of variables. Crucially, we now show that, in contrast the space complexity for the TRC algorithm is polynomial.

From our proof (Appendix C) the number of region edges, $E_{RG}$, updated during message passing in $RG_{TRC}$, is the absolute value of counting number plus one, via all intersections. These intersections are composed of all intersections with counting number -1 and those with counting number from $(3-n)$ down to -2, so we can calculate the number of region edges as given by Equation 5.6.

$$\begin{aligned} E_{RG} &= \sum_{j=1}^{(n-2)^2+1} (|c_{R_j}|+1) \\ &= [n-2+(n-2)(n-3)/2]+[n-2]+[2(n-2)^2-2n+4)] \\ &= (n-2)(5n-11)/2, \quad R_j \in \{\text{intersections}\}, n>3 \end{aligned} \tag{5.6}$$

The number of messages being updated is polynomial in Equation 5.6, so the complexity of TRC is polynomial.

## 5.2.5 Relationship with the Join Graph approach

The proof in section 5.2.3 is based on information theory which may be less intuitive, when justifying the cognate intersection pruning algorithm, in terms of verifying the TRC region graph's effectiveness. This section converts the TRC region graph into a join graph to intuitively support the proof.

**Example 5.3**

We borrow the example BN, $G$, used in (Mateescu et al., 2010) with variables renamed for consistency purpose in this thesis and shown in Figure 5.10.

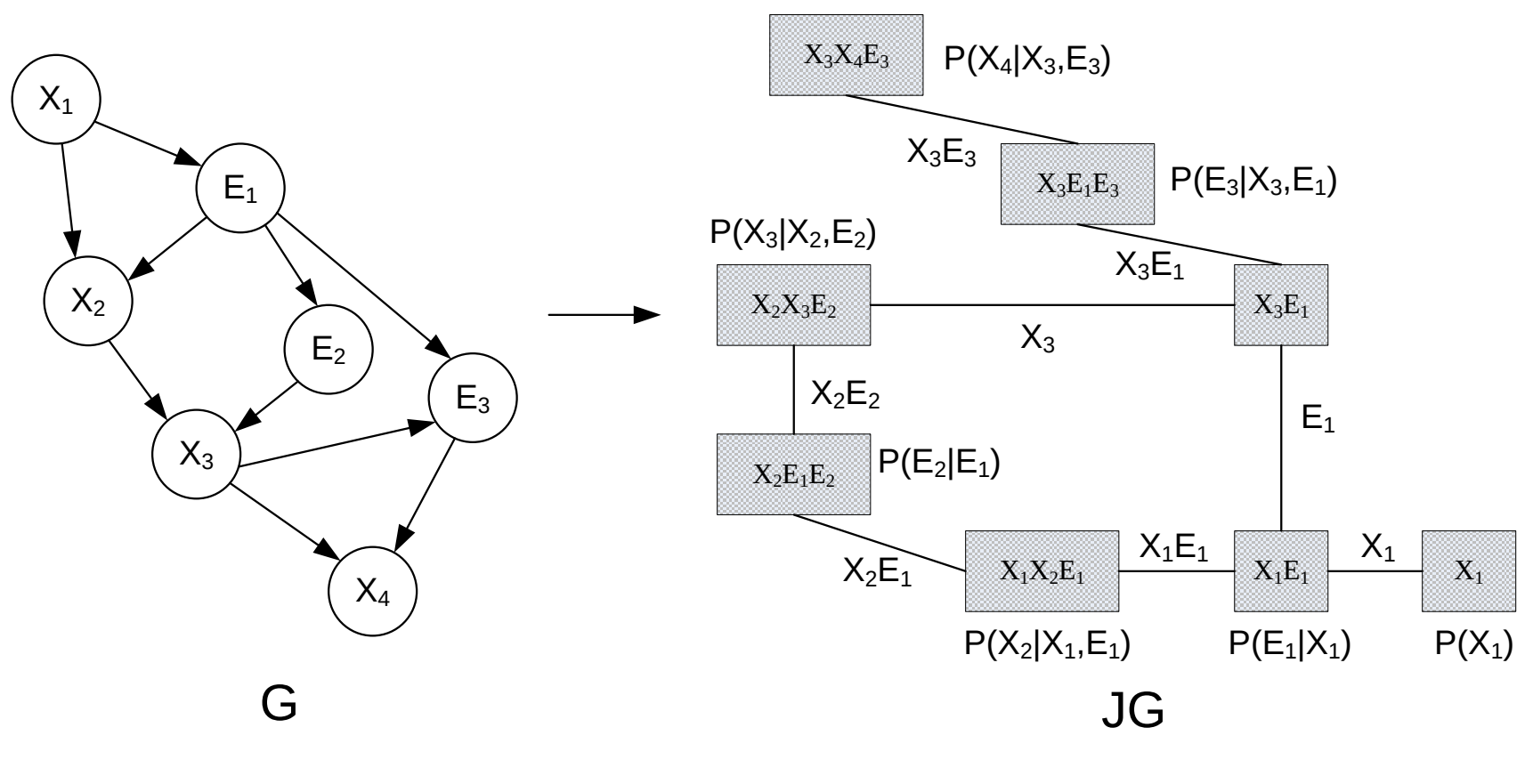

Figure 5.10 Generated Join Graph (with CPD listed aside each cluster) for the original BN, $G$, using the IJGP join graph structuring algorithm

The join graph, $JG$, in Figure 5.10 uses the IJGP algorithm (Mateescu et al., 2010), with bounded mini-cluster size of three. Intersections are labelled variables listed aside the edges between pairs of clusters in Figure 5.10 $JG$. The IJGP algorithm is guaranteed to converge in finite time shown in (Mateescu et al., 2010).

Figure 5.10 $G$ is not a full BFG but its factor size is limited to three, so it can still be considered a BFG. We can apply the same analogue used in TRC algorithm to partition $G$ into primary triplets and interaction triplets and generate a TRC region graph.

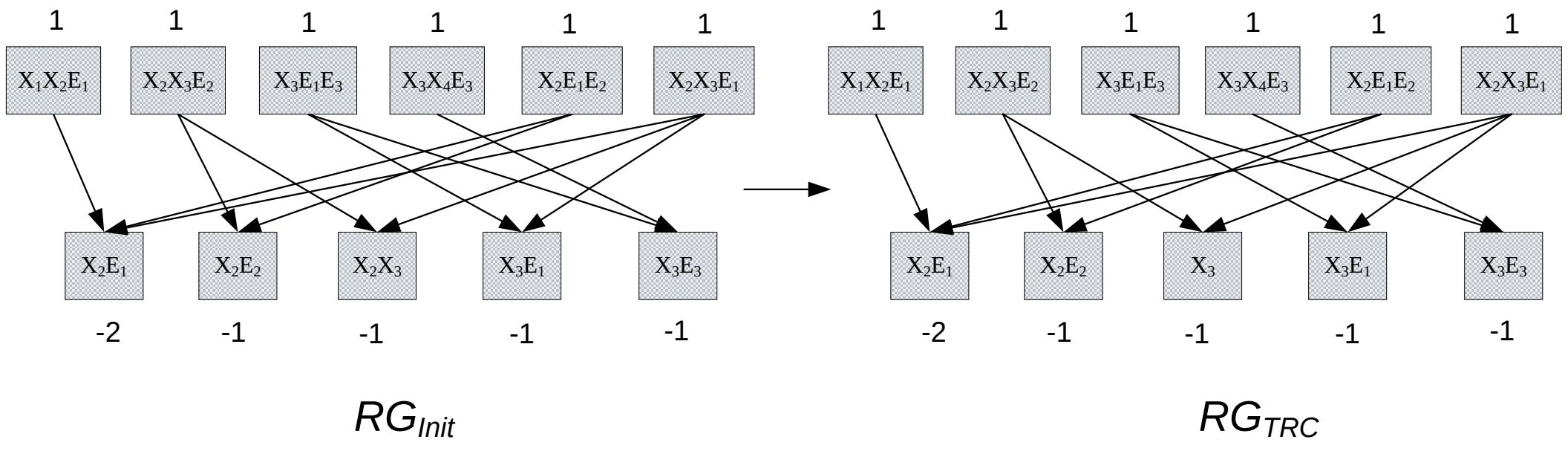

Figure 5.11 Generated TRC region graph for the original BN $G$ in Figure 5.10

Except for the four primary triplets[10] that are obviously included in the first level of the region graph, there are two interaction triplets, $X_2E_1E_2$ and $X_2X_3E_1$ are added to incorporate two moral edges $(X_2, E_2)$ and $(X_3, E_1)$. Figure 5.11 shows the initial region graph generated by these triplets. There is only one cognate intersection $X_2X_3$ that needs to be pruned in $RG_{Init}$. After applying cognate intersection pruning algorithm we obtain $RG_{TRC}$ in Figure 5.11. It can be verified that each variable is counted exactly once in $RG_{TRC}$.

[10] CPDs $P(X_1)$ and $P(E_1 | X_1)$ are multiplied to $P(X_2 | X_1, E_1)$.

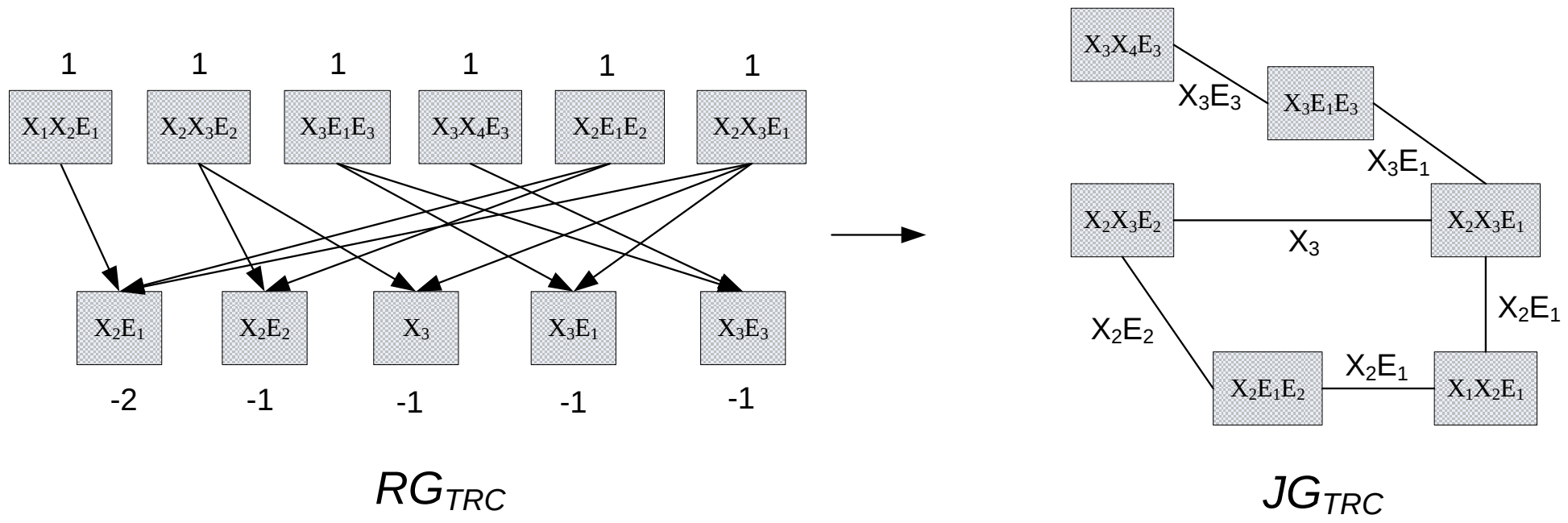

Figure 5.12 Convert TRC region graph into join graph

After converting the TRC region graph to a join graph, we obtain $JG_{TRC}$ in Figure 5.12. and can verify that $JG_{TRC}$ is equivalent with $JG$ in Figure 5.10, with the small clusters ($X_1$ and $X_1E_1$) being absorbed into the triplet $X_1X_2E_1$. The cognate intersection pruning algorithm pruned the intersection $X_2X_3$ into $X_3$, so variable $X_2$ will not be propagated cyclically in $JG_{TRC}$, given that $JG_{TRC}$ is a cyclic graph. (Mateescu et al., 2010) has proven that this pruning will not change the CI assumptions captured by the join graph, which is also used in IJGP algorithm for avoid over-counting problems on cyclic join graph.

The cognate intersection algorithm not only avoids over-counting in the TRC region graph but also generates an optimal region graph equivalent to a join graph produced using IJGP.

Figure 5.13 illustrates the join graph generated for the TRC region graph in Figure 5.9. Two dashed circle lines mark out the loops in this join graph, where the original un-pruned intersections are listed aside each pruned intersection. We can verify that the cognate intersection pruning algorithm avoids the cyclic propagation of the pruned variables in the join graph $JG_{TRC}$.

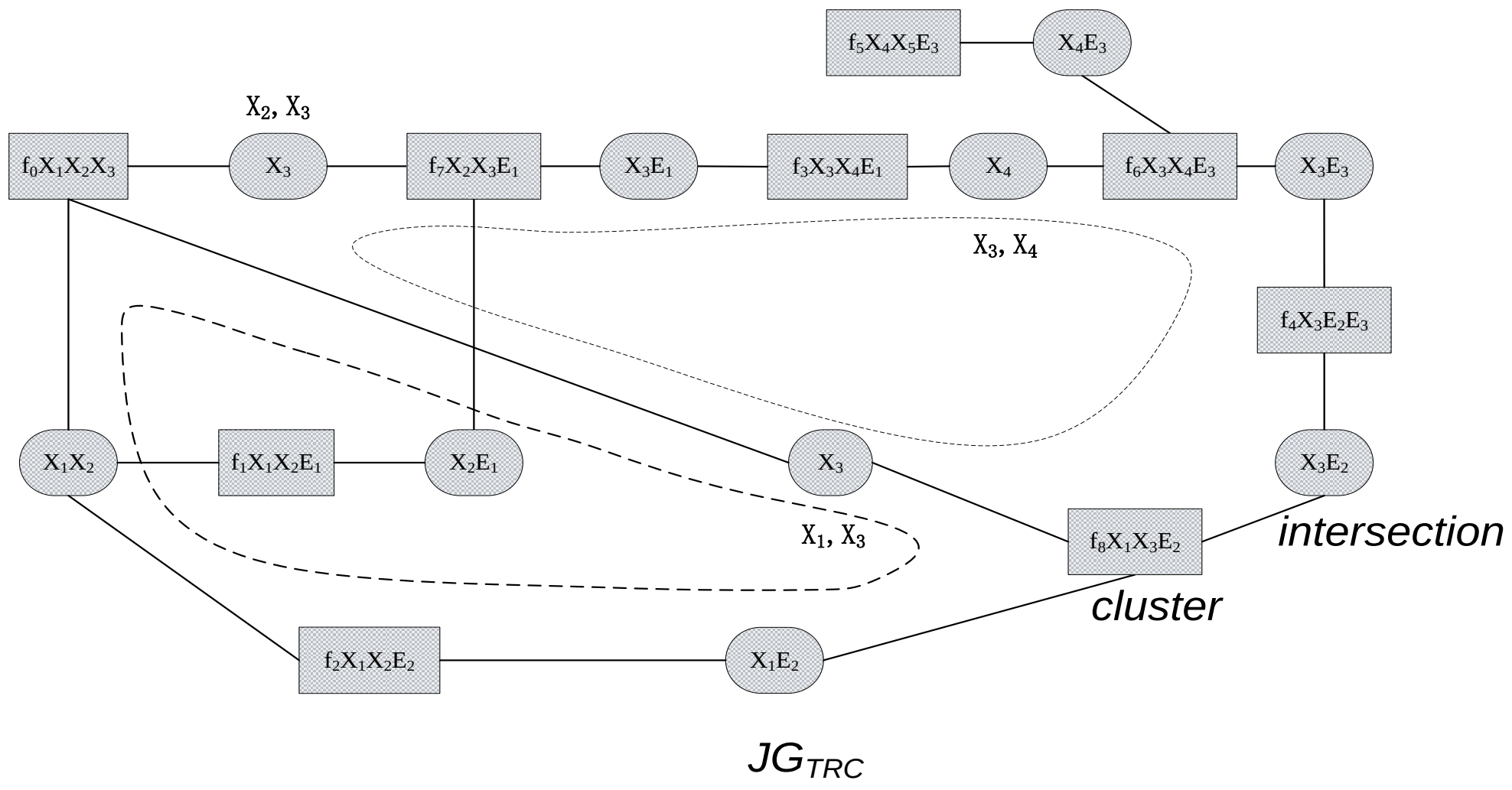

Figure 5.13 Join graph for TRC region graph in Figure 5.10

The TRC algorithm has similar characteristics with join graph based approach, i.e. it guarantees convergence in finite time. Differed from IJGP, TRC algorithm operates on an acyclic graph (and its join graph has no cyclic propagation of variables), which does not need to worry about cyclic graph based problems. If we convert the CVM three-level region graph in Figure 5.5 into a join graph, we will obtain a join graph with cyclic propagation of the third level region variables. If it converges it tends to be very accurate, i.e. for $\kappa_5$ BFG it may converge for some instances and the result is accurate, but in higher dimensional cases the CVM three-level region graph fails to converge, since by converting it to join graph it contains many cyclic propagations of the third level region variables.

# 5.3. DDBP algorithm

The TRC algorithm is sufficient for inference on full-BFG models containing discrete variables. However, to perform inference on high dimensional continuous DCCD models, we must first discretize the continuous variables. We use DD to carry out dynamic discretization during belief propagation and name the combined method ***Dynamically Discretized Belief Propagation*** (DDBP). DDBP can be viewed as a replacing JT in the DDJT algorithm with TRC.

We formalize the DDBP algorithm as show in Algorithm 6.

***Algorithm 6 Dynamically Discretized Belief Propagation (DDBP)***

**Input**: original BN $G$ (BF decomposable and dimension $d > 4$)

**Output**: original BN $G$ with marginals

1. **reorder** all nodes in $G$ from ancestors to descendants with new label $X_i$
2. **convert** $G$ into a DCCD $G$ by adding edges from ancestors to descendants
3. **reassign** all CPDs for $f(X_i \mid X_1, ..., X_{i-1})$ by blocking unrelated ancestors
4. **transform** $G$ into binary factorized version complete-BFG $G'$
5. **call Algorithm 2** *Triplet Region Construction* to generate $RG_{TRC}$
6. **assign** each region's factor $\phi(R_k)$ by multiplying the NPTs $p(X_i \mid pa\{X_i\})$ for all variables into the relevant primary triplets
7. **assign** uniform factors for all interaction triplets and separators
8. **initialize** each region's discretization $\mathbf{\Psi}_k^{(0)}$, by its support $\mathbf{\Omega}_k$
9. **do compute** the approximate NPTs, $p^{(0)}(X_i \mid pa\{X_i\})$, on $\Psi_{X_i}^{(l-1)}$ for each variable $X_i$ and initialize each region's factor
10. **query** the BN from node $\mathbf{X}_E = \mathbf{e}$
11. **while** total relative entropy error > target threshold
12. **create** a new discretization $\mathbf{\Psi}_k^{(l)}$ for region domain $\mathbf{\Omega}_k$
13. **define** updating order by depth first search algorithm
14. **perform** *two way messaging*
15. **check** BP convergence by *low discrepancy threshold*
16. **compute** the new discretized potential $\phi^{(l-1)}(R_k)$
17. **compute** the approximate total relative entropy error
18. **end for**
19. **until** convergence of the posterior marginal for each region by stable entropy error stopping rule and low entropy error stopping rule
20. **extract** marginal for each node from the relevant region and copy to $G$
21. **return** $G$

Algorithm 6 DDBP algorithm

Both DD and TRC based belief propagation have their stopping rules. The TRC stopping rule is a *low discrepancy rule,* which states the discrepancy of region beliefs generated from the previous and current rounds of propagation must be below a certain threshold. For example, if the old beliefs are generated by the first $M$ updates, then the current beliefs are generated by $2M$ updates, where $M$ is the number of connections between the first and second levels in the region graph. The convergence check determines the discrepancy between the old beliefs and current beliefs (the convergence metric is determined by monitoring each individual probability between old and current beliefs). If the discrepancy fails to fall below a certain threshold, e.g. 1.0E-6, the message updates continue. In our tests TRC guarantees convergence in finite time. For DD the stopping rule is similar in that the entropy error also has to fall below a specified threshold. DDBP features an inner loop of BP iterations and DD outer loop iterations. DDBP converges under the convergence of both BP and DD.

## 5.4. Experiments

We report seven experiments to evaluate the TRC algorithm (Experiments 1-2) and DDBP algorithm (Experiments 3-7). The parameterisation for the BN models used is listed in Appendix B. We compare our results to analytical results and MCMC approximations only for the purpose of illustration and validation of our algorithm. Other related algorithms, such as EP and PBP are not compared, because off-the-shelf implementations of these algorithms (for the types of empirical problems investigated here) are not available, and would involve considerable analytical and programming effort.

The DDBP algorithm was written in Java and used libraries available on the AgenaRisk product (AgenaRisk, 2014) using Java JDK 6. This also allowed comparison with DDJT since this is already available in AgenaRisk.

### 5.4.1 Experiment 1: Inference for a $\kappa_5$ BFG model with all binary variables using TRC

We first test TRC alone without involving continuous variables, using a $\kappa_5$ BFG in Figure 5.14 with all binary variables.

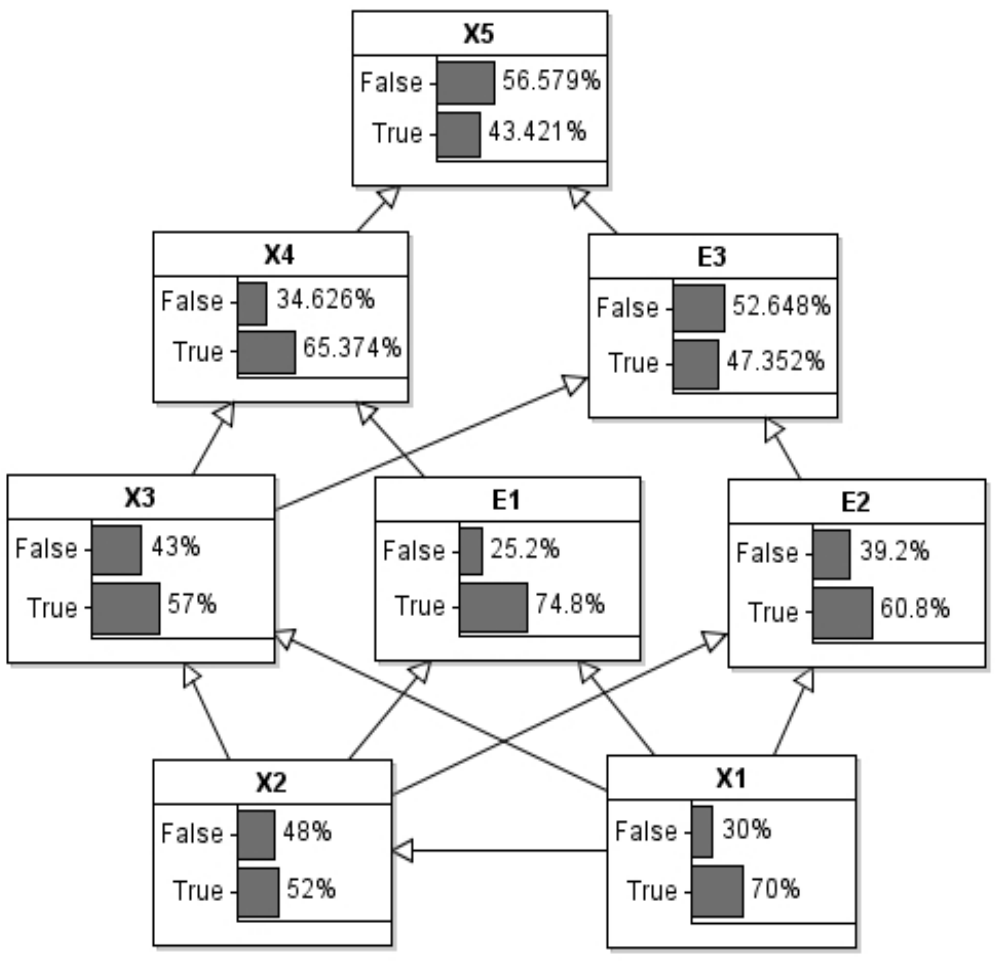

Figure 5.14 TRC result for a $\kappa_5$ BFG with all binary variables

The parameter settings for the NPTs in this experiment are listed in Appendix D. Table 5.1 shows TRC produced good approximations compared to exact results produced under JT. We have tested higher dimensional full-BFG models with more discrete states rather than just using binary state variables. The results are accurate. We have also compared TRC with Bethe method, CVM (three levels), and basic mean field approach (with default message orders in the package) using the fastInf package (Jaimovich, et al, 2014) and the TRC result is better than those produced by these competing approaches.

| Variables | False (TRC) | False (JT) | True (TRC) | True (JT) | KL Distance |
|---|---|---|---|---|---|
| $X_1$ | 0.30000 | 0.30000 | 0.70000 | 0.70000 | 0 |
| $X_2$ | 0.47999 | 0.48000 | 0.52001 | 0.52000 | 4.3119e-10 |
| $X_3$ | 0.43000 | 0.43000 | 0.57000 | 0.57000 | 8.2997e-12 |
| $X_4$ | 0.52647 | 0.51879 | 0.47353 | 0.48121 | 0.00017056 |
| $X_5$ | 0.56579 | 0.59202 | 0.43421 | 0.40798 | 0.0020309 |
| $E_1$ | 0.39200 | 0.39200 | 0.60800 | 0.60800 | 3.1343e-11 |
| $E_2$ | 0.25201 | 0.25200 | 0.74799 | 0.74800 | 1.5233e-10 |
| $E_3$ | 0.34625 | 0.34616 | 0.65375 | 0.65384 | 2.8359e-8 |

Table 5.1 TRC using binary variables

## 5.4.2 Experiment 2: Inference for a $\kappa_8$ BFG with evidence using TRC

The model contains binary (Boolean) variables only, with evidence set on variable $X_8 = False$. NPT settings are listed in Appendix B. In Figure 5.15, variables $X_3$, $E_{27}$ and $E_{71}$ have identical NPT settings, so these variables will have very close marginal distributions.

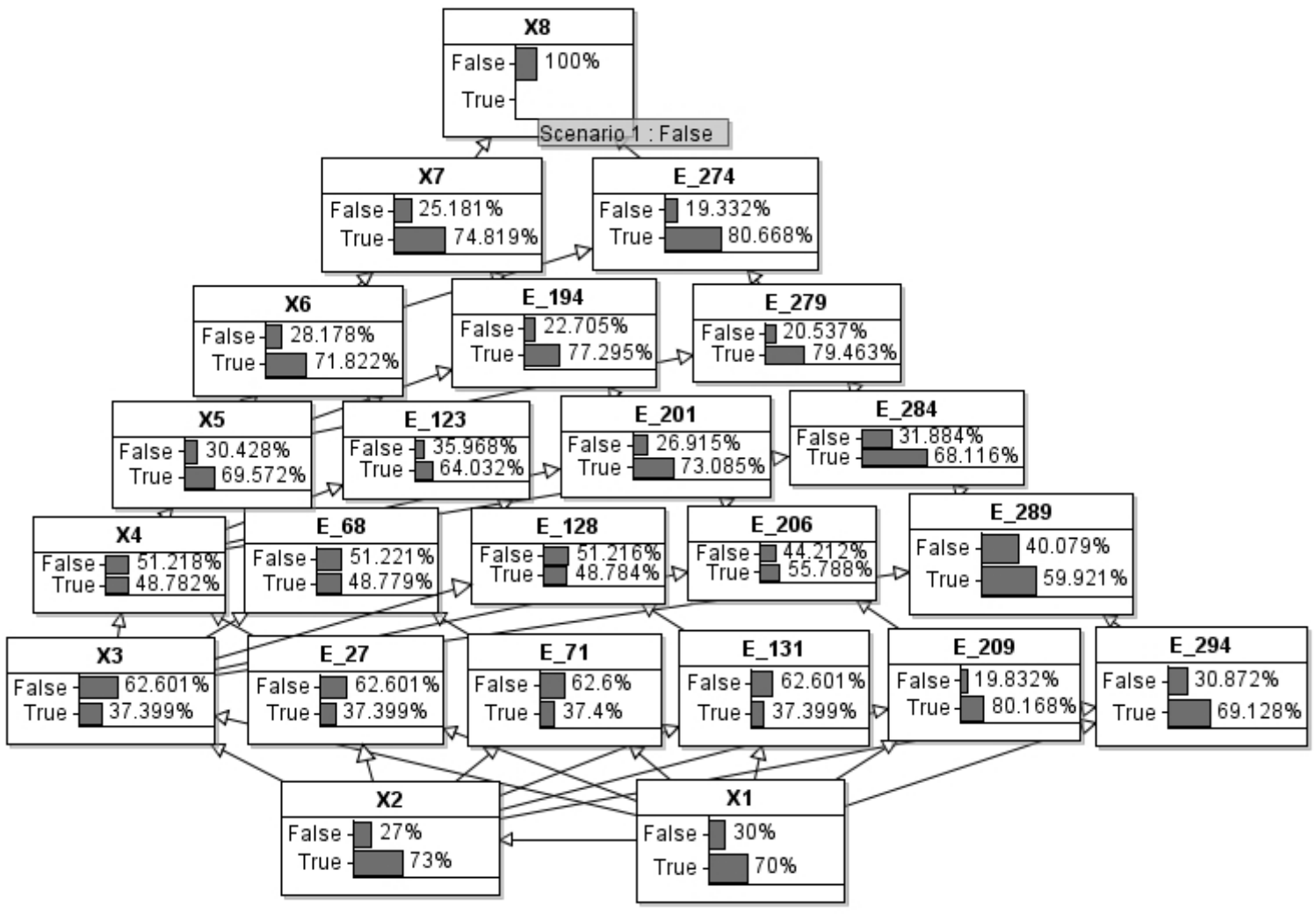

Figure 5.15 a $\kappa_8$ *BFG* with all binary variables

| Variables | False (TRC) | False (JT) | True (TRC) | True (JT) | KL Distance |
|---|---|---|---|---|---|
| $X_1$ | 0.30000 | 0.30000 | 0.70000 | 0.70000 | 0 |
| $X_2$ | 0.27000 | 0.27000 | 0.73000 | 0.73000 | 0 |
| $X_3$ | 0.62601 | 0.62601 | 0.37399 | 0.37399 | 0 |
| $X_4$ | 0.51218 | 0.51218 | 0.48782 | 0.48782 | 0 |
| $X_5$ | 0.30433 | 0.30428 | 0.69567 | 0.69572 | 5.90439E-09 |
| $X_6$ | 0.28166 | 0.28178 | 0.71834 | 0.71822 | 3.55828E-08 |
| $X_7$ | 0.25179 | 0.25181 | 0.74821 | 0.74819 | 1.0616E-09 |
| $E_{27}$ | 0.62600 | 0.62601 | 0.37400 | 0.37399 | 2.13563E-10 |
| $E_{71}$ | 0.62600 | 0.62600 | 0.37400 | 0.37400 | 0 |

Table 5.2 TRC using binary variables

We compare TRC with the JT results in Table 5.2. The maximum KL distance in Table 5.2 is 3.55828e-08 which confirms the performance of TRC. Also, the strongly correlated variables $X_2$, $E_{27}$ and $E_{71}$ are approximated very well.

### 5.4.3 Experiment 3: Inference for 20 dimensional CG-DCCD model using DDBP

We test DDBP using the 20-dimensional conditional Gaussian model shown in Figure 5.16. Parameter settings for the original BN model is shown in the Appendix D (20 dimensions). Results for mean and standard deviation statistics are listed in Table 5.3:

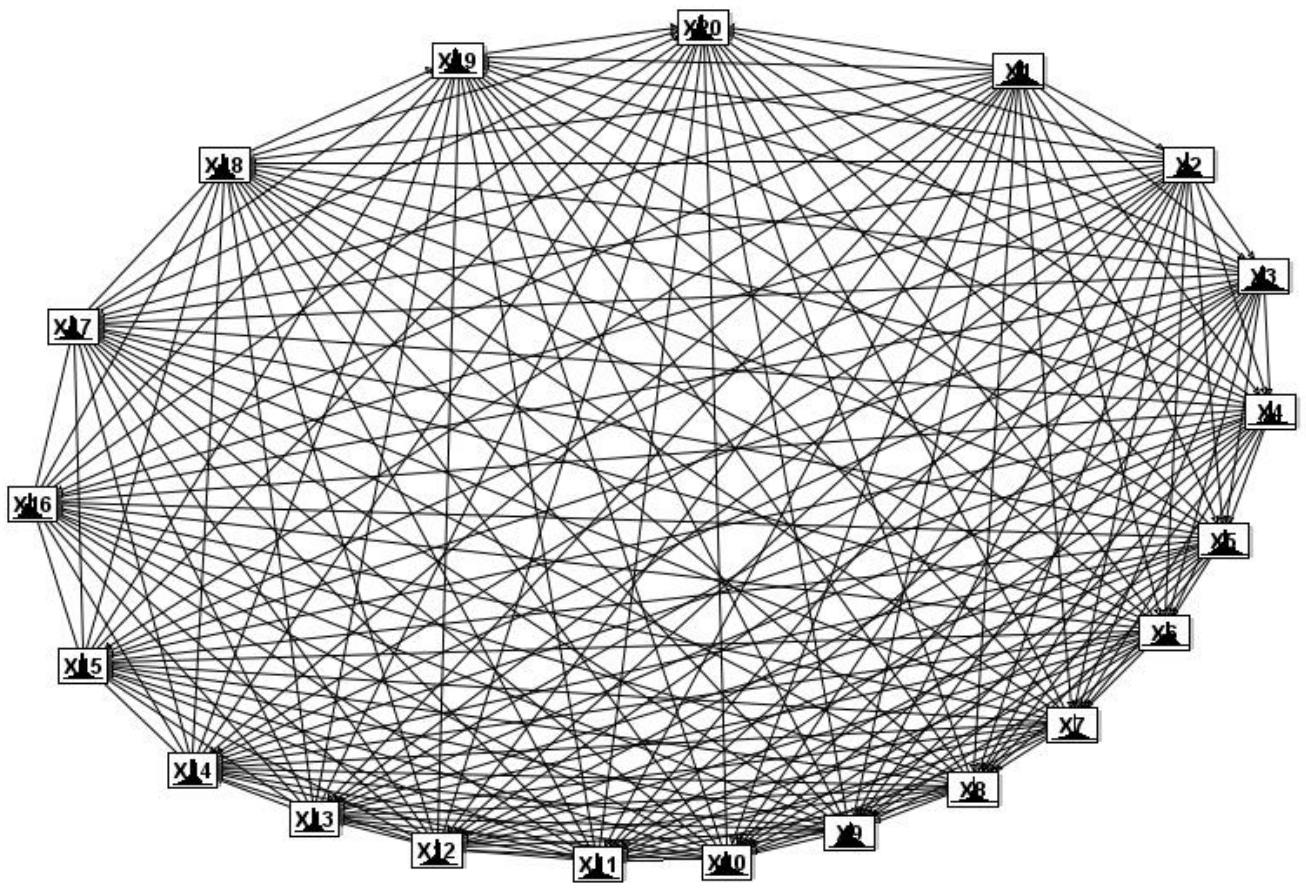

Figure 5.16  20 dimensional CG-DCCD model

| Node | Exact Mean | DDBP Mean | Mean diff | Exact SD | DDBP SD | SD diff. |
|---|---|---|---|---|---|---|
| $X_1$ | 2 | 2.0000 | 0.000% | 2 | 2.0138 | 0.690% |
| $X_2$ | 3 | 2.9970 | 0.100% | 3 | 3.0170 | 0.567% |
| $X_3$ | 4 | 3.9990 | 0.025% | 4 | 4.0130 | 0.325% |
| $X_4$ | 5 | 5.0000 | 0.000% | 5 | 5.0218 | 0.436% |
| $X_5$ | 6 | 5.9990 | 0.017% | 6 | 5.9976 | 0.040% |
| $X_6$ | 7 | 7.0000 | 0.000% | 7 | 6.9960 | 0.057% |
| $X_7$ | 8 | 7.9980 | 0.025% | 8 | 7.9913 | 0.109% |
| $X_8$ | 9 | 8.9980 | 0.022% | 9 | 8.9715 | 0.317% |
| $X_9$ | 10 | 9.9920 | 0.080% | 10 | 9.9560 | 0.440% |
| $X_{10}$ | 11 | 10.997 | 0.027% | 11 | 10.942 | 0.527% |
| $X_{11}$ | 12 | 11.997 | 0.025% | 12 | 11.883 | 0.975% |
| $X_{12}$ | 13 | 12.997 | 0.023% | 13 | 12.907 | 0.715% |
| $X_{13}$ | 14 | 14.000 | 0.000% | 14 | 13.945 | 0.393% |
| $X_{14}$ | 15 | 15.010 | 0.067% | 15 | 14.898 | 0.680% |
| $X_{15}$ | 16 | 16.010 | 0.063% | 16 | 15.899 | 0.631% |
| $X_{16}$ | 17 | 16.999 | 0.006% | 17 | 16.870 | 0.765% |
| $X_{17}$ | 18 | 18.027 | 0.150% | 18 | 17.826 | 0.967% |
| $X_{18}$ | 19 | 19.018 | 0.095% | 19 | 18.842 | 0.832% |
| $X_{19}$ | 20 | 20.056 | 0.280% | 20 | 19.870 | 0.650% |
| $X_{20}$ | 21 | 21.094 | 0.448% | 21 | 20.967 | 0.157% |

Table 5.3 statistics for 20 dimensional CG-DCCD model (DDBP done with DD iterations 35, maximum[11] GBP iterations 60)

Compared with the exact result the accuracy for the mean statistics is high for each variable and a small discrepancy (less than 1%) is produced on standard deviation (SD) statistics. The results suggest a stable approximation without degradation, despite increased dimensionality. The region graph generated by CVM with three levels fails to converge for this model.

### 5.4.4 Experiment 4: Inference for 10 dimensional CG-DCCD model with observations using DDBP

This experiment tests the accuracy of DDBP when the model contains an observation. Here we have a 10 dimensional CG-DCCD model with observation for variable $X_{10} = -10$. The parameter settings for the original ten dimensional BN model are shown in Appendix B. We compare the result with that achieved using MCMC with a single chain and 1.0E6 updates, as well as the exact solution. Table 5.4 shows that the results for both DDBP and MCMC methods are very similar.

| Node | MCMC Mean | DDBP Mean | MCMC SD | DDBP SD | Exact SD |
|---|---|---|---|---|---|
| $X_1$ | 1.616 | 1.611 | 1.990 | 2.000 | 2 |
| $X_2$ | 2.424 | 2.415 | 2.988 | 3.000 | 3 |
| $X_3$ | 3.231 | 3.213 | 3.981 | 4.040 | 4 |
| $X_4$ | 4.052 | 4.026 | 4.970 | 5.030 | 5 |
| $X_5$ | 4.854 | 4.838 | 5.973 | 6.020 | 6 |
| $X_6$ | 5.663 | 5.642 | 6.966 | 7.000 | 7 |
| $X_7$ | 6.461 | 6.455 | 7.958 | 7.990 | 8 |
| $X_8$ | 7.280 | 7.270 | 8.968 | 8.950 | 9 |
| $X_9$ | 8.088 | 8.068 | 9.940 | 9.950 | 10 |

Table 5.4 10 dimensional CG-DCCD model with observation $X_{10} = -10$ (DDBP performed with DD iterations 30, maximum GBP iterations 180)

[11] This is the maximum iteration settings for BP, though it may converge prior to the limit.

## 5.4.5 Experiment 5: Pair correlation test for 15 dimensional CG-DCCD model

Here we use empirical methods to determine how well the DDBP approximates the pair wise covariance/correlation structure in an MGD model and focus on how well the Pearson correlation coefficient, $\rho$, is approximated.

The parameter settings for the original BN model are shown in Appendix D. The DDBP model was run with 30 DD iterations and a maximum 240 GBP iterations.

We use combination pairs of variables $X_1, X_6, X_{11}$ and $X_{15}$ for this experiment (there are $C_{15}^2 = 105$ combinations in total). Figure 5.17 shows the $\kappa_{15}$ BFG model corresponding to the 15 dimensional CG-DCCD model.

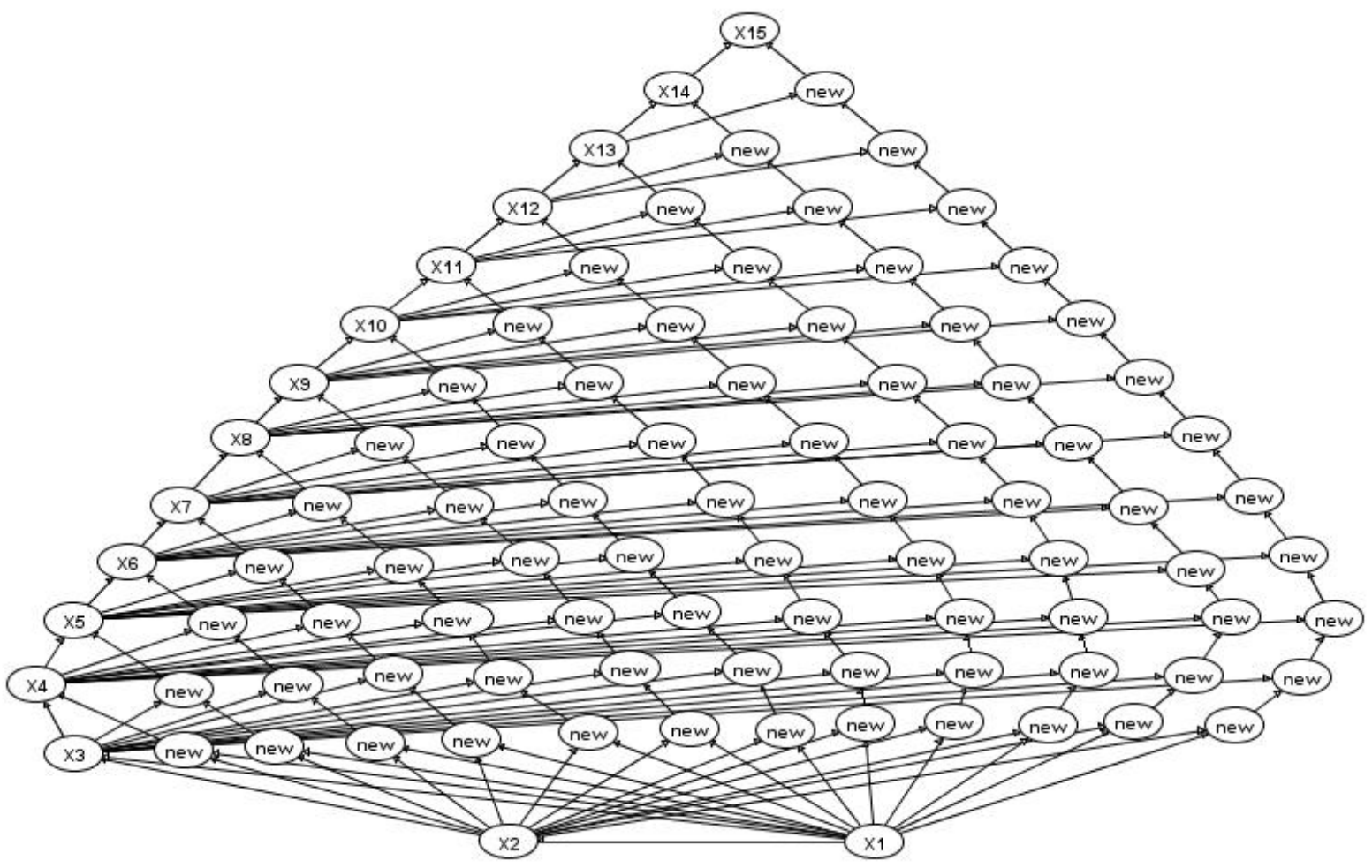

Figure 5.17 $\kappa_{15}$ BFG model (Intermediate variables are labelled as "new" which are added during the full-BF procedure. Nodes $X_1$ to $X_{15}$ are the original variables.)

The results are shown in Table 5.5 alongside the exact correlation.

| Node Pair | Exact $\rho$ | Approximated $\rho$ |
|---|---|---|
| $X_1X_6$ | 0.1 | 0.1010 |
| $X_1X_{11}$ | 0.1 | 0.1048 |
| $X_1X_{15}$ | 0.1 | 0.1083 |
| $X_6X_{11}$ | 0.1 | 0.1036 |
| $X_6X_{15}$ | 0.1 | 0.1042 |
| $X_{11}X_{15}$ | 0.1 | 0.1026 |

Table 5.5 pair correlation test

We use this weak correlation model for testing the accuracy of DDBP. Note that for efficiency considerations this correlation test is only based on a discretized static model produced by DD, to approximate the joint distribution. Then to produce the conditional distribution for each pair in Table 5.5, we fixed the state of one variable and enter evidence on each of its discrete states to query the posterior distribution of the other variables during propagation. The accuracy of the results is lower but computation time is faster. Despite this the result is still a very close approximation to the exact correlation coefficient, $\rho$.

### 5.4.6 Experiment 6: Inference in a linear model using DDBP

This experiment applies DDBP inference to a linear model, which is a sparse model and not a DCCD. Therefore the model is first converted, from a sparse model, to a DCCD by adding edges and assigning zero weights. This experiment shows how to use DDBP to perform inference on general (initially non-DCCD) models.

We use a Bayesian linear regression model (notation is borrowed from (Bishop, 2006)) in Figure 5.18 (a), with the model specification:

$\mathbf{t} = y(\mathbf{x}, \mathbf{w}) + \varepsilon$, where $y(\mathbf{x}, \mathbf{w}) = w_0 + w_1 x_1 + ... + w_D x_D = \mathbf{w}^T \phi(\mathbf{x})$,

The error term $\varepsilon$ is a zero mean Gaussian random variable with precision, $\beta$. For convenience we define $\phi_0(\mathbf{x}) = 1$. The model in Figure 5.18 (b) is defined as:

$w_0 \sim Normal(\mu = 0, \sigma^2 = 1000)$, regression coefficient

$w_1 \sim Normal(\mu = 0, \sigma^2 = 1000)$, bias parameter

$\beta \sim Gamma(1, 10)$, prior for the precision parameter over $\mathbf{t}$

$\text{var} \sim 1/\beta$, noise parameter, inverse of beta

$\mathbf{t} \sim Normal(\boldsymbol{\mu}, \beta^{-1})$, $t_1, ..., t_N$, target variables

$\mathbf{x} \sim Normal(\mu = 0, \sigma^2 = 1000)$, $x_1, ..., x_N$, explanatory variables

$\phi_n(\mathbf{x}_n)$, basis functions

$\boldsymbol{\mu} \sim \mathbf{w}^T \phi(\mathbf{x}_n)$, mean parameter over $\mathbf{t}$

The likelihood: $p(\mathbf{t} \mid \mathbf{x}, \mathbf{w}, \beta) = \prod_{n=1}^{N} N(t_n \mid \mathbf{w}^T \phi(\mathbf{x}_n), \beta^{-1})$, we assume noise is known so we set $\varepsilon \sim Normal(\mu = 0, \sigma^2 = 0.25)$.

For a simple illustration Figure 5.18 (b) shows the model structure for the plate model used in DDBP for the case $n = 3$ in (the algorithm is independent of the size of $n$).

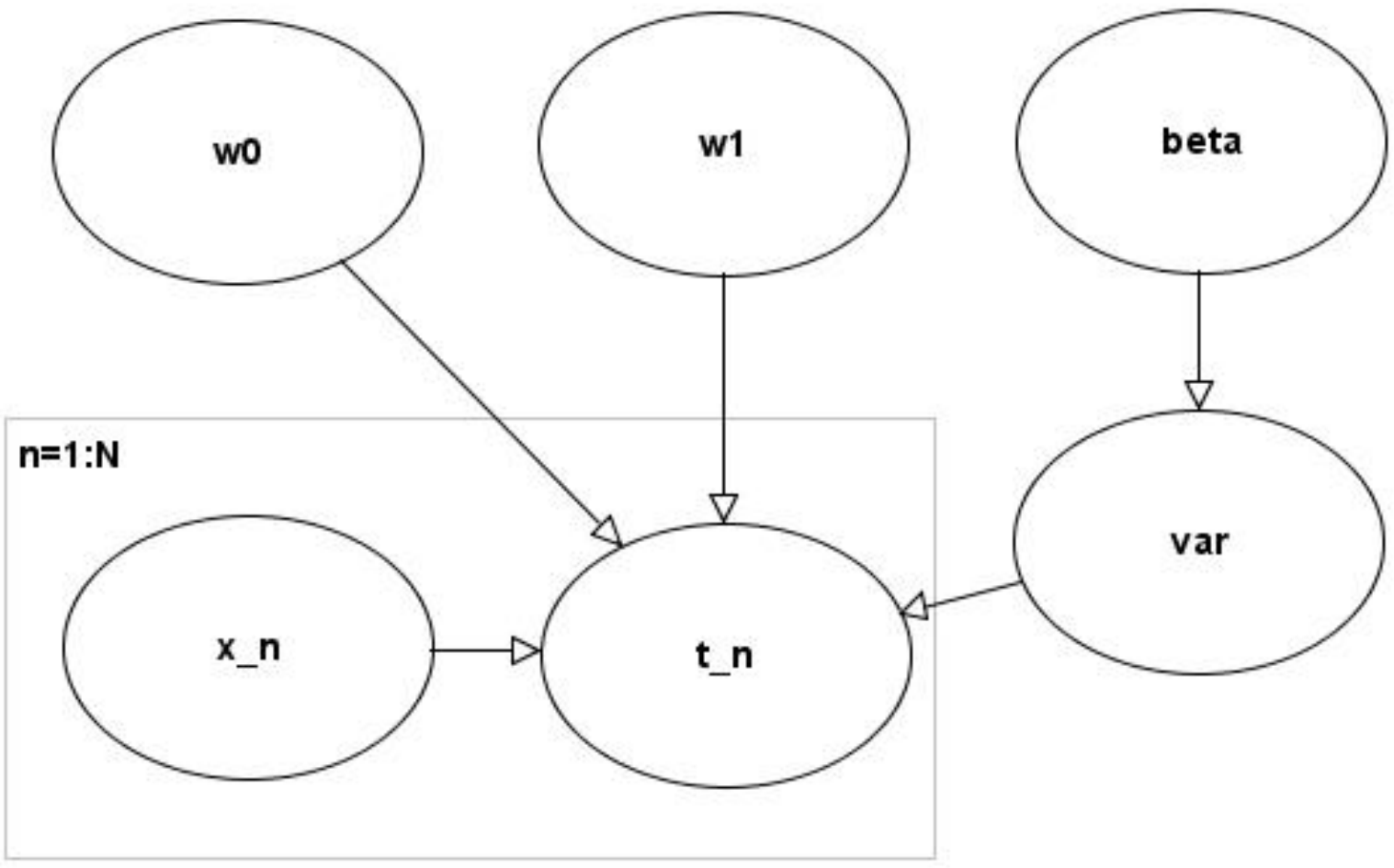

(a)

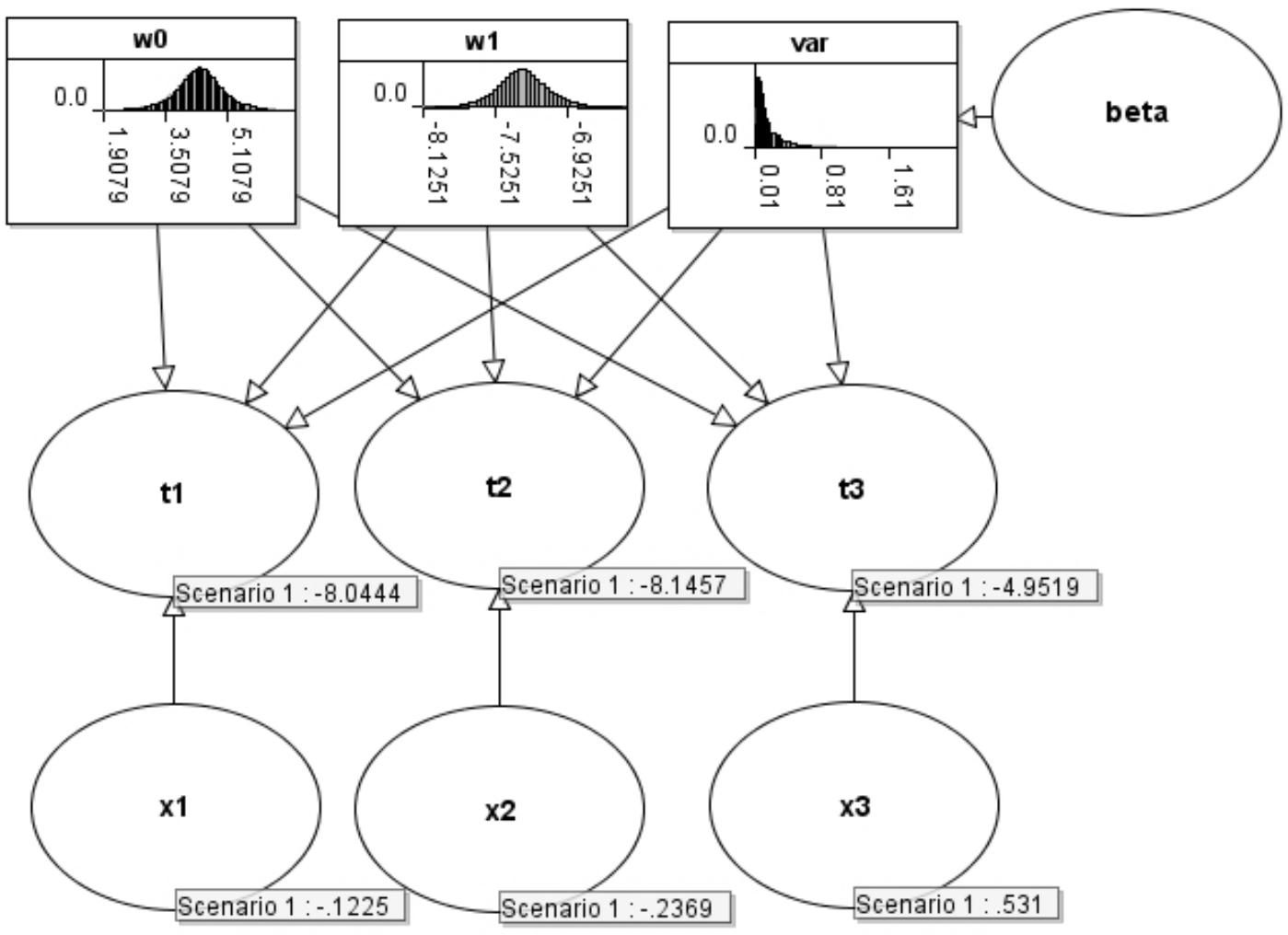

(b)

Figure 5.18 (a) Bayesian linear regression plate model

(b) Regression model when $n = 3$ for (a)

The original model in Figure 5.18 (b) is a sparse continuous model, so it can be converted to a 10 dimensional DCCD, and then subject to the BF process to produce a $\kappa_{10}$ full-BFG shown in Figure 5.19. Note that two variables, $t_2$ and $t_3$, have exact copies in the form of intermediate variables (this only happens when the original model is not DCCD) and given that these variables are observed the intermediate variables are set with the same observation values

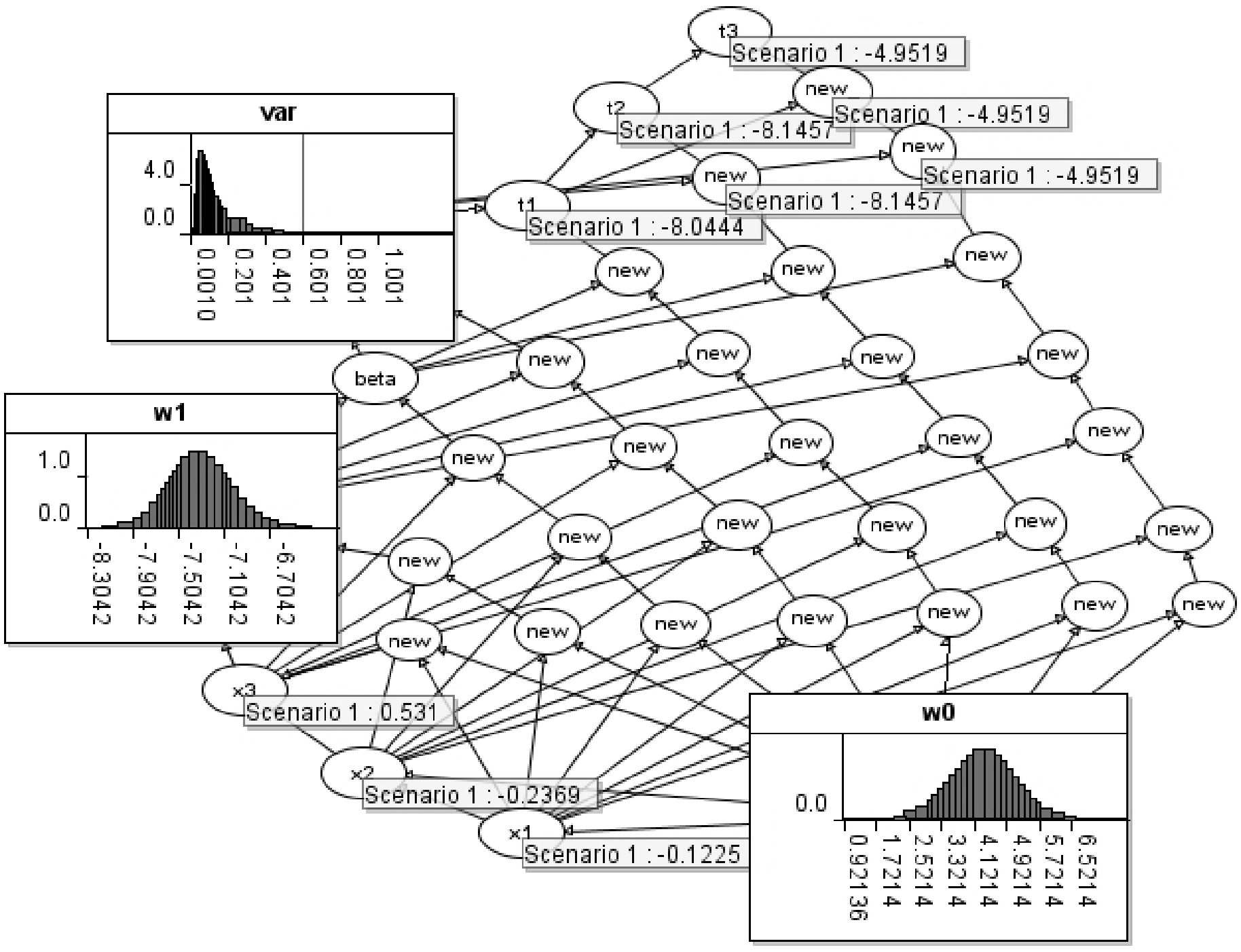

Figure 5.19 Converted $\kappa_{10}$ BFG model from Figure 5.18 (b) with observations

**(x,t) = {(-0.1225, -8.0444), (-0.2369, -8.1457), (0.531, -4.9519)}**

Table 5.6 data observations used in this experiment

We use the data in Table 5.6 and compare the experimental results for two models in Figure 5.18 (b) and Figure 5.19 using DDJT and DDBP, with 50 DD iterations on each, as well as against the Ordinary Least Square (OLS) solution.

The results are shown in Table 5.7.

| Parameters | Mean (DDJT) | Mean (DDBP) | Mean (OLS) | Variance (DDJT) | Variance (DDBP) | Variance (OLS) |
|---|---|---|---|---|---|---|
| $w_0$ | 4.3528 | 4.3413 | 4.3550 | 0.7660 | 0.8000 | 0.2340 |
| $w_1$ | -7.2952 | -7.300 | -7.2960 | 0.1000 | 0.0870 | 0.0275 |
| *var* | 0.2500 | 0.1975 | 0.0000 | 0.3200 | 0.1300 | 0.0400 |

Table 5.7 results comparing DDJT, OLS and DDBP

The results for DDJT and DDBP are very close, although the approximation of variable *var* is relatively poor. We believe that this is due to the discretization including a long tail and the bins' range in the tail could be slightly different. (This is supported by that fact that the median for *var* is 0.11 for both DDJT and DDBP).

### 5.4.7 Experiment 7: Aggregation of inter-dependent random variables

Final experiment illustrates the Bayesian risk aggregation of inter-dependent random variables. This is particularly an application of high dimensional models we have performed using DDBP algorithm.

The sum of $d$ dependent random variables is used in finance to calculate an overall capital charge in order to calculate the risk position $S_d = X_1 + X_2 + ... + X_d$ derived from a portfolio of $d$ random valuations.

Suppose we have a multivariate vector $\mathbf{X}$ that represents some kind of portfolio of risks held. $\mathbf{X} = (X_1,..., X_d)'$ is defined on some probability space. If the distribution function of $\mathbf{X}$ is $F_{\mathbf{X}}$, the aggregation of the risks portfolio is the sum of $X_1,..., X_d$, that takes the form:

$$P[X_1 + ... + X_d \le s] = \int_{\Im(s)} dF_{\mathbf{X}}(x_1,..., x_d), \; s \in \Re, \qquad (5.7)$$

where $\Im(s) = \{x \in \Re^d : \sum_{i=1}^{d} x_i \le s\}$.

Numerical evaluation of Equation 5.7 is a rather onerous task; one often has to rely on the bespoke design of MC methods. An alternative to this is the analytic numerical algorithm, AEP, (Arbenz, Embrechts, & Puccetti, 2011), which uses hyper-cubes to iteratively approximate the target distribution by calculating a simplex with probability masses. This algorithm is theoretically promising but in practice can only cope with five or fewer dimensions because of its space complexity.

Our algorithm has the advantages that it has avoided the space constraints as it reduced the computation complexity from exponential to polynomial, and if the joint distribution can be expressed by a conditional form, it retains its generalization for any conditionally deterministic operations not limited to Equation 5.7.

Consider the sum of a seven dimensional CG-DCCD model. For a seven dimensional model we produce the full-BF model shown in Figure 5.20 (a). Figure 5.20 (b) shows the full-BF transformed BN when we add the sum of seven dimensions to Figure 5.20 (a). Here intermediate variables $E_1,...,E_6$ are the incremental sum of the original variables $X_1,...,X_7$, where $E_5$ yields the total aggregation. The structure of Figure 5.20 (b) simply includes the addition of an individual path $E_1,...,E_6$ along $X_1$ to $X_7$ to Figure 5.20 (a). This operation makes no difference to the structure itself and results in a DCCD, as we would expect since Equation 5.7 is BF decomposable. Likewise, more general arithmetical operations defined on the vector $\mathbf{X}$ are just as applicable and would not change the topology of the full-BFG model. If we rearrange the labelled variables' positions in Figure 5.20 (b) it is equivalent to a $\kappa_8$ full-BFG, as shown in Figure 5.20 (c).

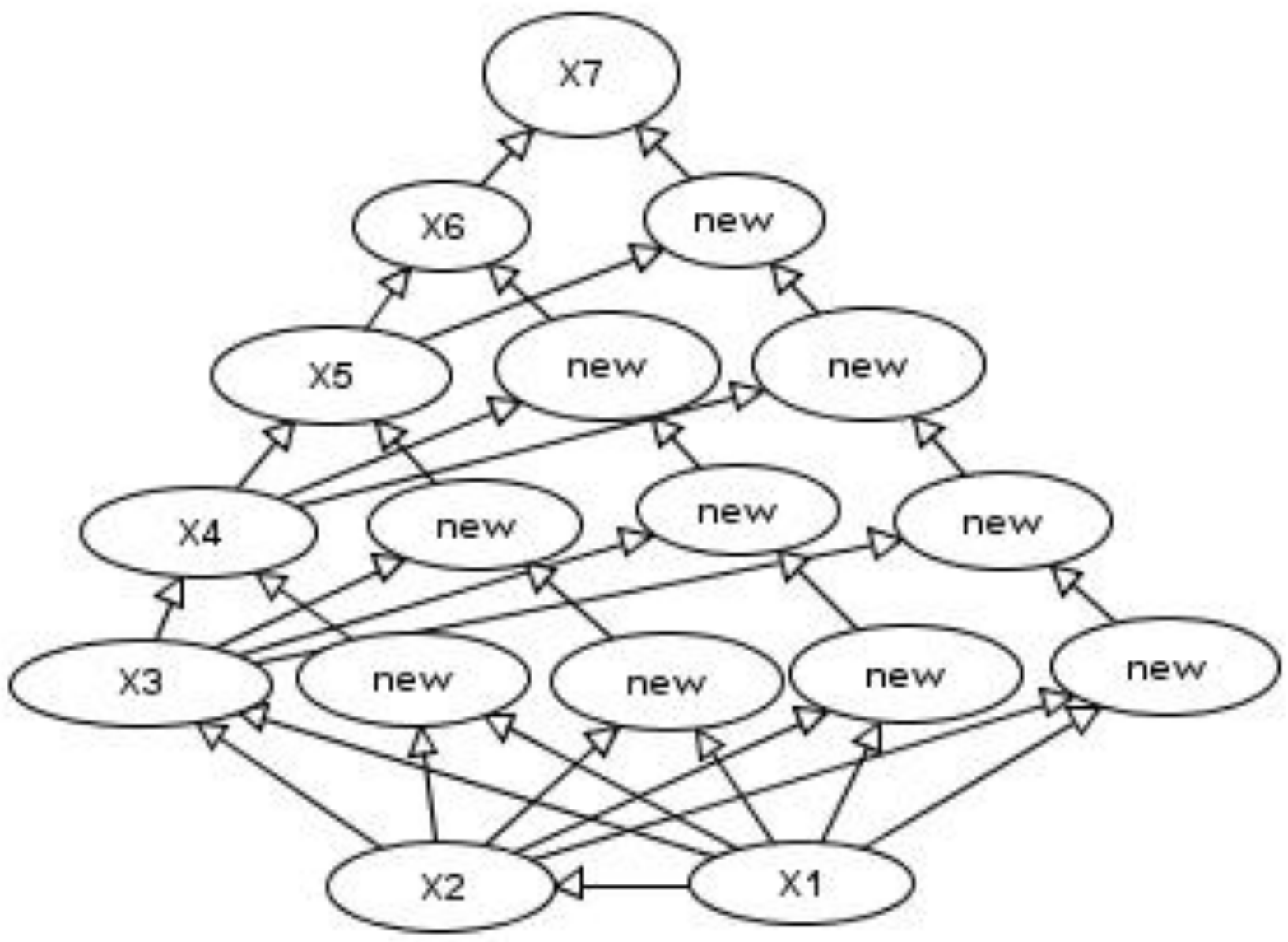

(a)

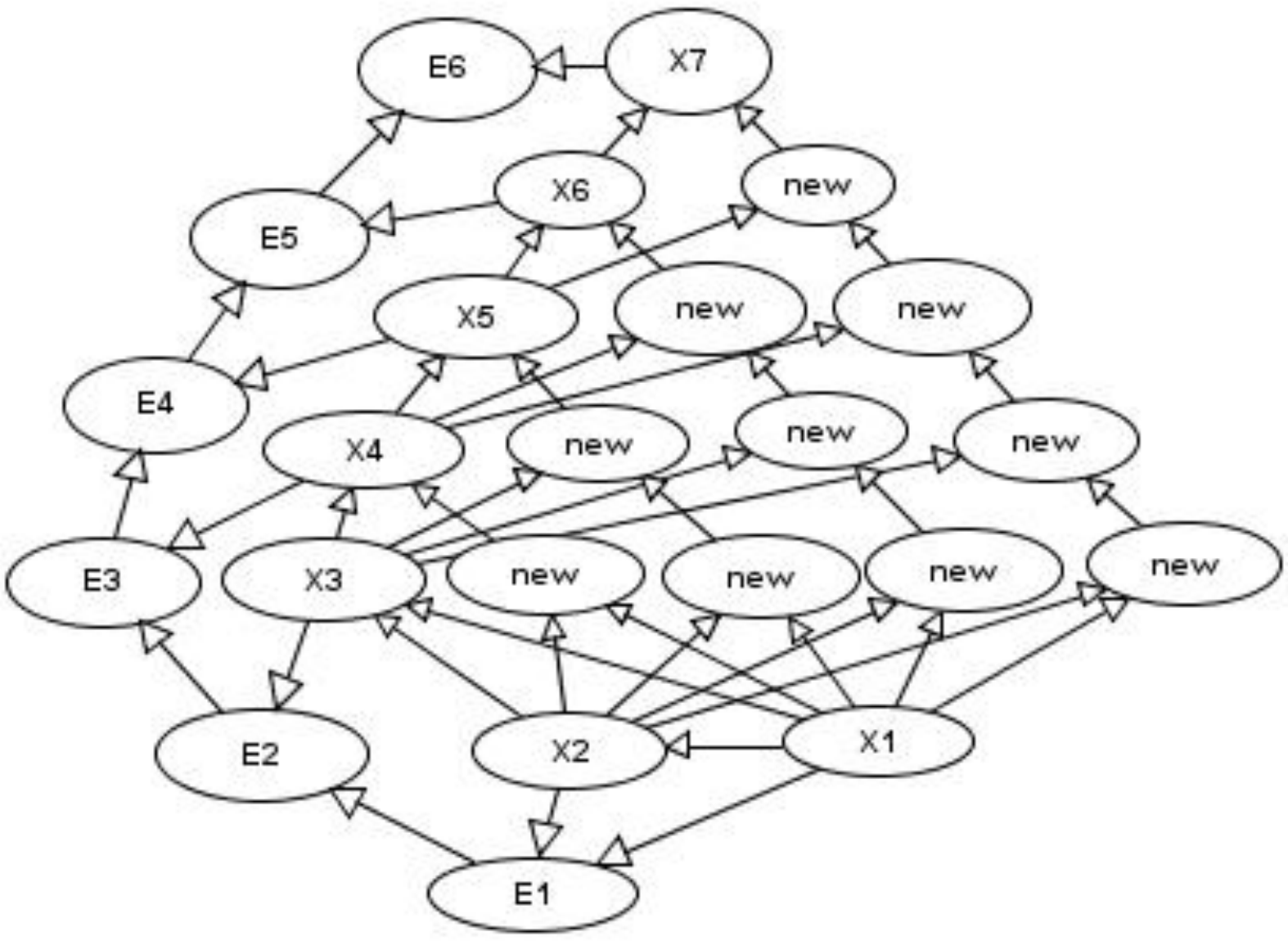

(b)

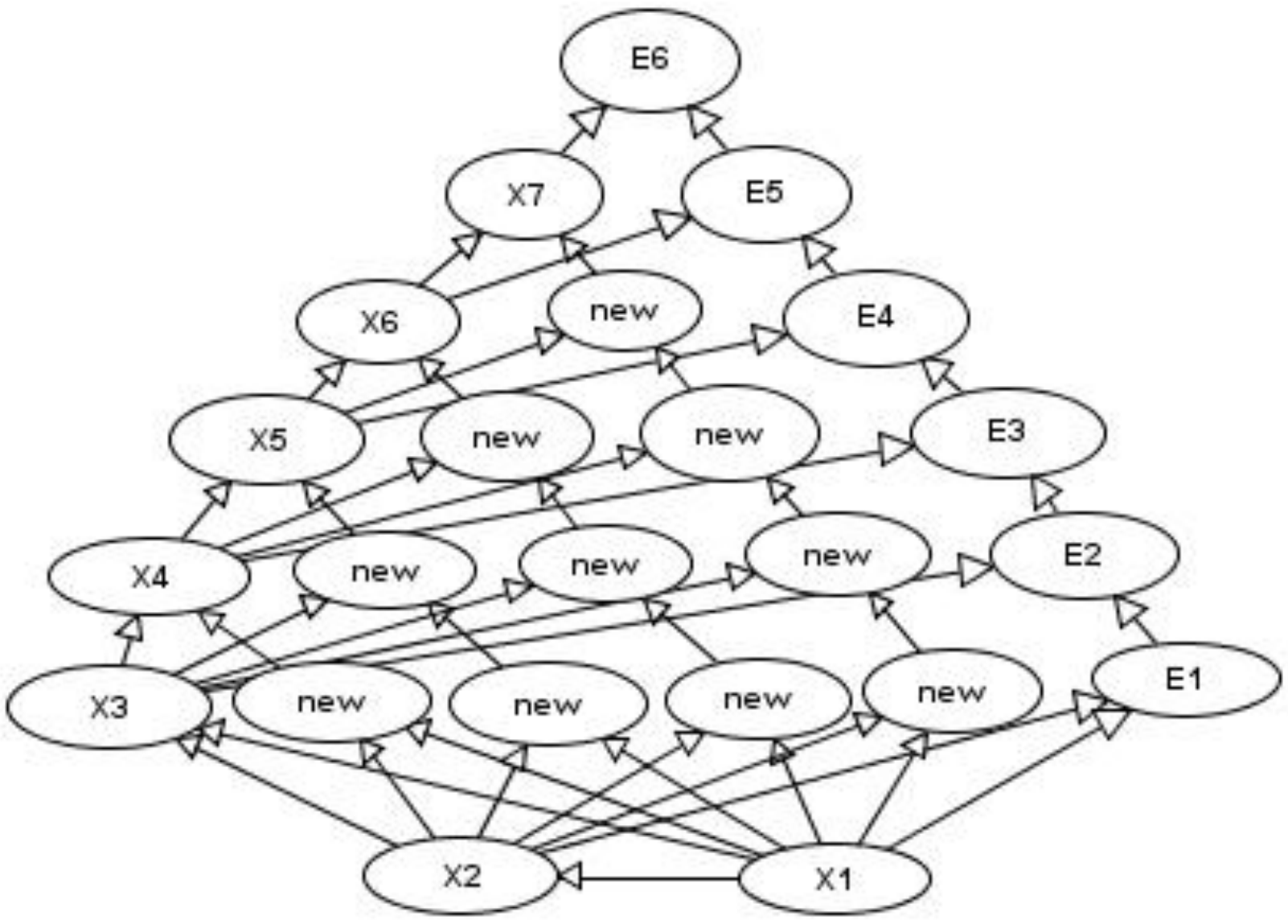

(c)

Figure 5.20 (a): $\kappa_7$ full-BFG model

(b): full-BFG transformed model for the sum of 7 dimensional CG-DCCD model

(c): Equivalent $\kappa_8$ full-BFG structure with (b)

The results of the approximation using a CG-DCCD compared to the exact values are shown in Table 5.8.

| Node | Exact Mean | DDBP Mean | Exact SD | DDBP SD |
|---|---|---|---|---|
| $E_1$ | 5 | 4.9970 | 3.77 | 4.12 |
| $E_2$ | 9 | 8.9820 | 5.85 | 6.05 |
| $E_3$ | 14 | 13.947 | 8.26 | 8.54 |
| $E_4$ | 20 | 19.942 | 11.00 | 11.06 |
| $E_5$ | 27 | 26.916 | 14.07 | 13.90 |
| $E_6$ | 35 | 34.956 | 17.47 | 17.15 |

Table 5.8 results of the sum of 7 dimensional CG-DCCD model

## 5.5. Summary

This chapter developed an inference approach which called DDBP by combining Dynamic Discretization and Belief Propagation algorithms to perform efficient inference for DCCD models. It overcomes the computational limitations of discrete approximation algorithms, such as DDJT, by limiting cluster size whilst ensuring the approximation retains desired properties (i.e. maxent-normal) in the triplet heuristics. The new DDBP approach uses an architecture comprising a series of algorithms, which take a BN and convert it to region graph where inference is done using Generalized Belief Propagation (GBP). This is done in two steps. Firstly we convert a BN to a DCCD and then into a full-Binary Factored Graph (full-BFG), to ensure it is I-equivalent to the original joint probability distribution. Next we take the full-BFG and use a Triplet Region Construction (TRC) algorithm to produce an optimized region graph upon which inference can take place. For discrete variables this is done directly on the region graph but when the model contains continuous variables we employ DD to approximate these variables.

The TRC algorithm we have developed is a variant of the CVM algorithm, designed to have three desirable properties for inference: a) it is acyclic and two-level, is b) balanced, in that the counting numbers are all one and c) it is maximum entropy normal. These properties mean that the algorithm is likely to converge quickly and produce accurate results. Unlike the JT algorithm, whose complexity is exponential in the worst case, the complexity of TRC is polynomial in all cases where the original joint distribution is BF-decomposable. TRC algorithm has close connections with join graph based approach which means it guarantees to converge in finite time.

With DDBP we can deal with arbitrary continuous variables and are not restricted to Gaussian families. Experiments undertaken to evaluate the algorithm include inference on two discrete binary full-BFG models and 10, 15 and 20 dimensional MGD models, with and without observations, a linear regression model and finally a large aggregation model. Results show that our algorithm converges and achieves comparably accurate results compared to exact solutions and competing approximation methods.

The DDBP algorithm offers a solution to performing Bayesian risk aggregation of inter-dependent random variables, i.e. illustrated in section 5.4.7. This positively answers research hypotheses 4 in Chapter 1, Section 1.2.

Note that to connect the work described in this Chapter to Chapter 4, it is easy to see that each *n-fold* convolution output of inter-dependent random variables can then be used to perform stochastic *N-fold* convolution using BFE.

# 6. Conclusions and Future Work

This thesis has addressed a general Bayesian risk aggregation framework, using a family of algorithms: BFE, TRC and DDBP. Together these algorithms positively satisfy the four hypotheses set out in Chapter 1 and, in sum, they provide general purpose approximate algorithms that provide accurate results.

Many popular financial methods can be accommodated to our approach and can now be extended to include causal risk factors. With regard to the BFE algorithm, on-going and future research is focused on more complex situations involving both copulas and common cause variables. The challenge here is to decompose these models into lower dimensional distributions where complexity is minimized by factorization. One final area of interest includes optimization of the results such that we might choose a set of actions in the model that maximize returns for minimum risk: deconvolution looks promising here.

This research proposed a new inference approach, DDBP, which is general purpose and improves the current discrete inference approaches for many classes of Bayesian network, and not simply those encountered in the area of financial applications. Further work to the DDBP and TRC algorithms include optimizing the efficiency and resolution of the prototype implementations of the algorithms. For example the current DD splitting strategy is based on univariate densities, whilst a finer-grained resolution can be achieved by splitting on the joint distribution (Langseth, Neil, & Marquez, 2013), i.e. on the clusters at the cost of efficiency. Likewise further work would demand greater focus on converting discrete and hybrid BNs to DCCD-BFGs using appropriate factorization methods. The search for methods to model arbitrary distributions with general dependencies, such as pair copulas, which can be represented using our factorization and discretization architecture, is ongoing so that inference may be carried out for arbitrary models. A thorough comparison with other related algorithms, such as EP and PBP, will be conducted in future research.

# Appendix

## Part A: Proof of Compound Density Factorization

Suppose the frequency probability density function of $N$, with sample space $Z$ is defined by:

$$f_N(x) = P(N = x) = P(\{z_j \in Z : N(z) = x\}) = a_j,\ j = 0,...,length(Z)$$

If we denote $S_j^{*z_j}$ as an *n-fold* convolution then equation (4.14) can be rewritten as:

$$F_{j-1} = P(E_{j-1} = True)P(F_{j-2}) + P(E_{j-1} = False)P(S_j^{*z_j})$$

And this can be expressed as:

$$(a_0 + a_1 + ... + a_{j-1})P(F_{j-2}) + a_j P(S_j^{*z_j})$$

$$= (a_0 + ... + a_{j-1})\left\{\frac{1}{a_0 + ... + a_{j-1}}(a_0 + ... + a_{j-2})P(F_{j-3}) + \frac{a_{j-1}}{a_0 + ... + a_{j-1}}P(S_{j-1}^{*z_{j-1}})\right\} + a_j P(S_j^{*z_j})$$

$$= (a_0 + ... + a_{j-2})P(F_{j-3}) + a_{j-1}P(S_{j-1}^{*z_{j-1}}) + a_j P(S_j^{*z_j})$$

...

$$= (a_0 + a_1 + a_2)\left\{\frac{a_0 + a_1}{a_0 + a_1 + a_2}P(F_0) + \frac{a_2}{a_0 + a_1 + a_2}P(S_2^{*z_2})\right\} + a_3 P(S_3^{*z_3}) + ... + a_j P(S_j^{*z_j})$$

$$= (a_0 + a_1)P(F_0) + a_2 P(S_2^{*z_2}) + ... + a_j P(S_j^{*z_j})$$

$$= (a_0 + a_1)\left\{\frac{a_0}{a_0 + a_1}P(S_0^{*z_0}) + \frac{a_1}{a_0 + a_1}P(S_1^{*z_1})\right\} + a_2 P(S_2^{*z_2}) + ... + a_j P(S_j^{*z_j})$$

$$= a_0 P(S_0^{*z_0}) + a_1 P(S_1^{*z_1}) + a_2 P(S_2^{*z_2}) + ... + a_j P(S_j^{*z_j})$$

This is equivalent to equation (4.3).

## Part B: Proof of BFE deconvolution is equivalent to deconvolution in the full model

Here we provide a small proof that an *N-fold* deconvolution performed by BFE produces the same results as one performed on a full BN.

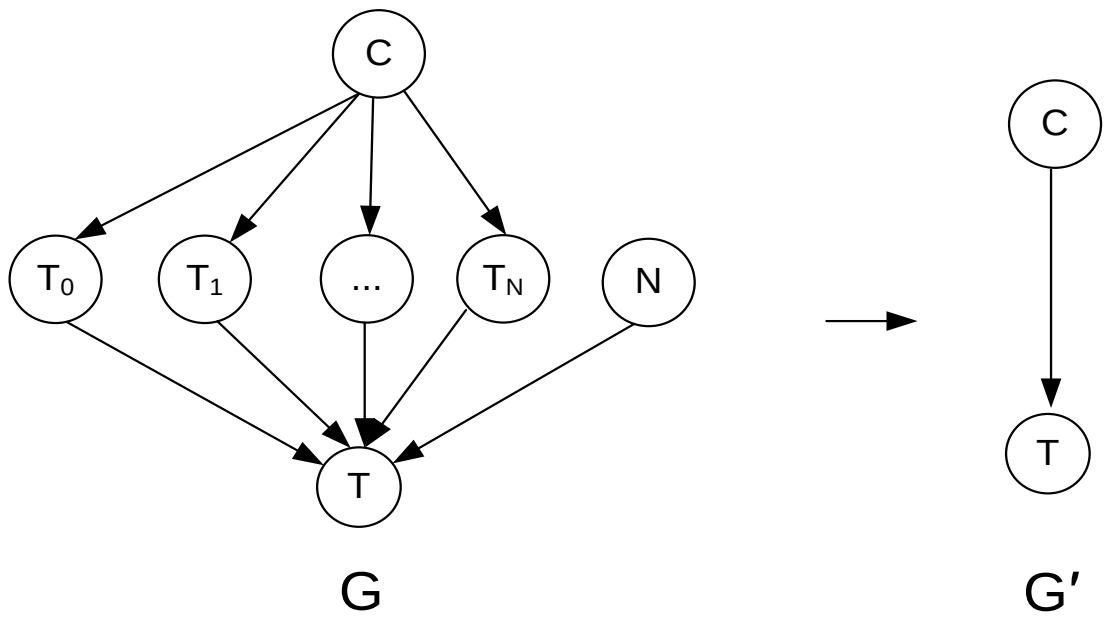

Figure A.1 Variable eliminations for *N-fold* deconvolution

As before, each $T_j, j = 0,...,N$ is a constant *n-fold* convolution node, $N$ is discretized frequency distribution controlling the weightings for each $T_j$ component, so $N$ is discrete and finite. For simplicity we consider a BN mode, $G$, without binary factorization and where the severity variables, $S_i$, can be assumed to have been eliminated already. The variable **C** is vector of common cause variables $\{C_0,...,C_i\}$. Thus the full model is represented by graph $G$ in Figure A.1 and the BFE reduced model, after convolution, is represented by $G'$.

In BN $G$ let $U = \{T_0...T_N\}$, therefore:

$$P(C \mid T = t_0) = \frac{P(C, T = t_0)}{P(T = t_0)} = P(C, T = t_0) \text{ since } P(T = t_0) = 1$$

$$\Rightarrow P(C \mid T = t_0) = \sum_{U,N} P(C, U, T = t_0, N)$$

$$= \sum_{U,N} P(T = t_0 \mid U, N) P(U \mid C) P(N) P(C)$$

$$= P(C) \sum_{U} P(T = t_0 \mid U) P(U \mid C)$$

Now we know that the variables in $U$ are mutually exclusive and so:

$$P(T, U, C, N) = \left\{ a_0 P(T \mid T_0, C) + a_1 P(T \mid T_1, C) + ... + a_L P(T \mid T_L, C) \right\} P(C)$$

and:

$$\Rightarrow P(T = t_o \mid U, N)$$

$$= P(C) \sum_{U} \left\{ a_0 P(T = t_o \mid T_0, C) + a_1 P(T = t_o \mid T_1, C) + ... + a_L P(T = t_o \mid T_L, C) \right\} \quad \text{(A.1)}$$

$$= P(T = t_0 \mid C) P(C)$$

From BN $G'$, and using Bayes':

$$P(C \mid T=t_0) = \frac{P(T=t_0 \mid C)P(C)}{P(T=t_0)} = P(T=t_0 \mid C)P(C) \tag{A.2}$$

(A.1) equals (A.2).

The proof still holds if $N$ is dependent on $C$ since only the mixture weights will change.

## *Part C: Proof of the TRC region graph for $\kappa_n$ full-BFG is maxent-normal with correct counting numbers*

In the general case we can prove that, for a $\kappa_n$ full-BFG, we can always construct $H_{RG}$ in the form of Equation 5.3 because each triplet's entropy can be cancelled by its connected second level regions and the single variables it contains. The proof can be simplified to satisfying the following two constraints:

1. The entropy for each first level region triplet minus the entropies of its connected second level region plus the single variables exclusively contained in this first level region triplet, but not in any of the connected second level regions, suffices to construct an $I$ term.
2. All single variables contained in an $I$ term must not be contained, as a single variables, in any other $I$ term.

The total entropy then satisfies $\sum_{i=1}^{N} H(b_i)$, where $N$ is the number of all variables in the region graph.

By inspecting the region connections in the region graph, the construction order for $I$ terms may not be unique, but if the above two constrains are satisfied it suffices to satisfy the maxent-normal property for the region graph. Also, for the second constraint, if the number of all variables being added to the $I$ terms is equal to $N$, then each must be added once, guaranteeing a counting number of one. To do otherwise would mean over or under counting the counting number for some variables.

Assuming the second constraint is satisfied, the proof of the first constraint can be further simplified by the fact that the entropies of all triplets minus the entropies of all second level regions are exactly equal to the entropy $\sum_{i=1}^{N} H(b_i)$. Therefore the required number of variables to be added, to construct all $I$ terms, is exactly $N$. We therefore need to prove that required number of variables to be added to the $I$ terms is equal to $N$.

Our region graph entropy is the sum of all region entropies, Equation A.0:

$$H_{RG} = H(\sum_{t=1}^{L1} c_t R_t + \sum_{i=1}^{L2} c_i R_i) \quad \text{(A.0)}$$

where there are $M$ first level regions, $R_t$ , $t = 1,....,M$ and $L2$ second level regions, $R_i$, $i = 1,....,L2$. The counting numbers for the first and second level regions is $c_t$ and $c_i$.

We can write Equation A.0 in the form of Equation 5.2 thus:

$$H_{RG} = \sum_{i=1}^{N} H(b_i) + \left[ H(\sum_{t=1}^{M} c_t R_t + \sum_{i=1}^{L2} c_i R_i) - \sum_{i=1}^{N} H(b_i) \right],$$

And $I\ terms = \left[ H(\sum_{t=1}^{M} c_t R_t + \sum_{i=1}^{L2} c_i R_i) - \sum_{i=1}^{N} H(b_i) \right]$, now $H_{RG}$ is maximised when the $I$ terms are minimised and equal to zero and this is equivalent to finding the value $\gamma$ from the region graph that satisfies Equation A.1:

$$I\ terms = \sum_{a=1}^{M} I(b_a) = H(\sum_{t=1}^{M} c_t R_t + \sum_{i=1}^{L2} c_i R_i + \gamma) = 0 \quad \text{(A.1)}$$

Where $\gamma$ is the number of all single variables that need to be added to “balance” the region graph.

We then need to prove Equation (A.2):

$$H(\gamma) = -\sum_{i=1}^{N} H(b_i) \quad \text{(A.2)}$$

Where $\sum_{i=1}^{N} H(b_i)$ is the sum of the entropies of the original and intermediate variables in the region graph, of which there are $n+(n-2)(n-3)/2=(n^2-3n+6)/2$. So Equation A.2 can be rewritten as:

$$H(\gamma)=-\sum_{i=1}^{N} H(b_i)=-H((n^2-3n+6)/2) \qquad \text{(A.3)}$$

Next we need to determine $H(\gamma)$ from the region graph and determine whether it satisfies Equation A.1 and is equal to Equation A.3. We know that all first level regions have identical counting number, so Equation A.1 can be reduced to $H(\sum_{t=1}^{M} R_t + \sum_{i=1}^{L2} c_i R_i + \gamma) = 0$, which is equivalent to Equation A.4.

$$H(\gamma)=-\left[ H(\sum_{t=1}^{M} R_t)+H(\sum_{i=1}^{L2} c_i R_i) \right] \qquad \text{(A.4)}$$

We can convert the log entropy forms into weights, because the final log terms can be omitted on both sides of Equation (A.4), which is then equivalent to Equation A.3.

$$w(\gamma)=-\left[ w(\sum_{t=1}^{M} R_t)+w(\sum_{i=1}^{L2} c_i R_i) \right] \qquad \text{(A.5)}$$

where $w$ is the weight of each region, e.g. a triplet has three variables so the weight is three.

The following steps 1-4 compute the term $w(\sum_{t=1}^{M} R_t)$ (the weights of all triplets in Equation A.5, and steps 5-8 compute the term $w(\sum_{i=1}^{L2} c_i R_i)$ (the weights of all intersections in Equation A.5.

We determine the term $w(\sum_{t=1}^{M} R_t)$ for all triplets in region level 1:

1. Let $n$ be the number of original nodes in a complete-BFG, $G'$, so the number of intermediate nodes in $G'$ is:

   $1+2+...+n-3=(n-2)(n-3)/2, n>3$

2. From the parent to child relationships in $G'$, the number of primary triplets is determined by the sum of the number of original variables and intermediate variables minus 2, as there are two factors absorbed in triplets. So we have $n-2+(n-2)(n-3)/2$ primary triplets.

3. The number of interaction triplets is the number of moral edges and it is also the number of intermediate nodes, so we have $(n-2)(n-3)/2$ interaction triplets.

4. The number of first level triplets is then:

$$M = n-2+(n-2)(n-3)/2+(n-2)(n-3)/2 = (n-2)^2$$

   and the weight of all triplets is then $w(\sum_{t=1}^{M} R_t) = 3(n-2)^2$.

We next determine the term $w(\sum_{i=1}^{L2} c_i R_i)$ for intersections in region level 2.

5. The number of second level intersections is determined by the number of first level triplets plus one and is $(n-2)^2+1$.

6. There are $n-2$ cognate intersections being pruned with counting number $3-n$ (equivalent with $n-3$ intersections having counting number -1) and other non-pruned intersections all with counting number -1. The number of all intersections with two nodes is:

$$-2[(n-3)+[(n-2)^2+1-(n-2)-1]] = -2[(n-2)^2-1]$$

7. The weight of all pruned intersections is the sum of their counting numbers, as they all only contain single variables:

$$-(n-3)-(n-4)-\ldots-1 = -(n-2)(n-3)/2$$

8. Summing steps 6 and 7 gives the total weights from the intersections:

$$w(\sum_{i=1}^{L2} c_i R_i) = -2[(n-2)^2-1]-(n-2)(n-3)/2$$

We can now determine $w(\gamma)$ by adding step 8 to step 4:

$$w(\gamma) = -\left[ w(\sum_{t=1}^{M} R_t) + w(\sum_{i=1}^{L2} c_i R_i) \right]$$
$$\Rightarrow w(\gamma) = -\left[ 3(n-2)^2 - 2[(n-2)^2 - 1] - (n-2)(n-3)/2 \right]$$
$$\Rightarrow w(\gamma) = -(n^2 - 3n + 6)/2$$

The entropies for these variables are given by Equation A.6:

$$H(\gamma) = -H((n^2 - 3n + 6)/2) \qquad \text{(A.6)}$$

We have proven that Equation A.6 is identical to Equation A.3, as required. Q.E.D.

## *Part D: Parameterization for experiments in section 5.4tion for experiments in section 5.4*

*Parameterization for Experiment 1*

| | | | | | | | | |
|---|---|---|---|---|---|---|---|---|
| $X_1$ | 0.3 | 0.7 | | | | | | |
| $X_2\|X_1$ | 0.2 | 0.8 | 0.6 | 0.4 | | | | |
| $X_3\|X_1,X_2$ | 0.4 | 0.6 | 0.7 | 0.3 | 0.5 | 0.5 | 0.1 | 0.9 |
| $E_1\|X_1,X_2$ | 0.5 | 0.5 | 0.4 | 0.6 | 0.1 | 0.9 | 0.3 | 0.7 |
| $E_2\|X_1,X_2$ | 0.2 | 0.8 | 0.3 | 0.7 | 0.4 | 0.6 | 0.5 | 0.5 |
| $X_4\|X_3,E_1$ | 0.1 | 0.9 | 0.2 | 0.8 | 0.7 | 0.3 | 0.4 | 0.6 |
| $E_3\|X_3,E_2$ | 0.3 | 0.7 | 0.2 | 0.8 | 0.5 | 0.5 | 0.9 | 0.1 |
| $X_5\|E_3,X_4$ | 0.8 | 0.2 | 0.4 | 0.6 | 0.4 | 0.6 | 0.7 | 0.3 |

Table 1 NPT settings for Experiment 1

(the order of parents are labelled as above)

*Parameterization for Experiment 2*

| | | | | | | | | |
|---|---|---|---|---|---|---|---|---|
| $X_1$ | 0.3 | 0.7 | | | | | | |
| $X_2\|X_1$ | 0.2 | 0.8 | 0.3 | 0.7 | | | | |
| $X_3\|X_1,X_2$ | 0.2 | 0.8 | 0.4 | 0.6 | 0.6 | 0.4 | 0.8 | 0.2 |
| $E_{27}\|X_1,X_2$ | 0.2 | 0.8 | 0.4 | 0.6 | 0.6 | 0.4 | 0.8 | 0.2 |
| $E_{71}\|X_1,X_2$ | 0.2 | 0.8 | 0.4 | 0.6 | 0.6 | 0.4 | 0.8 | 0.2 |
| $E_{131}\|X_1,X_2$ | 0.2 | 0.8 | 0.4 | 0.6 | 0.6 | 0.4 | 0.8 | 0.2 |
| $E_{209}\| X_1,X_2$ | 0.47 | 0.53 | 0.1 | 0.9 | 0.17 | 0.83 | 0.23 | 0.77 |
| $E_{294}\| X_1,X_2$ | 0.5 | 0.5 | 0.23 | 0.77 | 0.29 | 0.71 | 0.33 | 0.67 |
| $X_4\|E_{27},X_3$ | 0.4 | 0.6 | 0.5 | 0.5 | 0.6 | 0.4 | 0.7 | 0.3 |
| $E_{68}\| E_{71},X_3$ | 0.4 | 0.6 | 0.5 | 0.5 | 0.6 | 0.4 | 0.7 | 0.3 |
| $E_{128}\| E_{131},X_3$ | 0.4 | 0.6 | 0.5 | 0.5 | 0.6 | 0.4 | 0.7 | 0.3 |
| $E_{206}\| E_{209},X_3$ | 0.375 | 0.625 | 0.41 | 0.59 | 0.44 | 0.56 | 0.47 | 0.53 |
| $E_{289}\| X_3,E_{294}$ | 0.33 | 0.67 | 0.41 | 0.59 | 0.375 | 0.625 | 0.44 | 0.56 |
| $X_5\|X_4,E_{68}$ | 0.23 | 0.77 | 0.33 | 0.67 | 0.29 | 0.71 | 0.375 | 0.625 |
| $E_{123}\| X_4,E_{128}$ | 0.44 | 0.56 | 0.375 | 0.625 | 0.44 | 0.56 | 0.17 | 0.83 |
| $E_{201}\| X_4,E_{206}$ | 0.29 | 0.71 | 0.23 | 0.77 | 0.17 | 0.83 | 0.375 | 0.625 |
| $E_{284}\| E_{289},X_3$ | 0.41 | 0.59 | 0.375 | 0.625 | 0.17 | 0.83 | 0.375 | 0.625 |
| $X_6\|E_{123},X_5$ | 0.17 | 0.83 | 0.23 | 0.77 | 0.29 | 0.71 | 0.33 | 0.67 |
| $E_{194}\| E_{201},X_5$ | 0.44 | 0.56 | 0.17 | 0.83 | 0.17 | 0.83 | 0.23 | 0.77 |
| $E_{279}\| E_{284},X_5$ | 0.1 | 0.9 | 0.23 | 0.77 | 0.17 | 0.83 | 0.23 | 0.77 |
| $X_7\|E_{194},X_6$ | 0.33 | 0.67 | 0.375 | 0.625 | 0.23 | 0.77 | 0.17 | 0.83 |
| $E_{274}\| X_6,E_{279}$ | 0.1 | 0.9 | 0.23 | 0.77 | 0.17 | 0.83 | 0.23 | 0.77 |
| $X_8\|X_7,E_{274}$ | 0.33 | 0.67 | 0.23 | 0.77 | 0.17 | 0.83 | 0.23 | 0.77 |

Table 2 NPT settings for Experiment 2

*Parameterization for CG-DCCD models (Experiments 3, 4, 5, 7):*

This shows the mean vector and covariance matrix for 20 dimensional CG-DCCD model setting. The lower dimensions parametric used in experiments are compatible.

| **Mean** | **Cov.** | $X_1$ | $X_2$ | $X_3$ | $X_4$ | $X_5$ | $X_6$ | $X_7$ | $X_8$ | $X_9$ | $X_{10}$ | $X_{11}$ | $X_{12}$ | $X_{13}$ | $X_{14}$ | $X_{15}$ | $X_{16}$ | $X_{17}$ | $X_{18}$ | $X_{19}$ | $X_{20}$ |
|---|---|---|---|---|---|---|---|---|---|---|---|---|---|---|---|---|---|---|---|---|---|
| 2 | $X_1$ | 4 | 0.6 | 0.8 | 1 | 1.2 | 1.4 | 1.6 | 1.8 | 2 | 2.2 | 2.4 | 2.6 | 2.8 | 3 | 3.2 | 3.4 | 3.6 | 3.8 | 4 | 4.2 |
| 3 | $X_2$ | 0.6 | 9 | 1.2 | 1.5 | 1.8 | 2.1 | 2.4 | 2.7 | 3 | 3.3 | 3.6 | 3.9 | 4.2 | 4.5 | 4.8 | 5.1 | 5.4 | 5.7 | 6 | 6.3 |
| 4 | $X_3$ | 0.8 | 1.2 | 16 | 2 | 2.4 | 2.8 | 3.2 | 3.6 | 4 | 4.4 | 4.8 | 5.2 | 5.6 | 6 | 6.4 | 6.8 | 7.2 | 7.6 | 8 | 8.4 |
| 5 | $X_4$ | 1 | 1.5 | 2 | 25 | 3 | 3.5 | 4 | 4.5 | 5 | 5.5 | 6 | 6.5 | 7 | 7.5 | 8 | 8.5 | 9 | 9.5 | 10 | 10.5 |
| 6 | $X_5$ | 1.2 | 1.8 | 2.4 | 3 | 36 | 4.2 | 4.8 | 5.4 | 6 | 6.6 | 7.2 | 7.8 | 8.4 | 9 | 9.6 | 10.2 | 10.8 | 11.4 | 12 | 12.6 |
| 7 | $X_6$ | 1.4 | 2.1 | 2.8 | 3.5 | 4.2 | 49 | 5.6 | 6.3 | 7 | 7.7 | 8.4 | 9.1 | 9.8 | 10.5 | 11.2 | 11.9 | 12.6 | 13.3 | 14 | 14.7 |
| 8 | $X_7$ | 1.6 | 2.4 | 3.2 | 4 | 4.8 | 5.6 | 64 | 7.2 | 8 | 8.8 | 9.6 | 10.4 | 11.2 | 12 | 12.8 | 13.6 | 14.4 | 15.2 | 16 | 16.8 |
| 9 | $X_8$ | 1.8 | 2.7 | 3.6 | 4.5 | 5.4 | 6.3 | 7.2 | 81 | 9 | 9.9 | 10.8 | 11.7 | 12.6 | 13.5 | 14.4 | 15.3 | 16.2 | 17.1 | 18 | 18.9 |
| 10 | $X_9$ | 2 | 3 | 4 | 5 | 6 | 7 | 8 | 9 | 100 | 11 | 12 | 13 | 14 | 15 | 16 | 17 | 18 | 19 | 20 | 21 |
| 11 | $X_{10}$ | 2.2 | 3.3 | 4.4 | 5.5 | 6.6 | 7.7 | 8.8 | 9.9 | 11 | 121 | 13.2 | 14.3 | 15.4 | 16.5 | 17.6 | 18.7 | 19.8 | 20.9 | 22 | 23.1 |
| 12 | $X_{11}$ | 2.4 | 3.6 | 4.8 | 6 | 7.2 | 8.4 | 9.6 | 10.8 | 12 | 13.2 | 144 | 15.6 | 16.8 | 18 | 19.2 | 20.4 | 21.6 | 22.8 | 24 | 25.2 |
| 13 | $X_{12}$ | 2.6 | 3.9 | 5.2 | 6.5 | 7.8 | 9.1 | 10.4 | 11.7 | 13 | 14.3 | 15.6 | 169 | 18.2 | 19.5 | 20.8 | 22.1 | 23.4 | 24.7 | 26 | 27.3 |
| 14 | $X_{13}$ | 2.8 | 4.2 | 5.6 | 7 | 8.4 | 9.8 | 11.2 | 12.6 | 14 | 15.4 | 16.8 | 18.2 | 196 | 21 | 22.4 | 23.8 | 25.2 | 26.6 | 28 | 29.4 |
| 15 | $X_{14}$ | 3 | 4.5 | 6 | 7.5 | 9 | 10.5 | 12 | 13.5 | 15 | 16.5 | 18 | 19.5 | 21 | 225 | 24 | 25.5 | 27 | 28.5 | 30 | 31.5 |
| 16 | $X_{15}$ | 3.2 | 4.8 | 6.4 | 8 | 9.6 | 11.2 | 12.8 | 14.4 | 16 | 17.6 | 19.2 | 20.8 | 22.4 | 24 | 256 | 27.2 | 28.8 | 30.4 | 32 | 33.6 |
| 17 | $X_{16}$ | 3.4 | 5.1 | 6.8 | 8.5 | 10.2 | 11.9 | 13.6 | 15.3 | 17 | 18.7 | 20.4 | 22.1 | 23.8 | 25.5 | 27.2 | 289 | 30.6 | 32.3 | 34 | 35.7 |
| 18 | $X_{17}$ | 3.6 | 5.4 | 7.2 | 9 | 10.8 | 12.6 | 14.4 | 16.2 | 18 | 19.8 | 21.6 | 23.4 | 25.2 | 27 | 28.8 | 30.6 | 324 | 34.2 | 36 | 37.8 |
| 19 | $X_{18}$ | 3.8 | 5.7 | 7.6 | 9.5 | 11.4 | 13.3 | 15.2 | 17.1 | 19 | 20.9 | 22.8 | 24.7 | 26.6 | 28.5 | 30.4 | 32.3 | 34.2 | 361 | 38 | 39.9 |
| 20 | $X_{19}$ | 4 | 6 | 8 | 10 | 12 | 14 | 16 | 18 | 20 | 22 | 24 | 26 | 28 | 30 | 32 | 34 | 36 | 38 | 400 | 42 |
| 21 | $X_{20}$ | 4.2 | 6.3 | 8.4 | 10.5 | 12.6 | 14.7 | 16.8 | 18.9 | 21 | 23.1 | 25.2 | 27.3 | 29.4 | 31.5 | 33.6 | 35.7 | 37.8 | 39.9 | 42 | 441 |

Table 3 Mean and covariance matrix for 20 dimensional CG-DCCD model